\documentclass[]{uiuc_paper}
\pdfoutput=1

\usepackage{booktabs}
\usepackage{colortbl}
\usepackage{etoc}

\usepackage{amssymb}
\usepackage{amsmath}

\title{CryoGaussian: Heterogeneous Cryo-EM Reconstruction Made Explicit with Gaussian Splatting}

\ifdefined\affillogo
\author[1]{Ziwen Wang}{zw1663@nyu.edu}
\author[2]{Jiajun Fan}{jiajunf3@illinois.edu}
\author[2]{Chaoran Cheng}{chaoran7@illinois.edu}
\author[2]{Ruihan Guo}{guoruihan.sansi@gmail.com}
\author[2]{Yanru Qu}{yanruqu2@illinois.edu}
\author[1]{Juexiao Zhou}{juexiao.zhou@gmail.com}
\author[2]{Ge Liu}{geliu@illinois.edu}

\affiliation[1]{\affillogo{\includegraphics{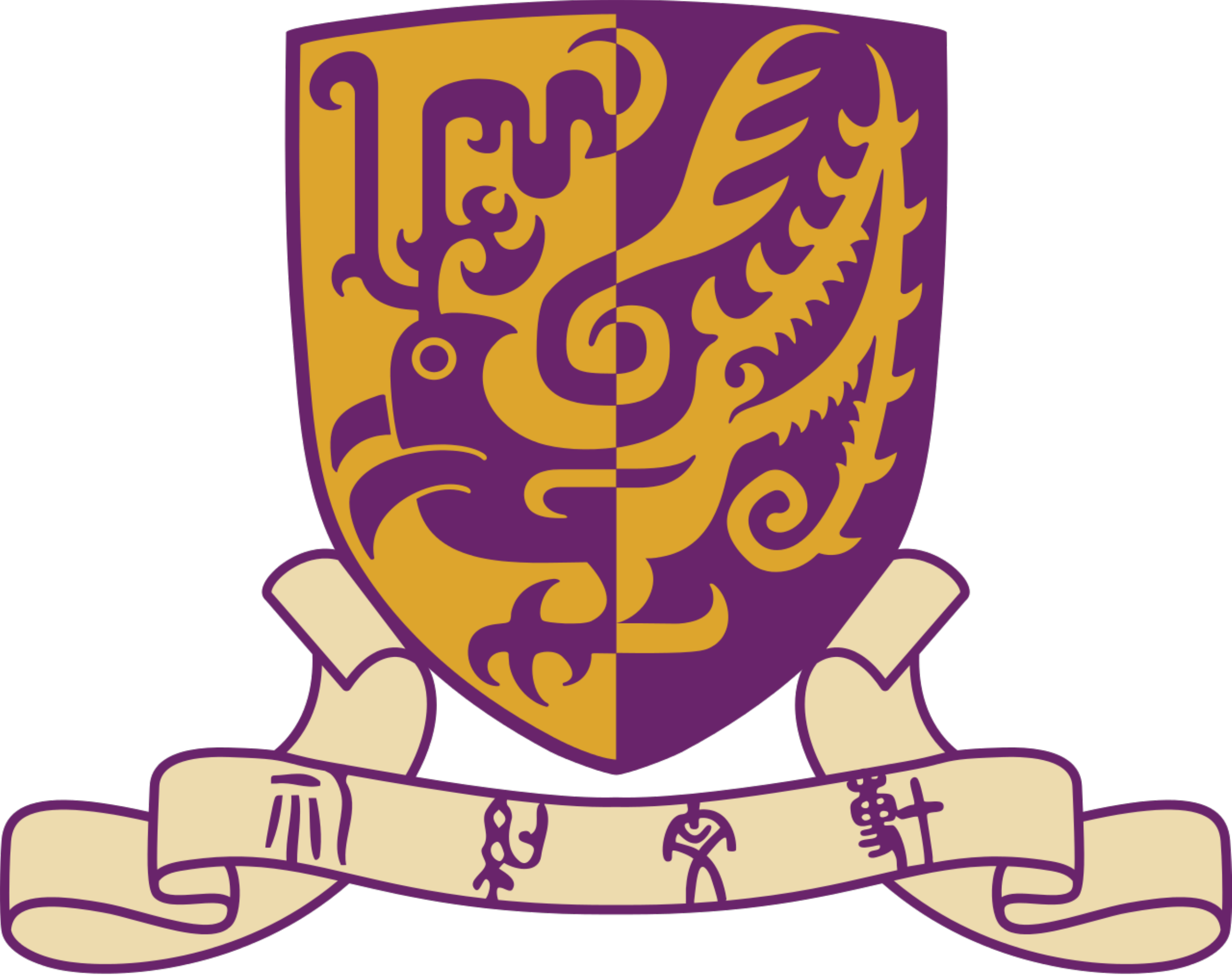}}{School of Data Science, The Chinese University of Hong Kong, Shenzhen}}
\affiliation[2]{\affillogo{\includegraphics{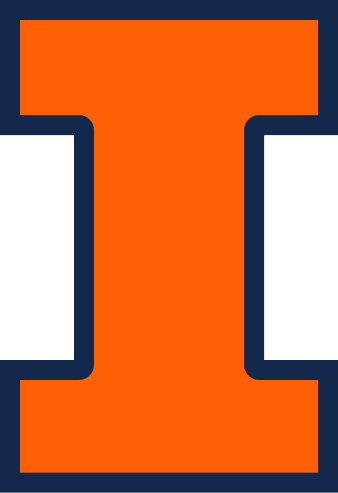}}{Siebel School of Computing and Data Science, University of Illinois Urbana-Champaign}}
\else
\author{Ziwen Wang}\affiliation{School of Data Science, The Chinese University of Hong Kong, Shenzhen}\email{zw1663@nyu.edu}
\author{Jiajun Fan}\affiliation{Siebel School of Computing and Data Science, University of Illinois Urbana-Champaign}\email{jiajunf3@illinois.edu}
\author{Chaoran Cheng}\affiliation{Siebel School of Computing and Data Science, University of Illinois Urbana-Champaign}\email{chaoran7@illinois.edu}
\author{Ruihan Guo}\affiliation{Siebel School of Computing and Data Science, University of Illinois Urbana-Champaign}\email{guoruihan.sansi@gmail.com}
\author{Yanru Qu}\affiliation{Siebel School of Computing and Data Science, University of Illinois Urbana-Champaign}\email{yanruqu2@illinois.edu}
\author{Juexiao Zhou}\affiliation{School of Data Science, The Chinese University of Hong Kong, Shenzhen}\email{juexiao.zhou@gmail.com}
\author{Ge Liu}\affiliation{Siebel School of Computing and Data Science, University of Illinois Urbana-Champaign}\email{geliu@illinois.edu}
\fi

\abstract{Single-particle cryo-electron microscopy (cryo-EM) determines macromolecular
structures from noisy particle images. However, rare, transient compositional
states and subtle, continuous conformational variation are difficult to
resolve. We present CryoGaussian, a Gaussian splatting method for
heterogeneous cryo-EM reconstruction that fits an explicit Gaussian density
representation directly to the particle images through a differentiable
rasterizer rendering orthographic line integrals modulated by the contrast
transfer function. The Gaussians are initialized randomly, with their number
derived from the estimated spatial support of the molecular density and their minimum
size bounded by the Nyquist frequency of the particle images. To capture the
compositional and conformational variation across particle images, we learn
particle-image and per-Gaussian latents that jointly modulate the base
position, amplitude, and scale of each Gaussian. Each structural state is then
reconstructed by fitting the Gaussians conditioned on its inferred latent. On
CryoBench, CryoGaussian achieves state-of-the-art reconstruction accuracy for
both compositional and conformational heterogeneity. On experimental data, it
resolves a rare compositional state of the 50S ribosomal subunit (class C4,
1.3\% of particles) with rRNA helix 68 docked in the absence of the central
protuberance, and continuous conformational motion of the SF3b and helicase
subcomplexes of the pre-catalytic spliceosome. We further optimize the
rasterizer for additive cryo-EM projection, reducing training time by
12.8--28.0$\times$. Reconstructed density maps are available at
\url{https://huggingface.co/datasets/ZiwenWang28/CryoGaussian}.
}

\date{\today}

\begin{document}

\maketitle

\etocdepthtag.toc{mainmatter}
\hyphenation{Cryo-Bench Cryo-Hype Cryo-KRAQEN Cryo-SPIRE Cryo-Splat RECOVAR SOLVAR Ribo-sembly Dyna-Might}

\begin{figure*}[t]
    \centering
    \includegraphics[width=1\linewidth]{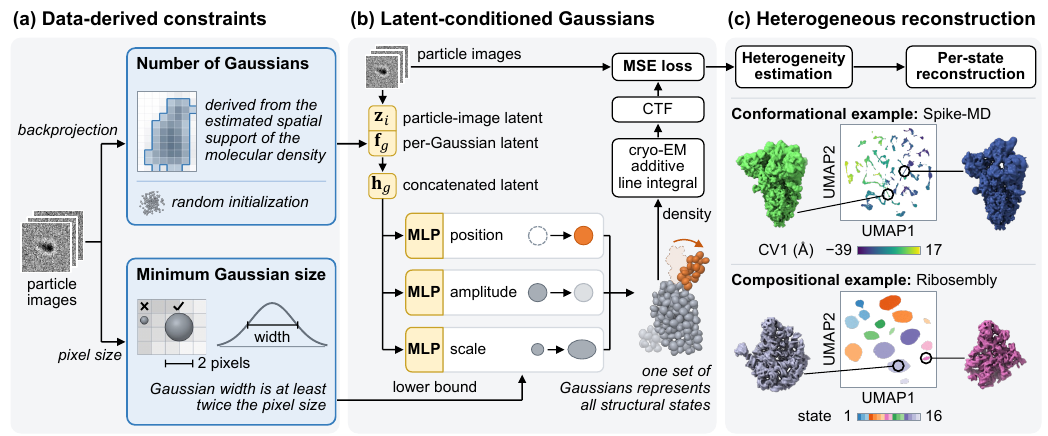}
\caption{\textbf{CryoGaussian overview.}
\textbf{(a)} The number of randomly initialized Gaussians is derived from
the estimated spatial support of the molecular density
(Appendix~\ref{app:space_bandwidth}), and the pixel size sets their minimum
size.
\textbf{(b)} Separate MLPs decode position, amplitude, and scale
modulations of each Gaussian from the concatenated particle-image and
per-Gaussian latents (Appendix~\ref{app:gaussian_parameterization}).
\textbf{(c)} Heterogeneity is estimated by jointly learning these latents, and
compositional and conformational states are then reconstructed from the
inferred latents.}
    \label{fig:main_flowchart}
\end{figure*}

\section{Introduction}

Single-particle cryo-electron microscopy (cryo-EM) reconstructs
three-dimensional biomolecular density at near-atomic resolution from noisy
two-dimensional particle images~\citep{cheng2015primer,yip2020atomic}.
Resolving compositional and conformational heterogeneity reveals molecular
organization and function, as illustrated by respiratory supercomplexes and
proteasome states~\citep{zheng2024supercomplexes,dong2019proteasome}.
Biological interpretation requires both reliable structural detail and an
account of the compositional and conformational states of the imaged
biomolecules.

\begin{table}[t]
    \centering
    \caption{\textbf{Training efficiency.} On one GPU, CryoGaussian achieves
    multiplicative speedups over the CryoSplat
    rasterizer~\citep{chen2026cryosplat} through per-Gaussian gradient
    accumulation and projection scheduling. On eight NVIDIA RTX PRO 6000
    Blackwell GPUs, we measure its further speedup, time per iteration, and
    memory saved (Table~\ref{tab:computational_cost}).}
    \label{tab:rasterizer_speedup}
    \footnotesize
    \setlength{\tabcolsep}{2.7pt}
    \renewcommand{\arraystretch}{1.1}
    \begin{tabular}{@{}lrrrrr@{}}
        \toprule
        Dataset
        & \begin{tabular}[c]{@{}r@{}}Gradient\\accumulation\end{tabular}
        & \begin{tabular}[c]{@{}r@{}}Projection\\scheduling\end{tabular}
        & 8 GPUs
        & \begin{tabular}[c]{@{}r@{}}Time\\(ms)\end{tabular}
        & \begin{tabular}[c]{@{}r@{}}Memory\\saved (\%)\end{tabular} \\
        \midrule
        IgG-1D & 2.39$\times$ & 7.69$\times$ & 5.41$\times$ & 17.31 & 7.98 \\
        IgG-RL & 2.27$\times$ & 7.96$\times$ & 5.79$\times$ & 34.33 & 9.56 \\
        Ribosembly & 2.46$\times$ & 11.37$\times$ & 6.48$\times$ & 73.89 & 8.97 \\
        Tomotwin-100 & 2.62$\times$ & 10.23$\times$ & 5.14$\times$ & 72.90 & 9.74 \\
        Spike-MD & 2.38$\times$ & 6.68$\times$ & 6.16$\times$ & 216.49 & 10.17 \\
        \addlinespace[2pt]
        EMPIAR-10076 & 2.35$\times$ & 5.43$\times$ & 6.18$\times$ & 520.19 & 9.63 \\
        EMPIAR-10180 & 2.64$\times$ & 8.51$\times$ & 5.59$\times$ & 376.64 & 9.95 \\
        \bottomrule
    \end{tabular}
\end{table}

Each particle image is a noisy projection of an unknown density modulated by
the contrast transfer function (CTF), and the structural state underlying
each particle image is also unknown. Signal-to-noise ratios (SNRs) as low as
0.01 obscure fine
features~\citep{bendory2020single}, while overfitting can introduce misleading
density~\citep{scheres2012overfitting}. Discrete
classification can miss rare, transient states or average over continuous
variation~\citep{rabuckgibbons2022mining,punjani20233dflex}.
Density-preserving deformation models including 3DFlex capture motion but are
ill-suited to compositional changes~\citep{punjani20233dflex}. Other
continuous representations accommodate compositional and conformational
heterogeneity~\citep{zhong2021cryodrgn,gilles2025recovar,luo2023opusdsd}, yet
resolving heterogeneity alone does not ensure faithful per-state
reconstruction: even with identical latents, reconstruction procedures
recover different structural detail~\citep{gilles2025recovar}. Accurate
heterogeneity estimation and high-quality per-state reconstruction pose
distinct challenges.

Gaussian mixtures provide physically motivated representations of molecular
density and structural
variation~\citep{doyle1968relativistic,peng1996robust,chen2021e2gmm}, and 3D
Gaussian Splatting (3DGS) makes explicit Gaussians computationally attractive
through differentiable rasterization~\citep{kerbl2023gaussian}. Cryo-EM,
however, records orthographic line integrals modulated by the CTF, so
projected Gaussians add without occlusion and
require a rasterizer adapted to cryo-EM image
formation~\citep{singer2020computational}. CryoSplat adapts Gaussian splatting
to this projection model for homogeneous \mbox{reconstruction}, the special case of
a single structural state, from randomly
initialized Gaussians~\citep{chen2026cryosplat}, whereas Gaussian-based
heterogeneity methods such as e2gmm, DynaMight, CryoSPIRE, and GaussianEM
still require initialization from a reference map or atomic
structure~\citep{chen2021e2gmm,schwab2024dynamight,shekarforoush2025cryospire,he2025gaussianem}.
These limitations motivate us to employ Gaussian splatting, free of
structural priors and grounded in the physics of cryo-EM image formation,
to achieve both accurate
heterogeneity estimation and high-quality per-state reconstruction.

We introduce CryoGaussian, a Gaussian splatting method for heterogeneous
cryo-EM reconstruction that fits an explicit Gaussian density representation
directly to particle images. The Gaussians are initialized randomly, without prior structural
information, with their number derived from the
estimated spatial support of the molecular density and their minimum size bounded by
the Nyquist frequency of the particle images. To estimate compositional and
conformational heterogeneity, we learn particle-image and per-Gaussian latents
that jointly modulate the base position, amplitude, and scale of each
Gaussian. To reconstruct each structural state, we fit the
latent-conditioned Gaussians at its inferred latent.

\noindent\textbf{Main contributions.}
(i) We introduce CryoGaussian, which fits an explicit Gaussian density
representation directly to particle images from random initialization under
data-derived constraints.
(ii) We model both compositional and conformational heterogeneity with
latent-conditioned Gaussians and reconstruct each structural state from the
inferred latents.
(iii) We optimize differentiable rasterization for additive cryo-EM
projection, whose linearity permits per-Gaussian gradient accumulation and
projection scheduling, and distribute training, reducing end-to-end
training time by 12.8--28.0$\times$ on one GPU and a further
5.1--6.5$\times$ on eight GPUs (Table~\ref{tab:rasterizer_speedup}).
(iv) On CryoBench, we demonstrate accurate heterogeneity estimation and
state-of-the-art reconstruction accuracy for both compositional and
conformational heterogeneity; on experimental data, we resolve a rare
assembly intermediate of the 50S ribosomal subunit (class C4, 1.3\% of
particles) and continuous motion of the SF3b and helicase subcomplexes of the
pre-catalytic spliceosome.

\section{Methods}
\label{sec:methods}

CryoGaussian performs heterogeneous reconstruction from $N$ particle images
$\{Y_i\}_{i=1}^{N}$ with pixel size $a$ and dimensions $H\times W$. As is standard in heterogeneous
reconstruction~\citep{zhong2021cryodrgn,punjani20233dflex,gilles2025recovar,jeon2024cryobench},
the orientations $R_i\in\mathrm{SO}(3)$, in-plane translations
$\mathbf t_i$, and CTFs $C_i$ of the particle images are provided as inputs
and held fixed. Figure~\ref{fig:main_flowchart} gives an overview.

\subsection{Gaussian density representation and cryo-EM image formation}
\label{sec:gaussian_projection}

The biomolecular density for a particle-image latent $\mathbf z$ is
\begin{equation}
  \rho_{\Theta}(\mathbf x;\mathbf z)
  =\sum_{g=1}^{G}p_g(\mathbf z)
    \mathcal N_3\!\left(\mathbf x;
      \boldsymbol\mu_g(\mathbf z),\boldsymbol\Sigma_g(\mathbf z)\right),
  \label{eq:gaussian_density}
\end{equation}
where $G$ is the number of Gaussians and $\mathcal N_3$ is a normalized
three-dimensional Gaussian. Each Gaussian has a position
$\boldsymbol\mu_g(\mathbf z)\in\mathbb R^3$, covariance
$\boldsymbol\Sigma_g(\mathbf z)\succ0$, and amplitude \mbox{$p_g(\mathbf z)>0$};
$\Theta$ comprises the Gaussian parameters, per-Gaussian
latents, and decoder weights
(Section~\ref{sec:heterogeneous_learning}).

Under the weak-phase-object and projection approximations, cryo-EM image formation
comprises line integration along the electron beam followed by CTF
modulation~\citep{singer2020computational}. With $P=[I_2\;\mathbf0]$
selecting the physical image-plane coordinates $\mathbf v$,
the line integral of the rotated density is analytic:
\begin{equation}
\begin{aligned}
  &{}[\mathcal P_{R_i}\rho_{\Theta}](\mathbf v;\mathbf z_i)
  =\sum_{g=1}^{G}p_g(\mathbf z_i)
    \mathcal N_2\!\Bigl(\mathbf v;
       PR_i\boldsymbol\mu_g(\mathbf z_i), \\
  &\hspace{5em}PR_i\boldsymbol\Sigma_g(\mathbf z_i)R_i^{\mathsf T}P^{\mathsf T}\Bigr),
\end{aligned}
\label{eq:analytic_projection}
\end{equation}
where $\mathcal N_2$ is a normalized two-dimensional Gaussian. We evaluate these
two-dimensional Gaussians on the image grid with spacing $a$ and apply CTF modulation:
\begin{equation}
  \widehat Y_i^{\mathrm{pre}}(\mathbf v)
  =\kappa[\mathcal P_{R_i}\rho_{\Theta}](\mathbf v;\mathbf z_i),
  \qquad
  \widehat Y_i
  =\mathcal F_2^{-1}\!\left[C_i\,\mathcal F_2[\widehat Y_i^{\mathrm{pre}}]\right].
  \label{eq:ctf_projection}
\end{equation}
Here $\mathcal F_2$ is the discrete Fourier transform and
$\kappa$ a fixed conversion from projected density to image
intensity. Gaussian contributions add independently of depth order, so our
rasterizer accumulates them additively instead of the
front-to-back alpha compositing of 3DGS~\citep{kerbl2023gaussian}, as does
CryoSplat~\citep{chen2026cryosplat}.

Applying the supplied in-plane translations through the Fourier shift theorem
gives aligned targets $Y_i^{\mathrm{aligned}}$. We minimize the image-domain mean
squared error, as in neural and Gaussian-based cryo-EM
methods~\citep{zhong2021cryodrgn,shekarforoush2025cryospire,chen2026cryosplat}:
\begin{equation}
  \mathcal L(\Theta,Z)
  =\frac{1}{NHW}\sum_{i=1}^{N}\sum_{\mathbf v}
    \left[\widehat Y_i(\mathbf v;\Theta,\mathbf z_i)
          -Y_i^{\mathrm{aligned}}(\mathbf v)\right]^2,
\label{eq:image_objective}
\end{equation}
where $Z=\{\mathbf z_i\}_{i=1}^{N}$ and the inner sum runs over the image
grid, subject to the constraints of
Section~\ref{sec:gaussian_capacity}.

\begin{figure*}[t]
    \centering
    \includegraphics[width=1\linewidth]{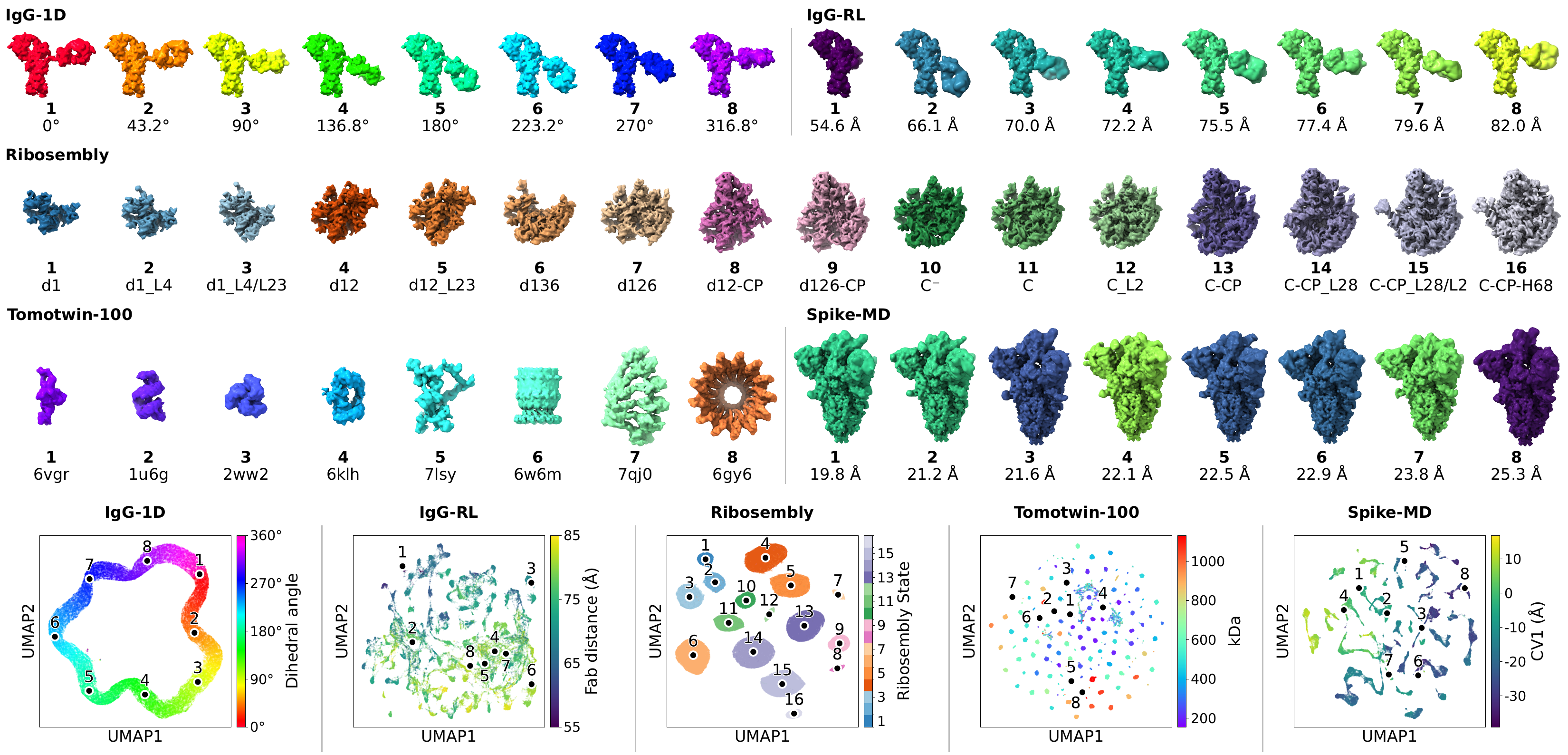}
    \caption{\textbf{Heterogeneous reconstruction on CryoBench.}
Representative reconstructed densities (top)
and UMAP projections~\citep{mcinnes2018umap} of inferred latents (bottom) are linked by matching
numbers. Colors indicate the ground-truth IgG-1D dihedral angle, IgG-RL Fab
distance, Ribosembly assembly state, Tomotwin-100 molecular mass, and
Spike-MD collective variable CV1. Spike-MD labels indicate receptor-binding domain
opening. All 16 Ribosembly states are shown, named as in
Qin et~al.~\citep{qin2023assembly}.
Appendix~\ref{app:additional_cryobench} shows all evaluated reconstructions
with reference densities and FSC curves
(Figures~\ref{fig:igg1d_reconstruction}--\ref{fig:spikemd_reconstruction}),
densities of latent clusters
(Figure~\ref{fig:cryobench_cluster_densities}), and latent metrics
(Figures~\ref{fig:heterogeneity_latent_metrics}--\ref{fig:ctf_information_imbalance}).}
    \label{fig:cryobench_overview}
\end{figure*}

\begin{table*}[t]
    \centering
    \caption{\textbf{Per-conformation FSC AUC on CryoBench.}
Mean (standard deviation) and median across evaluated reference states, with
background masking (Appendix~\ref{app:reference_fsc}). Higher is better. Bold and underline mark the best and
second best per column, including ties at the displayed precision. Conf.,
Comp., and All average the means over the conformational (IgG-1D, IgG-RL,
Spike-MD), the compositional (Ribosembly, Tomotwin-100), and all five
datasets, for methods with values on every dataset, so that
averages are comparable across methods. For CryoSPIRE$^{*}$,
values are taken directly from its publication
(Appendix~\ref{app:baseline_configurations}) because we were
unable to find a publicly available implementation. CryoHype and CryoKRAQEN
have published per-image values only
(Table~\ref{tab:cryobench_per_image_fsc}).
Dashes mark results not included.}
    \label{tab:cryobench_per_conformation_fsc}
    \setlength{\tabcolsep}{3pt}
    \setlength{\aboverulesep}{0pt}
    \setlength{\belowrulesep}{0pt}
    \renewcommand{\arraystretch}{1.3}
    \resizebox{\linewidth}{!}{%
        \begin{tabular}{@{}l*{5}{cc}ccc@{}}
            \toprule
            Method
            & \multicolumn{2}{c}{IgG-1D}
            & \multicolumn{2}{c}{IgG-RL}
            & \multicolumn{2}{c}{Ribosembly}
            & \multicolumn{2}{c}{Tomotwin-100}
            & \multicolumn{2}{c}{Spike-MD}
            & \multicolumn{3}{c}{Average of means} \\
            \cmidrule(lr){2-3}
            \cmidrule(lr){4-5}
            \cmidrule(lr){6-7}
            \cmidrule(lr){8-9}
            \cmidrule(lr){10-11}
            \cmidrule(lr){12-14}
            & Mean (std) & Median
            & Mean (std) & Median
            & Mean (std) & Median
            & Mean (std) & Median
            & Mean (std) & Median
            & Conf. & Comp. & All \\
            \midrule
            CryoDRGN
            & 0.366 (0.003) & 0.366
            & 0.349 (0.008) & 0.348
            & 0.415 (0.019) & 0.415
            & \underline{0.321 (0.034)} & 0.322
            & 0.340 (0.009) & 0.340
            & 0.352 & \underline{0.368} & 0.358 \\
            CryoDRGN-AI-fixed
            & 0.366 (0.001) & 0.366
            & 0.355 (0.007) & 0.354
            & 0.372 (0.032) & 0.374
            & 0.206 (0.041) & 0.212
            & 0.301 (0.012) & 0.304
            & 0.341 & 0.289 & 0.320 \\
            OPUS-DSD
            & 0.340 (0.006) & 0.341
            & 0.346 (0.029) & 0.349
            & 0.372 (0.046) & 0.382
            & 0.256 (0.038) & 0.266
            & 0.228 (0.030) & 0.242
            & 0.305 & 0.314 & 0.308 \\
            3DFlex
            & 0.336 (0.002) & 0.336
            & 0.339 (0.007) & 0.339
            & -- & --
            & -- & --
            & 0.304 (0.010) & 0.306
            & -- & -- & -- \\
            3DVA
            & 0.351 (0.003) & 0.351
            & 0.341 (0.006) & 0.341
            & 0.375 (0.038) & 0.372
            & 0.095 (0.040) & 0.083
            & 0.324 (0.010) & 0.325
            & 0.339 & 0.235 & 0.297 \\
            RECOVAR
            & 0.391 (0.001) & 0.391
            & 0.372 (0.008) & 0.371
            & \underline{0.430 (0.016)} & \textbf{0.432}
            & 0.306 (0.093) & \underline{0.330}
            & \underline{0.363 (0.011)} & \underline{0.366}
            & \underline{0.375} & \underline{0.368} & \underline{0.372} \\
            3D Class
            & 0.297 (0.019) & 0.291
            & 0.309 (0.010) & 0.307
            & 0.289 (0.081) & 0.288
            & 0.046 (0.026) & 0.037
            & 0.307 (0.023) & 0.308
            & 0.304 & 0.168 & 0.250 \\
            CryoSPIRE$^{*}$
            & \textbf{0.402 (0.002)} & \textbf{0.402}
            & \underline{0.386 (0.014)} & \textbf{0.389}
            & 0.427 (0.014) & \underline{0.424}
            & -- & --
            & -- & --
            & -- & -- & -- \\
            SOLVAR
            & \underline{0.392 (0.001)} & \underline{0.392}
            & 0.377 (0.007) & 0.376
            & \textbf{0.432 (0.015)} & \textbf{0.432}
            & 0.298 (0.095) & 0.322
            & 0.355 (0.012) & 0.356
            & \underline{0.375} & 0.365 & 0.371 \\
            \midrule
            \rowcolor{gray!15}
            CryoGaussian (Ours)
            & \textbf{0.402 (0.001)} & \textbf{0.402}
            & \textbf{0.387 (0.010)} & \underline{0.387}
            & \textbf{0.432 (0.016)} & \textbf{0.432}
            & \textbf{0.336 (0.034)} & \textbf{0.343}
            & \textbf{0.365 (0.012)} & \textbf{0.367}
            & \textbf{0.385} & \textbf{0.384} & \textbf{0.384} \\
            \bottomrule
        \end{tabular}%
    }
\end{table*}

\begin{table*}[t]
    \centering
    \caption{\textbf{Per-image FSC AUC on CryoBench.}
Mean (standard deviation) and median across evaluated particle images, with
background masking (Appendix~\ref{app:reference_fsc}); notation as in
Table~\ref{tab:cryobench_per_conformation_fsc}. For methods marked $^{*}$ and
$^{**}$, values are taken directly from their publications
(Appendix~\ref{app:baseline_configurations}), $^{*}$ because we were
unable to find a publicly available implementation and $^{**}$ because we were unable
to find implementations of the mechanisms of CryoHype for the other datasets.
Tables~\ref{tab:seed_variability} and~\ref{tab:cryobench_masking} report
seed variability and unmasked FSC AUC.}
    \label{tab:cryobench_per_image_fsc}
    \setlength{\tabcolsep}{3pt}
    \setlength{\aboverulesep}{0pt}
    \setlength{\belowrulesep}{0pt}
    \renewcommand{\arraystretch}{1.3}
    \resizebox{\linewidth}{!}{%
        \begin{tabular}{@{}l*{5}{cc}ccc@{}}
            \toprule
            Method
            & \multicolumn{2}{c}{IgG-1D}
            & \multicolumn{2}{c}{IgG-RL}
            & \multicolumn{2}{c}{Ribosembly}
            & \multicolumn{2}{c}{Tomotwin-100}
            & \multicolumn{2}{c}{Spike-MD}
            & \multicolumn{3}{c}{Average of means} \\
            \cmidrule(lr){2-3}
            \cmidrule(lr){4-5}
            \cmidrule(lr){6-7}
            \cmidrule(lr){8-9}
            \cmidrule(lr){10-11}
            \cmidrule(lr){12-14}
            & Mean (std) & Median
            & Mean (std) & Median
            & Mean (std) & Median
            & Mean (std) & Median
            & Mean (std) & Median
            & Conf. & Comp. & All \\
            \midrule
            CryoDRGN
            & 0.351 (0.028) & 0.356
            & 0.331 (0.016) & 0.333
            & 0.412 (0.023) & 0.415
            & 0.316 (0.046) & 0.321
            & 0.340 (0.009) & 0.340
            & 0.341 & 0.364 & 0.350 \\
            CryoDRGN-AI-fixed
            & 0.364 (0.002) & 0.364
            & 0.348 (0.012) & 0.350
            & 0.372 (0.032) & 0.375
            & 0.202 (0.044) & 0.207
            & 0.301 (0.012) & 0.303
            & 0.338 & 0.287 & 0.317 \\
            OPUS-DSD
            & 0.335 (0.026) & 0.339
            & 0.343 (0.016) & 0.346
            & 0.362 (0.083) & 0.382
            & 0.237 (0.049) & 0.251
            & 0.229 (0.027) & 0.242
            & 0.302 & 0.300 & 0.301 \\
            3DFlex
            & 0.335 (0.003) & 0.335
            & 0.337 (0.007) & 0.337
            & -- & --
            & -- & --
            & 0.304 (0.011) & 0.306
            & -- & -- & -- \\
            3DVA
            & 0.349 (0.004) & 0.350
            & 0.333 (0.014) & 0.335
            & 0.375 (0.038) & 0.375
            & 0.088 (0.040) & 0.077
            & 0.324 (0.010) & 0.323
            & 0.335 & 0.232 & 0.294 \\
            RECOVAR
            & 0.386 (0.005) & 0.388
            & 0.363 (0.011) & 0.363
            & 0.429 (0.018) & \textbf{0.432}
            & 0.258 (0.109) & 0.254
            & \underline{0.362 (0.011)} & \underline{0.365}
            & \underline{0.370} & 0.344 & 0.360 \\
            3D Class
            & 0.297 (0.019) & 0.291
            & 0.309 (0.010) & 0.307
            & 0.289 (0.081) & 0.288
            & 0.046 (0.026) & 0.037
            & 0.307 (0.023) & 0.308
            & 0.304 & 0.168 & 0.250 \\
            CryoSPIRE$^{*}$
            & \underline{0.396 (0.013)} & \textbf{0.400}
            & \textbf{0.375 (0.020)} & \textbf{0.391}
            & 0.422 (0.015) & 0.421
            & -- & --
            & -- & --
            & -- & -- & -- \\
            CryoHype$^{**}$
            & -- & --
            & -- & --
            & -- & --
            & \textbf{0.346 (0.033)} & \textbf{0.353}
            & -- & --
            & -- & -- & -- \\
            SOLVAR
            & 0.388 (0.003) & 0.389
            & \underline{0.369 (0.009)} & 0.368
            & \underline{0.431 (0.017)} & \textbf{0.432}
            & 0.246 (0.104) & 0.238
            & 0.354 (0.012) & 0.355
            & \underline{0.370} & 0.339 & 0.358 \\
            CryoKRAQEN$^{*}$
            & 0.375 (0.004) & 0.376
            & 0.352 (0.010) & 0.352
            & 0.418 (0.019) & \underline{0.422}
            & \underline{0.335 (0.021)} & 0.337
            & 0.328 (0.010) & 0.329
            & 0.352 & \underline{0.377} & \underline{0.362} \\
            \midrule
            \rowcolor{gray!15}
            CryoGaussian (Ours)
            & \textbf{0.397 (0.004)} & \underline{0.398}
            & 0.366 (0.020) & \underline{0.370}
            & \textbf{0.432 (0.014)} & \textbf{0.432}
            & \underline{0.335 (0.035)} & \underline{0.341}
            & \textbf{0.364 (0.012)} & \textbf{0.366}
            & \textbf{0.376} & \textbf{0.384} & \textbf{0.379} \\
            \bottomrule
        \end{tabular}%
    }
\end{table*}

\begin{figure*}[t]
    \centering
    \includegraphics[width=1\linewidth]{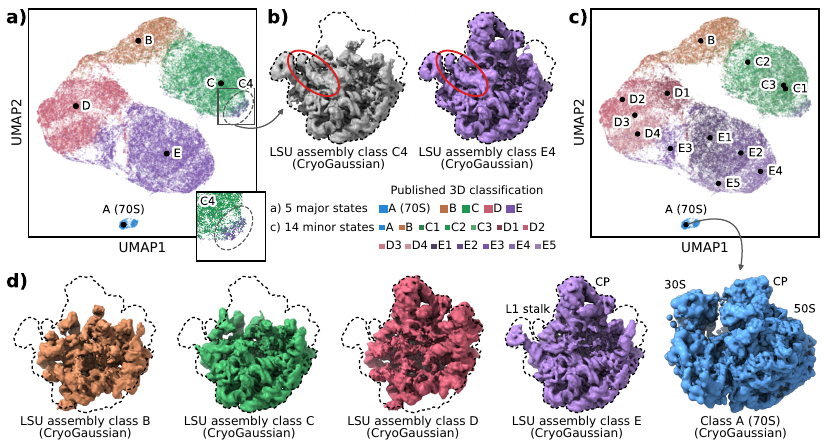}
    \caption{\textbf{Compositional heterogeneity in the assembling 50S ribosomal
subunit (EMPIAR-10076).} \textbf{(a, c)} UMAP projections of the inferred
latents, colored by published major and minor class
assignments~\citep{davis2016assembly};
black points mark class centers. The inset locates the 1,113 particle images
of class C4~\citep{zhong2021cryodrgn}. \textbf{(b)} C4 and E4 densities.
\textbf{(d)} Major assembly classes B--E and the 70S class A.
Dashed lines trace the outline of the mature 50S subunit, and red ellipses mark rRNA
helix 68 (H68). CP, central protuberance; LSU, large subunit.
Appendix~\ref{app:additional_experimental_heterogeneity} shows the twelve
minor classes C1--C3, D1--D4, and E1--E5 with map-to-map FSC against the
deposited reconstructions (Figure~\ref{fig:empiar10076_supplied_states}) and
the CryoSPARC refinement of C4
(Figure~\ref{fig:empiar10076_c4_comparison}).
Appendix~\ref{app:particle_selection} presents particle selection by
CryoGaussian
(Figures~\ref{fig:empiar10076_filtering}--\ref{fig:empiar10076_cross_comparison}).}
    \label{fig:empiar10076_main}
\end{figure*}

\begin{figure*}[t]
    \centering
    \includegraphics[width=0.92\linewidth]{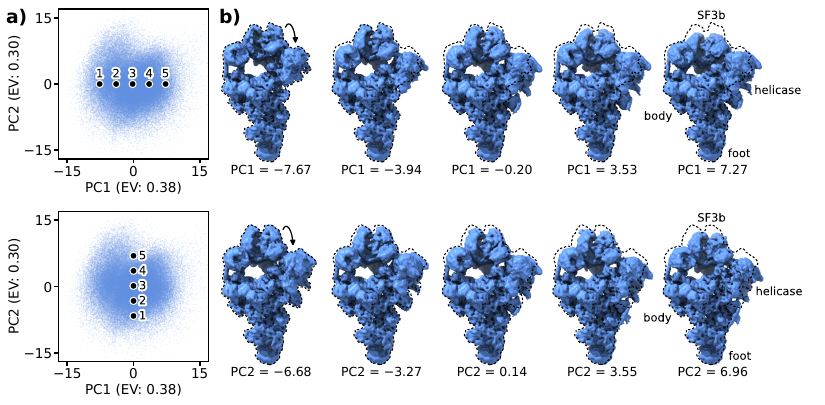}
    \caption{\textbf{Conformational variation in the pre-catalytic spliceosome
(EMPIAR-10180).} \textbf{(a)} Principal-component projections of inferred
latents, with five numbered positions along PC1 (top) and PC2 (bottom).
EV denotes the fraction of latent variance explained by each component.
\textbf{(b)} Densities reconstructed at positions 1--5, ordered from left
to right in each row. Dashed outlines reproduce each row's first
density; arrows indicate the direction of conformational change.
Appendix~\ref{app:additional_experimental_heterogeneity} analyzes the full
particle stack (Figure~\ref{fig:empiar10180_full_dataset}).}
    \label{fig:empiar10180_main}
\end{figure*}

\subsection{Data-derived constraints}
\label{sec:gaussian_capacity}

The pixel size $a$ defines the Nyquist frequency
$1/(2a)$~\citep{shannon1949communication}. We use the corresponding sampling
scale $d=2a$ for the minimum size and number of Gaussians
(Section~\ref{sec:ablations_robustness}).

\noindent\textbf{Minimum Gaussian size.}
Inspired by the sampling-frequency constraint on Gaussian size in
Mip-Splatting~\citep{yu2024mipsplatting}, we bound the full width at half
maximum (FWHM) of each Gaussian below by the sampling scale $d$, below
which Gaussians add no resolvable detail and fit noise.
For every effective principal-axis scale $\sigma$ (standard deviation),
\begin{equation}
  \mathrm{FWHM}=2\sqrt{2\ln2}\,\sigma\geq d,
  \qquad d=2a.
  \label{eq:fwhm_constraint}
\end{equation}

\noindent\textbf{Number of Gaussians.}
Gaussian-mixture cryo-EM methods set the number of Gaussians before
optimization~\citep{chen2024improving,schwab2024dynamight,shekarforoush2025cryospire}.
We determine $G$ from the pixel size, which bounds the frequency band of the
density, and from the spatial support of the density, estimated from a
CTF-weighted backprojection $\rho_{\mathrm{BP}}$ of the particle images. For a
relative density threshold $\tau\in(0,1)$, the estimated spatial support
$\Omega_\tau$ is the union of the $n_\tau$ voxel cells of side $a$ in
which the nonnegative part of $\rho_{\mathrm{BP}}$ is at least
$\tau$ times its maximum, including disconnected regions
(Appendix~\ref{app:capacity_backprojection}), and has volume
$V_\tau=n_\tau a^3$.

The space--bandwidth product characterizes the approximate effective dimension of
signals concentrated within a spatial region and a frequency
band~\citep{slepian1964multidimensional,simons2011spatiospectral}, and in cryo-EM it has
been used to set the effective dimension of density
representations~\citep{katsevich2015covariance}. We use the rotationally symmetric sampling band
$B_d=\{\boldsymbol\xi:\|\boldsymbol\xi\|_2\leq1/d\}$, with frequencies in cycles
per unit length, and set $G$ to the space--bandwidth product of the
estimated support (Appendix~\ref{app:space_bandwidth}),
\begin{equation}
  \mathcal S_\tau=V_\tau|B_d|
  =\frac{4\pi}{3}\frac{V_\tau}{d^3},
  \qquad
  G=\mathcal R(\mathcal S_\tau)
   =\mathcal R\!\left(\frac{\pi}{6}n_\tau\right),
  \label{eq:gaussian_count}
\end{equation}
where $\mathcal R$ rounds to an integer. $G$ is the effective dimension of
densities confined to the estimated support at the sampling scale $d$.
We use $\tau=0.02$, which
approximates the relative boundary density of the 95-percent mass ellipsoid
of a three-dimensional Gaussian (Appendix~\ref{app:density_threshold}).

The backprojection enters CryoGaussian only through the volume of the
estimated support. Gaussian positions are initialized randomly and are not
restricted to this support, so no structure of the imaged biomolecules is
encoded before optimization, and their structural states, including rare ones
(Section~\ref{sec:empiar10076}), are inferred directly from the particle
images. Inspired by 3DGS-MCMC~\citep{kheradmand2024mcmc}, we fix $G$ and
periodically relocate low-amplitude Gaussians
(Appendix~\ref{app:gaussian_relocation}).

\subsection{Heterogeneous reconstruction}
\label{sec:heterogeneous_learning}

\noindent\textbf{Gaussian attribute decoding.}
Each particle image has a trainable particle-image latent
$\mathbf z_i\in\mathbb R^{d_z}$, and each Gaussian has a trainable
per-Gaussian latent $\mathbf f_g\in\mathbb R^{d_f}$.
Separate multilayer perceptrons (MLPs) decode position, amplitude, and
scale modulation from the concatenated latent
$\mathbf h_g(\mathbf z)=[\mathbf f_g;\mathbf z]$. The position and amplitude
decoders produce
\begin{equation}
\begin{aligned}
  \boldsymbol\mu_g(\mathbf z)
  &=\boldsymbol\mu_g^{\mathrm{base}}+M_{\mu}(\mathbf h_g(\mathbf z)), \\
  p_g(\mathbf z)
  &=p_g^{\mathrm{base}}\,
    \operatorname{sigmoid}\!\left(M_p(\mathbf h_g(\mathbf z))\right),
\end{aligned}
\label{eq:state_position_amplitude}
\end{equation}
where $\boldsymbol\mu_g^{\mathrm{base}}$ and $p_g^{\mathrm{base}}$ are the learned
base position and amplitude. The scale decoder modulates the three
principal-axis scales $\sigma_{gk}(\mathbf z)$, $k=1,2,3$, of the effective
covariance $\boldsymbol\Sigma_g(\mathbf z)
=Q_g\operatorname{diag}(\sigma_{g1}^2,\sigma_{g2}^2,\sigma_{g3}^2)Q_g^{\mathsf T}$,
each subject to the FWHM constraint in
Equation~\eqref{eq:fwhm_constraint}, where $Q_g\in\mathrm{SO}(3)$ is the
learned Gaussian rotation (Appendix~\ref{app:gaussian_parameterization}).

Particle-image latents parameterize the
structural state of each particle image, and per-Gaussian latents determine
how each Gaussian responds to that state. Each decoder is shared by all
Gaussians and all particle images, so that, for a given state, the response
of a Gaussian depends only on its per-Gaussian latent. The per-Gaussian
latents are
optimized jointly with the decoders, which allows Gaussians that vary
together to share similar latents without a predefined grouping.
Together, the three modulations allow each Gaussian to move,
change shape, and appear or disappear across states, so that compositional
and conformational variation are represented within one set of Gaussians.

\noindent\textbf{Heterogeneity estimation and per-state reconstruction.}
Heterogeneous reconstruction comprises two problems under the objective in
Equation~\eqref{eq:image_objective}, each evaluated in
Section~\ref{sec:cryobench_evaluation}. Heterogeneity estimation,
$\min_{\Theta_1,Z}\mathcal L(\Theta_1,Z)$, yields the structural state
$Z^\star$ of each particle image by optimizing the latents $Z$ jointly with
the parameters $\Theta_1$ of randomly initialized Gaussians, which are
fitted while the latents change. Per-state
reconstruction, $\min_{\Theta_2}\mathcal L(\Theta_2,Z^\star)$, yields the
density of each state by fitting randomly initialized latent-conditioned
Gaussians, whose parameters $\Theta_2$ include their per-Gaussian latents
and decoders, at the inferred latents $Z^\star$ under the same data-derived
constraints. For a selected latent $\mathbf z$, their decoders define the
density $\rho_{\Theta_2}(\mathbf x;\mathbf z)$, evaluated on a
three-dimensional grid without additional optimization.
Section~\ref{sec:ablations_robustness} compares this per-state
reconstruction with evaluating or continuing to optimize the Gaussians of
heterogeneity estimation at the inferred latents.

\subsection{Efficient cryo-EM Gaussian splatting}
\label{sec:efficient_optimization}

Additive cryo-EM projection has no depth ordering or early termination.
Our differentiable CUDA rasterizer accumulates gradients per Gaussian
instead of pixel-wise atomic scatter, as in Taming
3DGS~\citep{mallick2024taming} but without compositing state, and its
projection scheduling partitions long tile lists into
independent partial projections, the additive counterpart of tile load
balancing in 3DGS~\citep{gui2024balanced,wang2024adr}. Its backward pass
recomputes the projection geometry of each
Gaussian instead of retaining the forward workspaces, which lowers peak
memory. Training is distributed across GPUs, and the image-formation model
is unchanged
(Table~\ref{tab:rasterizer_speedup}; Appendix~\ref{app:renderer_optimization}).

\section{Experiments}
\label{sec:experiments}

\noindent\textbf{Datasets and baselines.}
We quantitatively compare CryoGaussian with existing methods on
CryoBench~\citep{jeon2024cryobench}, whose five synthetic datasets provide
ground-truth structures and states for per-state reconstruction
and heterogeneity estimation. IgG-1D, IgG-RL, and Spike-MD assess
conformational heterogeneity through one-dimensional Fab motion, randomized
linker conformations, and a molecular-dynamics ensemble, respectively, and
Ribosembly and Tomotwin-100 assess compositional heterogeneity through 16
unequally populated assembly states and a mixture of 100 distinct complexes,
respectively (Table~\ref{tab:cryobench_data}). We evaluate CryoGaussian on
experimental cryo-EM data, the assembling 50S ribosomal subunit
(EMPIAR-10076)~\citep{davis2016assembly} and the pre-catalytic spliceosome
(EMPIAR-10180)~\citep{plaschka2017spliceosome}, to assess its ability to
resolve compositional and conformational heterogeneity, respectively
(Table~\ref{tab:experimental_data}). Data preparation, baseline
configurations, the compared methods, and evaluation protocols are detailed
in Appendices~\ref{app:experimental_data},
\ref{app:baseline_configurations}, \ref{app:extended_related_work},
and~\ref{app:experimental_evaluation}, respectively.

\noindent\textbf{Implementation.}
Heterogeneity estimation and per-state reconstruction each use 30,000
iterations of Adam~\citep{kingma2015adam}, with particle-image and
per-Gaussian latent dimensions $d_z=4$ and $d_f=24$ selected by ablation
(Section~\ref{sec:ablations_robustness}). All datasets use the same
dimensions, relative density threshold, and base learning rates.
Appendix~\ref{app:gaussian_parameterization} details the Gaussian
parameterization, Appendix~\ref{app:training_ablation_settings} the
training settings, and Appendix~\ref{app:computational_cost} the training
time and peak GPU memory (Table~\ref{tab:computational_cost}).

\subsection{CryoBench evaluation}
\label{sec:cryobench_evaluation}

\noindent\textbf{Per-state reconstruction.}
Table~\ref{tab:cryobench_per_conformation_fsc} reports the area under the
Fourier shell correlation curve (FSC AUC) per conformation and
Table~\ref{tab:cryobench_per_image_fsc} the FSC AUC per image,
following the CryoBench protocols
(Appendix~\ref{app:reference_fsc}). CryoGaussian attains the highest mean
FSC AUC per conformation on all five datasets and per image on IgG-1D,
Ribosembly, and Spike-MD. We also average the
means over the conformational, compositional, and all five
datasets. Under the
per-image protocol, the ranking of the compared methods depends on the type
of heterogeneity: RECOVAR and SOLVAR~\citep{yadgar2026solvar} rank second
on the conformational but fourth and fifth, respectively, on the
compositional datasets, whereas CryoKRAQEN~\citep{gao2026cryokraqen} ranks second on the
compositional but fourth on the conformational datasets.
CryoGaussian ranks first in all three averages under both protocols,
attaining state-of-the-art reconstruction accuracy on
both compositional and conformational heterogeneity. The average of
CryoGaussian over all five datasets exceeds that of the next-ranked method
by 0.012 per conformation and 0.017 per image. Across 10 training
seeds, the standard deviation of the mean FSC AUC of CryoGaussian is at most
0.0012 on each dataset under both protocols
(Table~\ref{tab:seed_variability}), an order of magnitude below these
margins.
Figure~\ref{fig:cryobench_overview} shows reconstructed
densities and inferred latents.
Figures~\ref{fig:igg1d_reconstruction}--\ref{fig:spikemd_reconstruction}
show reconstruction--reference comparisons and FSC curves, and
Figure~\ref{fig:cryobench_cluster_densities} shows densities selected by
latent clustering (Appendix~\ref{app:additional_cryobench}).

\noindent\textbf{Heterogeneity estimation.}
We evaluate the heterogeneity learned by CryoGaussian with the CryoBench
metrics (Appendix~\ref{app:latent_evaluation}). On all five datasets,
CryoGaussian attains the highest neighborhood similarity to the
ground-truth heterogeneity embedding at the largest evaluated radius,
and the information imbalance from its inferred latents to
this embedding lies within 0.04--0.20, where 0 and 1
indicate equivalent and orthogonal information
(Figure~\ref{fig:heterogeneity_latent_metrics}). On Ribosembly and
Tomotwin-100, $k$-means clustering of the inferred latents into the
ground-truth number of states agrees with the state labels
with ARI of 0.974 and 0.823 and AMI of 0.982 and 0.926,
ranking first on Ribosembly and second on
Tomotwin-100; CryoGaussian is the only method in the top two on both datasets
(Table~\ref{tab:clustering_accuracy}). CryoBench further assesses whether latents capture non-structural factors
such as pose and CTF~\citep{jeon2024cryobench}. CryoGaussian's latents are
orthogonal to both: their information imbalance in both directions lies
within 0.99--1.02 with ground-truth rotations and 0.997--1.000 with CTF
parameters on all five datasets
(Figures~\ref{fig:pose_information_imbalance}
and~\ref{fig:ctf_information_imbalance}).

\subsection{Assembling 50S ribosomal subunit}
\label{sec:empiar10076}

\noindent\textbf{Compositional heterogeneity (EMPIAR-10076).}
We analyze the assembling 50S ribosomal subunit from an L17-depleted
sample~\citep{davis2016assembly} using the CryoDRGN-filtered subset of
87,327 particle images~\citep{zhong2021cryodrgn}. The inferred latents,
learned without class labels, distinguish the five major classes and capture
finer variation among the 14 minor classes of the published
classification~\citep{davis2016assembly}, including 70S class A
\mbox{(Figure~\ref{fig:empiar10076_main})}. Map-to-map FSC assesses the agreement
of the reconstructed assembly states with the deposited reconstructions
(Figure~\ref{fig:empiar10076_supplied_states};
Appendices~\ref{app:experimental_heterogeneity}
and~\ref{app:additional_experimental_heterogeneity}).

\noindent\textbf{Rare assembly intermediate C4.}
CryoGaussian resolves a rare assembly intermediate with rRNA helix 68
(H68) docked in the absence of the central protuberance (class C4, 1,113
of 87,327 particle images, 1.3\%)~\citep{zhong2021cryodrgn}
(Figure~\ref{fig:empiar10076_main}(b)). Homogeneous CryoSPARC
refinement of these 1,113 particle images shows
the same two features (Figure~\ref{fig:empiar10076_c4_comparison}(a)).

\noindent\textbf{Particle selection by CryoGaussian.}
CryoGaussian also performs the particle selection itself: trained on all
131,899 particle images without class
labels, it selects 97,651 particle images
(Figure~\ref{fig:empiar10076_filtering};
Appendix~\ref{app:particle_selection}).
The C4 intermediate is also resolved from this CryoGaussian-filtered subset,
and refinement of the 1,019 of the 1,113 C4 particle images retained in
this subset shows the same two features
(Figure~\ref{fig:empiar10076_c4_comparison}(b));
Figure~\ref{fig:empiar10076_filtered_states} shows the assembly states
reconstructed from this subset and their map-to-map FSC.
Figure~\ref{fig:empiar10076_cross_comparison} compares the major assembly
classes B--E reconstructed by CryoDRGN with those reconstructed by
CryoGaussian from each of the two filtered subsets: regions fragmented
in the CryoDRGN densities are
continuous in both CryoGaussian reconstructions. CryoGaussian
accomplishes both particle selection and the reconstruction of a rare
assembly intermediate on experimental data, starting from the unfiltered
particle stack.

\subsection{Pre-catalytic spliceosome}
\label{sec:empiar10180}

\noindent\textbf{Conformational heterogeneity (EMPIAR-10180).}
We analyze the \textit{Saccharomyces cerevisiae} pre-catalytic
spliceosome using the CryoDRGN-filtered subset of 139,720 particle
images~\citep{zhong2021cryodrgn}. Reconstructions along the first two
principal components
of the inferred latents resolve continuous motion of the SF3b and
helicase
subcomplexes~\citep{plaschka2017spliceosome,zhong2021cryodrgn} relative to
the body and foot of the complex (Figure~\ref{fig:empiar10180_main}).

\noindent\textbf{Structural subpopulations.}
Trained on the full stack of 327,489 particle images, CryoGaussian
separates the pre-catalytic spliceosome from structurally distinct
populations. Clustering the inferred latents isolates populations lacking
resolved SF3b density (14,063 particle images), showing additional density
near the U2 core (12,854 particle images), and exhibiting poorly resolved
density (10,052 particle images)~\citep{zhong2021cryodrgn}
(Figure~\ref{fig:empiar10180_full_dataset};
Appendix~\ref{app:experimental_heterogeneity}). CryoGaussian resolves
continuous conformational dynamics and compositionally distinct
subpopulations in one experimental dataset and enables particle
selection from the unfiltered stack.

\subsection{Ablations and robustness}
\label{sec:ablations_robustness}

Unless noted otherwise, the comparisons below are in mean FSC AUC and hold
on all five CryoBench datasets under both protocols; each study is detailed
in Appendix~\ref{app:sensitivity_robustness}.

\noindent\textbf{Number of Gaussians.}
Mean FSC AUC and, on IgG-1D, neighborhood similarity are highest at the
default relative density threshold $\tau$ of 0.02. Mean FSC AUC is
0.002--0.012 lower with more Gaussians at 0.01 and 0.033--0.067 lower with
the fewest Gaussians at 0.10 (Table~\ref{tab:tau_sensitivity};
Figure~\ref{fig:gaussian_count_sensitivity}).

\noindent\textbf{FWHM constraint and initialization.}
Without the FWHM constraint, mean FSC AUC is 0.016--0.030 lower
(Table~\ref{tab:fwhm_ablation}) and neighborhood similarity on IgG-1D is
lower (Figure~\ref{fig:fwhm_ablation}). Initializing Gaussian positions from
a consensus density fitted without latents instead of at random decreases
this similarity from 82.7\% to 55.8\% at SNR 0.01
(Figure~\ref{fig:initialization_sensitivity}).

\noindent\textbf{Per-state reconstruction strategy.}
We examine whether the Gaussians of heterogeneity estimation can replace the
randomly initialized Gaussians that CryoGaussian fits at the inferred latents
for per-state reconstruction. At the same total number of iterations, mean
FSC AUC is 0.008--0.016 lower with heterogeneity estimation alone and
0.005--0.010 lower with per-state reconstruction continued from the Gaussians
of heterogeneity estimation (Table~\ref{tab:reconstruction_strategy}).

\noindent\textbf{Decoders and latent dimensions.}
Among all 15 combinations of the position, amplitude, and scale decoders
and an additional rotation decoder, the combination of the
position, amplitude, and scale decoders attains the highest mean FSC AUC.
Removing the amplitude, position, or scale decoder from this combination
decreases mean FSC AUC by 0.013--0.027, 0.007--0.020, and 0.008--0.015,
respectively, and adding the rotation decoder to any combination decreases
mean FSC AUC (Tables~\ref{tab:decoder_ablation_per_conformation}
and~\ref{tab:decoder_ablation_per_image}). Over particle-image latent
dimensions of 2, 4, 8, and 10 and per-Gaussian latent dimensions of 16, 24,
and 32, mean FSC AUC is highest at 4 and 24
(Table~\ref{tab:latent_dimensions}). Although CryoGaussian differs from
CryoSPIRE~\citep{shekarforoush2025cryospire} in its heterogeneity model,
constraints on the Gaussians, and decoders, these dimensions match the
latent and Gaussian feature dimensions of CryoSPIRE.

\noindent\textbf{Image noise and evaluation masks.}
On IgG-1D, CryoGaussian attains the highest mean per-image FSC AUC among the
compared methods at SNR 0.01 and 0.005
(Figure~\ref{fig:igg1d_robustness}), with reconstructions at SNR 0.005 and
0.001 shown in Figures~\ref{fig:igg1d_snr005} and~\ref{fig:igg1d_snr001}.
Removing the evaluation mask changes mean FSC AUC by at most 0.001 on all
five datasets (Table~\ref{tab:cryobench_masking}).

\section{Related Work}
\label{sec:related_work}

\noindent\textbf{Cryo-EM reconstruction.} RELION, CryoSPARC, and EMAN2 reconstruct
voxelized densities~\citep{scheres2012relion,punjani2017cryosparc,tang2007eman2},
whereas neural fields represent density as a continuous function of Fourier
or real-space
coordinates~\citep{zhong2021cryodrgn,qu2025cryonerf,zhu2024cryonefen,herreros2025hetsiren}.
Empirical Bayesian methods regularize reconstruction using signal and noise
statistics~\citep{scheres2012bayesian}, while Blush and CryoFM
incorporate learned density priors~\citep{kimanius2024blush,zhou2025cryofm}. CryoGaussian uses an
explicit Gaussian density representation for heterogeneous reconstruction
without pretrained density priors.

\noindent\textbf{Heterogeneous reconstruction.} Discrete classification reconstructs
densities for particle classes~\citep{scheres2007disentangling,scheres2012relion}.
Continuous descriptions use linear subspaces in 3DVA, covariance estimation
in RECOVAR and SOLVAR, or latent-conditioned density decoders in CryoDRGN and
OPUS-DSD~\citep{punjani20213dva,gilles2025recovar,yadgar2026solvar,zhong2021cryodrgn,luo2023opusdsd}.
Hydra and CryoHype model mixtures of distinct complexes with neural
fields~\citep{levy2024mixture,gu2026cryohype}, and CryoKRAQEN learns
quantized embeddings~\citep{gao2026cryokraqen}. Motion-based approaches use
Gaussian displacements, rigid-body transformations, or deformation
fields~\citep{nakane2018multibody,punjani20233dflex,schwab2024dynamight,shekarforoush2025cryospire}.
CryoDRGN-AI directly optimizes particle-image latents, whereas cryoSTAR uses
atomic-model constraints to infer latents for per-state
reconstruction~\citep{levy2025cryodrgnai,li2024cryostar}. CryoGaussian infers
compositional and conformational heterogeneity by fitting latent-conditioned
Gaussians directly to particle images, and reconstructs each state
from the inferred latents.

\noindent\textbf{Gaussian mixtures in cryo-EM.} Gaussian mixtures fit density maps
and infer heterogeneity from particle
images~\citep{kawabata2008gmm,kawabata2018gaussian,chen2021e2gmm,chen2023integrating}.
Their initialization and number of Gaussians commonly incorporate structural
information.
DynaMight deforms a Gaussian pseudo-atomic model built from a consensus
reconstruction~\citep{schwab2024dynamight}, e2gmm fits a density map or
sets the number of Gaussians from atom
counts~\citep{chen2021e2gmm,chen2024improving}, and CryoSPIRE is a
hierarchical, part-based motion model initialized from a consensus
reconstruction, whose Gaussians are displaced with their anchors and
modulated in amplitude~\citep{shekarforoush2025cryospire}. GaussianEM samples
its Gaussians from a homogeneous reconstruction and infers their displacement
and density changes with image and Gaussian
encoders~\citep{he2025gaussianem}. CryoGaussian fits a data-derived number of
randomly initialized Gaussians directly to particle images and models
compositional and conformational heterogeneity through directly optimized
particle-image and per-Gaussian latents, without a part-based hierarchy.

\noindent\textbf{Gaussian splatting.} Classical volume splatting projects
reconstruction kernels into image-plane
footprints~\citep{westover1990footprint,zwicker2001ewa}, and 3DGS combines
explicit Gaussians with differentiable rasterization and depth-ordered alpha
compositing~\citep{kerbl2023gaussian}. Adaptations to projection imaging
include X-ray image synthesis and tomographic
reconstruction~\citep{cai2024radiative,zha2024r2gaussian,talegaonkar2025volumetric}
and the homogeneous cryo-EM reconstruction of CryoSplat and
GEM~\citep{chen2026cryosplat,qu2025gem}. Mip-Splatting constrains Gaussian size
by the sampling frequency~\citep{yu2024mipsplatting},
3DGS-MCMC relocates low-opacity Gaussians~\citep{kheradmand2024mcmc}, and
dynamic-scene extensions deform Gaussians over
time~\citep{yang2024deformable,wu20244dgs,bae2024pergaussian}.
CryoGaussian applies CTF-modulated orthographic projection to heterogeneous
reconstruction, with a Nyquist bound on Gaussian size, a data-derived number
of Gaussians, and Gaussian attributes conditioned on the structural state of
each particle image.

\section{Conclusion}
\label{sec:conclusion}

We presented CryoGaussian, a Gaussian splatting method for heterogeneous
cryo-EM reconstruction that fits an explicit Gaussian density representation
directly to particle images from random initialization under data-derived
constraints. CryoGaussian attains
state-of-the-art reconstruction accuracy across compositional and
conformational heterogeneity on CryoBench. On experimental data, it resolves
the rare 50S assembly intermediate C4 and continuous motion of the
pre-catalytic spliceosome, and enables particle selection from unfiltered
particle stacks. These results demonstrate
that CryoGaussian achieves both accurate
heterogeneity estimation and high-quality per-state reconstruction, with the
potential to
support mechanistic studies, structure-based drug and vaccine design, and
protein engineering (Appendix~\ref{app:extended_discussion}).

\clearpage
\bibliographystyle{unsrtcase}
\bibliography{main}

@article{cheng2015primer,
  author = {Cheng, Yifan and Grigorieff, Nikolaus and Penczek, Pawel A. and Walz, Thomas},
  title = {A Primer to Single-Particle Cryo-Electron Microscopy},
  journal = {Cell},
  volume = {161},
  number = {3},
  pages = {438--449},
  year = {2015},
  doi = {10.1016/j.cell.2015.03.050},
}

@article{yip2020atomic,
  title = {Atomic-resolution protein structure determination by {cryo-EM}},
  author = {Yip, Ka Man and Fischer, Niels and Paknia, Elham and Chari, Ashwin and Stark, Holger},
  journal = {Nature},
  volume = {587},
  number = {7832},
  pages = {157--161},
  year = {2020},
  doi = {10.1038/s41586-020-2833-4},
  url = {https://doi.org/10.1038/s41586-020-2833-4}
}

@article{zheng2024supercomplexes,
  author  = {Zheng, Wan and Chai, Pengxin and Zhu, Jiapeng and Zhang, Kai},
  title   = {High-resolution in situ structures of mammalian respiratory
             supercomplexes},
  journal = {Nature},
  year    = {2024},
  volume  = {631},
  number  = {8019},
  pages   = {232--239},
  doi     = {10.1038/s41586-024-07488-9},
}

@article{dong2019proteasome,
  author  = {Dong, Yuanchen and Zhang, Shuwen and Wu, Zhaolong and Li, Xuemei
             and Wang, Wei Li and Zhu, Yanan and Stoilova-McPhie, Svetla
             and Lu, Ying and Finley, Daniel and Mao, Youdong},
  title   = {{Cryo-EM} structures and dynamics of substrate-engaged human
             {26S} proteasome},
  journal = {Nature},
  year    = {2019},
  volume  = {565},
  number  = {7737},
  pages   = {49--55},
  doi     = {10.1038/s41586-018-0736-4},
}

@article{punjani20213dva,
  author = {Punjani, Ali and Fleet, David J.},
  title = {{3D} variability analysis: Resolving continuous flexibility and discrete heterogeneity from single particle {cryo-EM}},
  journal = {Journal of Structural Biology},
  volume = {213},
  number = {2},
  pages = {107702},
  year = {2021},
  doi = {10.1016/j.jsb.2021.107702},
}

@article{punjani20233dflex,
  author = {Punjani, Ali and Fleet, David J.},
  title = {{3DFlex}: determining structure and motion of flexible proteins from {cryo-EM}},
  journal = {Nature Methods},
  volume = {20},
  number = {6},
  pages = {860--870},
  year = {2023},
  doi = {10.1038/s41592-023-01853-8},
}

@article{zhong2021cryodrgn,
  author = {Zhong, Ellen D. and Bepler, Tristan and Berger, Bonnie and Davis, Joseph H.},
  title = {{CryoDRGN}: reconstruction of heterogeneous cryo-{EM} structures using neural networks},
  journal = {Nature Methods},
  volume = {18},
  number = {2},
  pages = {176--185},
  year = {2021},
  doi = {10.1038/s41592-020-01049-4},
}

@article{luo2023opusdsd,
  author = {Luo, Zhenwei and Ni, Fengyun and Wang, Qinghua and Ma, Jianpeng},
  title = {{OPUS-DSD}: deep structural disentanglement for {cryo-EM} single-particle analysis},
  journal = {Nature Methods},
  volume = {20},
  number = {11},
  pages = {1729--1738},
  year = {2023},
  doi = {10.1038/s41592-023-02031-6},
}

@article{gilles2025recovar,
  author = {Gilles, Marc Aur{\`e}le and Singer, Amit},
  title = {{Cryo-EM} heterogeneity analysis using regularized covariance estimation and kernel regression},
  journal = {Proceedings of the National Academy of Sciences},
  volume = {122},
  number = {9},
  pages = {e2419140122},
  year = {2025},
  doi = {10.1073/pnas.2419140122},
}

@article{kerbl2023gaussian,
  author     = {Kerbl, Bernhard and Kopanas, Georgios
                and Leimk{\"u}hler, Thomas and Drettakis, George},
  title      = {{3D} Gaussian Splatting for Real-Time Radiance Field Rendering},
  journal    = {ACM Transactions on Graphics},
  year       = {2023},
  volume     = {42},
  number     = {4},
  articleno  = {139},
  numpages   = {14},
  doi        = {10.1145/3592433},
}

@article{doyle1968relativistic,
  author = {Doyle, P. A. and Turner, P. S.},
  title = {Relativistic {Hartree--Fock} {X-ray} and electron scattering factors},
  journal = {Acta Crystallographica Section A},
  volume = {24},
  number = {3},
  pages = {390--397},
  year = {1968},
  doi = {10.1107/S0567739468000756},
}

@article{peng1996robust,
  author = {Peng, L. M. and Ren, G. and Dudarev, S. L. and Whelan, M. J.},
  title = {Robust Parameterization of Elastic and Absorptive Electron Atomic Scattering Factors},
  journal = {Acta Crystallographica Section A: Foundations of Crystallography},
  volume = {52},
  number = {2},
  pages = {257--276},
  year = {1996},
  doi = {10.1107/S0108767395014371},
}

@article{kawabata2008gmm,
  author = {Kawabata, Takeshi},
  title = {Multiple Subunit Fitting into a Low-Resolution Density Map of a Macromolecular Complex Using a {Gaussian} Mixture Model},
  journal = {Biophysical Journal},
  volume = {95},
  number = {10},
  pages = {4643--4658},
  year = {2008},
  doi = {10.1529/biophysj.108.137125},
}

@article{chen2021e2gmm,
  author = {Chen, Muyuan and Ludtke, Steven J.},
  title = {Deep learning-based mixed-dimensional {Gaussian} mixture model for characterizing variability in cryo-{EM}},
  journal = {Nature Methods},
  volume = {18},
  number = {8},
  pages = {930--936},
  year = {2021},
  doi = {10.1038/s41592-021-01220-5},
}

@article{schwab2024dynamight,
  author = {Schwab, Johannes and Kimanius, Dari and Burt, Alister and Dendooven, Tom and Scheres, Sjors H. W.},
  title = {{DynaMight}: estimating molecular motions with improved reconstruction from {cryo-EM} images},
  journal = {Nature Methods},
  volume = {21},
  pages = {1855--1862},
  year = {2024},
  doi = {10.1038/s41592-024-02377-5},
}

@article{singer2020computational,
  author  = {Singer, Amit and Sigworth, Fred J.},
  title   = {Computational Methods for Single-Particle Electron Cryomicroscopy},
  journal = {Annual Review of Biomedical Data Science},
  year    = {2020},
  volume  = {3},
  pages   = {163--190},
  doi     = {10.1146/annurev-biodatasci-021020-093826},
}

@inproceedings{chen2026cryosplat,
  author    = {Chen, Suyi and Ling, Haibin},
  title     = {{CryoSplat}: Gaussian Splatting for {Cryo-EM} Homogeneous Reconstruction},
  booktitle = {The Fourteenth International Conference on Learning Representations},
  year      = {2026},
  url       = {https://proceedings.iclr.cc/paper_files/paper/2026/hash/5f173562e7662b14fb5c5695f225ea46-Abstract-Conference.html}
}

@article{slepian1964multidimensional,
  author = {Slepian, David},
  title = {Prolate Spheroidal Wave Functions, Fourier Analysis and Uncertainty---{IV}: Extensions to Many Dimensions; Generalized Prolate Spheroidal Functions},
  journal = {Bell System Technical Journal},
  volume = {43},
  number = {6},
  pages = {3009--3057},
  year = {1964},
  doi = {10.1002/j.1538-7305.1964.tb01037.x},
}

@article{simons2011spatiospectral,
  author = {Simons, Frederik J. and Wang, Dong V.},
  title = {Spatiospectral concentration in the {Cartesian} plane},
  journal = {GEM - International Journal on Geomathematics},
  volume = {2},
  number = {1},
  pages = {1--36},
  year = {2011},
  doi = {10.1007/s13137-011-0016-z},
}

@inproceedings{jeon2024cryobench,
  author    = {Jeon, Minkyu and Raghu, Rishwanth and Astore, Miro
               and Woollard, Geoffrey and Feathers, Ryan and Kaz, Alkin
               and Hanson, Sonya M. and Cossio, Pilar and Zhong, Ellen D.},
  title     = {{CryoBench}: Diverse and challenging datasets for the
               heterogeneity problem in {cryo-EM}},
  booktitle = {Advances in Neural Information Processing Systems},
  year      = {2024},
  volume    = {37},
  doi       = {10.52202/079017-2840},
}

@article{scheres2012relion,
  title={{RELION}: implementation of a Bayesian approach to cryo-EM structure determination},
  author={Scheres, Sjors HW},
  journal={Journal of structural biology},
  volume={180},
  number={3},
  pages={519--530},
  year={2012},
  publisher={Elsevier}
}

@article{punjani2017cryosparc,
  title={{cryoSPARC}: algorithms for rapid unsupervised cryo-EM structure determination},
  author={Punjani, Ali and Rubinstein, John L and Fleet, David J and Brubaker, Marcus A},
  journal={Nature methods},
  volume={14},
  number={3},
  pages={290--296},
  year={2017},
  publisher={Nature Publishing Group}
}

@article{tang2007eman2,
  title={{EMAN2}: an extensible image processing suite for electron microscopy},
  author={Tang, Guang and Peng, Liwei and Baldwin, Philip R and Mann, Deepinder S and Jiang, Wen and Rees, Ian and Ludtke, Steven J},
  journal={Journal of structural biology},
  volume={157},
  number={1},
  pages={38--46},
  year={2007},
  publisher={Elsevier}
}

@misc{qu2025cryonerf,
  title = {{CryoNeRF}: reconstruction of homogeneous and heterogeneous {cryo-EM} structures using neural radiance field},
  author = {Qu, Huaizhi and Wang, Xiao and Zhang, Yuanyuan and Wang, Sheng and Noble, William Stafford and Chen, Tianlong},
  howpublished = {bioRxiv preprint},
  year = {2025},
  doi = {10.1101/2025.01.10.632460},
}

@inproceedings{zhou2025cryofm,
  author    = {Zhou, Yi and Li, Yilai and Yuan, Jing and Gu, Quanquan},
  title     = {{CryoFM}: A Flow-based Foundation Model for {Cryo-EM} Densities},
  booktitle = {The Thirteenth International Conference on Learning Representations},
  year      = {2025},
  url       = {https://proceedings.iclr.cc/paper_files/paper/2025/file/6450ea28ebbc8437bc38775157818172-Paper-Conference.pdf}
}

@article{levy2024mixture,
  title={Mixture of neural fields for heterogeneous reconstruction in cryo-EM},
  author={Levy, Axel and Raghu, Rishwanth and Shustin, David and Peng, Adele and Li, Huan and Clarke, Oliver and Wetzstein, Gordon and Zhong, Ellen},
  journal={Advances in Neural Information Processing Systems},
  volume={37},
  pages={56988--57017},
  year={2024}
}

@article{kawabata2018gaussian,
  title={{Gaussian}-input {Gaussian} mixture model for representing density maps and atomic models},
  author={Kawabata, Takeshi},
  journal={Journal of structural biology},
  volume={203},
  number={1},
  pages={1--16},
  year={2018},
  publisher={Elsevier}
}

@article{chen2023integrating,
  title   = {Integrating molecular models into {CryoEM} heterogeneity analysis
             using scalable high-resolution deep {Gaussian} mixture models},
  author  = {Chen, Muyuan and Toader, Bogdan and Lederman, Roy},
  journal = {Journal of Molecular Biology},
  volume  = {435},
  number  = {9},
  pages   = {168014},
  year    = {2023},
  doi     = {10.1016/j.jmb.2023.168014},
}

@article{chen2024improving,
  author = {Chen, Muyuan and Schmid, Michael F. and Chiu, Wah},
  title = {Improving resolution and resolvability of single-particle cryo{EM} structures using {Gaussian} mixture models},
  journal = {Nature Methods},
  volume = {21},
  number = {1},
  pages = {37--40},
  year = {2024},
  doi = {10.1038/s41592-023-02082-9},
}

@inproceedings{shekarforoush2025cryospire,
  author    = {Shekarforoush, Shayan and Lindell, David B.
               and Brubaker, Marcus A. and Fleet, David J.},
  title     = {Reconstructing Heterogeneous Biomolecules via Hierarchical
               Gaussian Mixtures and Part Discovery},
  booktitle = {Advances in Neural Information Processing Systems},
  volume    = {38},
  year      = {2025},
  url       = {https://openreview.net/forum?id=CCbXvvcdF9}
}

@inproceedings{kheradmand2024mcmc,
  author = {Kheradmand, Shakiba and Rebain, Daniel and Sharma, Gopal and Sun, Weiwei and Tseng, Yang-Che and Isack, Hossam and Kar, Abhishek and Tagliasacchi, Andrea and Yi, Kwang Moo},
  title = {{3D Gaussian Splatting as Markov Chain Monte Carlo}},
  booktitle = {Advances in Neural Information Processing Systems},
  volume = {37},
  year = {2024},
  url = {https://proceedings.neurips.cc/paper_files/paper/2024/hash/93be245fce00a9bb2333c17ceae4b732-Abstract-Conference.html}
}

@inproceedings{cai2024radiative,
  author    = {Cai, Yuanhao and Liang, Yixun and Wang, Jiahao and Wang, Angtian
               and Zhang, Yulun and Yang, Xiaokang and Zhou, Zongwei
               and Yuille, Alan},
  title     = {Radiative Gaussian Splatting for Efficient {X-Ray} Novel View Synthesis},
  booktitle = {Computer Vision -- ECCV 2024},
  series    = {Lecture Notes in Computer Science},
  volume    = {15059},
  publisher = {Springer},
  year      = {2025},
  pages     = {283--299},
  doi       = {10.1007/978-3-031-73232-4_16},
}

@inproceedings{zha2024r2gaussian,
  author    = {Zha, Ruyi and Lin, Tao Jun and Cai, Yuanhao and Cao, Jiwen
               and Zhang, Yanhao and Li, Hongdong},
  title     = {{R$^2$-Gaussian}: Rectifying Radiative Gaussian Splatting
               for Tomographic Reconstruction},
  booktitle = {Advances in Neural Information Processing Systems},
  volume    = {37},
  year      = {2024},
  doi       = {10.52202/079017-1427},
}

@misc{nistChiSquareTable,
  author = {{NIST/SEMATECH}},
  title = {Critical Values of the Chi-Square Distribution},
  howpublished = {e-Handbook of Statistical Methods, Section 1.3.6.7.4},
  year = {n.d.},
  note = {Accessed September 2026},
  url = {https://www.itl.nist.gov/div898/handbook/eda/section3/eda3674.htm}
}

@article{herreros2025hetsiren,
  author  = {Herreros, David and Mata, Carlos Perez and Noddings, Chari and Irene, Deli and Krieger, James and Agard, David A. and Tsai, Ming-Daw and Sorzano, Carlos Oscar Sanchez and Carazo, Jose Maria},
  title   = {Real-space heterogeneous reconstruction, refinement, and disentanglement of {CryoEM} conformational states with {HetSIREN}},
  journal = {Nature Communications},
  volume  = {16},
  number  = {1},
  pages   = {3751},
  year    = {2025},
  doi     = {10.1038/s41467-025-59135-0},
}

@article{kimanius2024blush,
  author  = {Kimanius, Dari and Jamali, Kiarash and Wilkinson, Max E. and L{\"o}vestam, Sofia and Velazhahan, Vaithish and Nakane, Takanori and Scheres, Sjors H. W.},
  title   = {Data-driven regularization lowers the size barrier of {cryo-EM} structure determination},
  journal = {Nature Methods},
  volume  = {21},
  number  = {7},
  pages   = {1216--1221},
  year    = {2024},
  doi     = {10.1038/s41592-024-02304-8},
}

@inproceedings{yu2024mipsplatting,
  author    = {Yu, Zehao and Chen, Anpei and Huang, Binbin and Sattler, Torsten and Geiger, Andreas},
  title     = {{Mip-Splatting}: Alias-free {3D Gaussian Splatting}},
  booktitle = {Proceedings of the IEEE/CVF Conference on Computer Vision and Pattern Recognition (CVPR)},
  pages     = {19447--19456},
  month     = jun,
  year      = {2024},
  doi       = {10.1109/CVPR52733.2024.01839},
}

@article{zhu2024cryonefen,
  author  = {Huang, Yue and Zhu, Chengguang and Yang, Xiaokang and Liu, Manhua},
  title   = {High-resolution real-space reconstruction of {cryo-EM} structures using a neural field network},
  journal = {Nature Machine Intelligence},
  volume  = {6},
  number  = {8},
  pages   = {892--903},
  year    = {2024},
  doi     = {10.1038/s42256-024-00870-2},
}

@article{li2024cryostar,
  author  = {Li, Yilai and Zhou, Yi and Yuan, Jing and Ye, Fei and Gu, Quanquan},
  title   = {{CryoSTAR}: leveraging structural priors and constraints for {cryo-EM} heterogeneous reconstruction},
  journal = {Nature Methods},
  volume  = {21},
  number  = {12},
  pages   = {2318--2326},
  year    = {2024},
  doi     = {10.1038/s41592-024-02486-1},
}

@article{scheres2012overfitting,
  author  = {Scheres, Sjors H. W. and Chen, Shaoxia},
  title   = {Prevention of overfitting in {cryo-EM} structure determination},
  journal = {Nature Methods},
  year    = {2012},
  volume  = {9},
  number  = {9},
  pages   = {853--854},
  doi     = {10.1038/nmeth.2115},
}

@article{rabuckgibbons2022mining,
  author  = {Rabuck-Gibbons, Jessica N. and Lyumkis, Dmitry
             and Williamson, James R.},
  title   = {Quantitative mining of compositional heterogeneity in
             {cryo-EM} datasets of ribosome assembly intermediates},
  journal = {Structure},
  year    = {2022},
  volume  = {30},
  number  = {4},
  pages   = {498--509.e4},
  doi     = {10.1016/j.str.2021.12.005},
}

@article{scheres2012bayesian,
  author  = {Scheres, Sjors H. W.},
  title   = {A Bayesian View on {Cryo-EM} Structure Determination},
  journal = {Journal of Molecular Biology},
  year    = {2012},
  volume  = {415},
  number  = {2},
  pages   = {406--418},
  doi     = {10.1016/j.jmb.2011.11.010},
}

@article{shannon1949communication,
  author  = {Shannon, Claude E.},
  title   = {Communication in the Presence of Noise},
  journal = {Proceedings of the IRE},
  volume  = {37},
  number  = {1},
  pages   = {10--21},
  year    = {1949},
  doi     = {10.1109/JRPROC.1949.232969},
}

@article{katsevich2015covariance,
  author  = {Katsevich, Eugene and Katsevich, Alexander and Singer, Amit},
  title   = {Covariance Matrix Estimation for the {Cryo-EM} Heterogeneity Problem},
  journal = {SIAM Journal on Imaging Sciences},
  volume  = {8},
  number  = {1},
  pages   = {126--185},
  year    = {2015},
  doi     = {10.1137/130935434},
}

@article{scheres2007disentangling,
  author  = {Scheres, Sjors H. W. and Gao, Haixiao and Valle, Mikel
             and Herman, Gabor T. and Eggermont, Paul P. B.
             and Frank, Joachim and Carazo, Jose-Maria},
  title   = {Disentangling conformational states of macromolecules in
             {3D-EM} through likelihood optimization},
  journal = {Nature Methods},
  volume  = {4},
  number  = {1},
  pages   = {27--29},
  year    = {2007},
  doi     = {10.1038/nmeth992},
}

@article{nakane2018multibody,
  author  = {Nakane, Takanori and Kimanius, Dari and Lindahl, Erik
             and Scheres, Sjors H. W.},
  title   = {Characterisation of molecular motions in {cryo-EM}
             single-particle data by multi-body refinement in {RELION}},
  journal = {eLife},
  volume  = {7},
  pages   = {e36861},
  year    = {2018},
  doi     = {10.7554/eLife.36861},
}

@article{levy2025cryodrgnai,
  author  = {Levy, Axel and Raghu, Rishwanth and Feathers, J. Ryan
             and Grzadkowski, Michal and Poitevin, Fr{\'e}d{\'e}ric
             and Johnston, Jake D. and Vallese, Francesca
             and Clarke, Oliver Biggs and Wetzstein, Gordon and Zhong, Ellen D.},
  title   = {{CryoDRGN-AI}: neural ab initio reconstruction of challenging
             {cryo-EM} and {cryo-ET} datasets},
  journal = {Nature Methods},
  volume  = {22},
  number  = {7},
  pages   = {1486--1494},
  year    = {2025},
  doi     = {10.1038/s41592-025-02720-4},
}

@inproceedings{gu2026cryohype,
  author    = {Gu, Jeffrey and Jeon, Minkyu and Ma, Ambri
               and Yeung-Levy, Serena and Zhong, Ellen D.},
  title     = {{CryoHype}: Reconstructing a thousand {cryo-EM} structures
               with transformer-based hypernetworks},
  booktitle = {Proceedings of the IEEE/CVF Conference on Computer Vision
               and Pattern Recognition},
  year      = {2026},
  pages     = {35280--35290},
  url       = {https://openaccess.thecvf.com/content/CVPR2026/html/Gu_CryoHype_Reconstructing_a_thousand_cryo-EM_structures_with_transformer-based_hypernetworks_CVPR_2026_paper.html}
}

@inproceedings{westover1990footprint,
  author    = {Westover, Lee},
  title     = {Footprint Evaluation for Volume Rendering},
  booktitle = {Proceedings of the 17th Annual Conference on Computer
               Graphics and Interactive Techniques},
  series    = {SIGGRAPH '90},
  year      = {1990},
  pages     = {367--376},
  publisher = {Association for Computing Machinery},
  doi       = {10.1145/97879.97919},
}

@inproceedings{zwicker2001ewa,
  author    = {Zwicker, Matthias and Pfister, Hanspeter
               and van Baar, Jeroen and Gross, Markus},
  title     = {{EWA} Volume Splatting},
  booktitle = {Proceedings of IEEE Visualization 2001},
  year      = {2001},
  pages     = {29--36},
  publisher = {IEEE},
  doi       = {10.1109/VISUAL.2001.964490},
}

@inproceedings{talegaonkar2025volumetric,
  author    = {Talegaonkar, Chinmay and Belhe, Yash
               and Ramamoorthi, Ravi and Antipa, Nicholas},
  title     = {Volumetrically Consistent {3D Gaussian} Rasterization},
  booktitle = {Proceedings of the IEEE/CVF Conference on Computer Vision
               and Pattern Recognition},
  year      = {2025},
  pages     = {10953--10963},
  doi       = {10.1109/CVPR52734.2025.01023},
}

@article{davis2016assembly,
  author  = {Davis, Joseph H. and Tan, Yong Zi and Carragher, Bridget
             and Potter, Clinton S. and Lyumkis, Dmitry
             and Williamson, James R.},
  title   = {Modular Assembly of the Bacterial Large Ribosomal Subunit},
  journal = {Cell},
  year    = {2016},
  volume  = {167},
  number  = {6},
  pages   = {1610--1622.e15},
  doi     = {10.1016/j.cell.2016.11.020},
}

@article{plaschka2017spliceosome,
  author  = {Plaschka, Clemens and Lin, Pei-Chun and Nagai, Kiyoshi},
  title   = {Structure of a pre-catalytic spliceosome},
  journal = {Nature},
  year    = {2017},
  volume  = {546},
  number  = {7660},
  pages   = {617--621},
  doi     = {10.1038/nature22799},
}

@article{hyndman1996hdr,
  author  = {Hyndman, Rob J.},
  title   = {Computing and Graphing Highest Density Regions},
  journal = {The American Statistician},
  volume  = {50},
  number  = {2},
  pages   = {120--126},
  year    = {1996},
  doi     = {10.1080/00031305.1996.10474359},
}

@article{qin2023assembly,
  author = {Qin, Bo and Lauer, Simon M. and Balke, Annika and Vieira-Vieira, Carlos H. and B{\"u}rger, J{\"o}rg and Mielke, Thorsten and Selbach, Matthias and Scheerer, Patrick and Spahn, Christian M. T. and Nikolay, Rainer},
  title = {{Cryo-EM} captures early ribosome assembly in action},
  journal = {Nature Communications},
  volume = {14},
  number = {1},
  pages = {898},
  year = {2023},
  doi = {10.1038/s41467-023-36607-9},
}

@article{wieczor2023spike,
  author = {Wiecz{\'o}r, Mi{\l}osz and Tang, Phu K. and Orozco, Modesto and Cossio, Pilar},
  title = {Omicron mutations increase interdomain interactions and reduce epitope exposure in the {SARS-CoV-2} spike},
  journal = {iScience},
  volume = {26},
  number = {2},
  pages = {105981},
  year = {2023},
  doi = {10.1016/j.isci.2023.105981},
}

@article{yadgar2026solvar,
  author = {Yadgar, Roey and Lederman, Roy R. and Shkolnisky, Yoel},
  title = {{SOLVAR}: Fast covariance-based heterogeneity analysis with pose refinement for {cryo-EM}},
  journal = {arXiv preprint arXiv:2602.17603},
  year = {2026}
}

@inproceedings{gao2026cryokraqen,
  author = {Gao, Wenyuan and Wu, Yutan and He, Xuming},
  title = {{CryoKRAQEN}: Kernel-Regularized Annealing for Quantized Embedding Networks in {Cryo-EM} Heterogeneous Reconstruction},
  booktitle = {Proceedings of the IEEE/CVF Conference on Computer Vision and Pattern Recognition},
  pages = {28298--28307},
  year = {2026}
}

@inproceedings{kingma2015adam,
  author    = {Kingma, Diederik P. and Ba, Jimmy},
  title     = {{Adam}: A Method for Stochastic Optimization},
  booktitle = {International Conference on Learning Representations},
  year      = {2015},
}

@inproceedings{zhao2025scaling,
  author    = {Zhao, Hexu and Weng, Haoyang and Lu, Daohan and Li, Ang and Li, Jinyang and Panda, Aurojit and Xie, Saining},
  title     = {On Scaling Up {3D Gaussian Splatting} Training},
  booktitle = {The Thirteenth International Conference on Learning Representations},
  year      = {2025},
  url       = {https://openreview.net/forum?id=pQqeQpMkE7},
}

@article{ye2025gsplat,
  author  = {Ye, Vickie and Li, Ruilong and Kerr, Justin and Turkulainen, Matias and Yi, Brent and Pan, Zhuoyang and Seiskari, Otto and Ye, Jianbo and Hu, Jeffrey and Tancik, Matthew and Kanazawa, Angjoo},
  title   = {{gsplat}: An Open-Source Library for {Gaussian Splatting}},
  journal = {Journal of Machine Learning Research},
  volume  = {26},
  number  = {34},
  pages   = {1--17},
  year    = {2025},
  url     = {https://jmlr.org/papers/v26/24-1476.html},
}

@article{wang2026proteinzero,
  title = {{ProteinZero}: Self-Improving Protein Generation via Online Reinforcement Learning},
  author = {Wang, Ziwen and Fan, Jiajun and Guo, Ruihan and Nguyen, Thao and Ji, Heng and Liu, Ge},
  journal = {Transactions on Machine Learning Research},
  year = {2026},
  issn = {2835-8856},
  url = {https://openreview.net/forum?id=pVvm0CQAr6}
}

@article{peng2026pocketxmol,
  title = {Unified modeling of {3D} molecular generation via atomic interactions with {PocketXMol}},
  author = {Peng, Xingang and Guo, Ruihan and Guo, Fenglin and Wang, Ziyi and Sun, Jiayu and Guan, Jiaqi and Jia, Yinjun and Xu, Yan and Huang, Yanwen and Zhang, Muhan and Peng, Jian and Wang, Xinquan and Han, Chuanhui and Wang, Zihua and Ma, Jianzhu},
  journal = {Cell},
  volume = {189},
  number = {7},
  pages = {1904--1922.e28},
  year = {2026},
  doi = {10.1016/j.cell.2026.01.003},
  url = {https://doi.org/10.1016/j.cell.2026.01.003}
}

@article{boiko2023autonomous,
  title = {Autonomous chemical research with large language models},
  author = {Boiko, Daniil A. and MacKnight, Robert and Kline, Ben and Gomes, Gabe},
  journal = {Nature},
  volume = {624},
  number = {7992},
  pages = {570--578},
  year = {2023},
  doi = {10.1038/s41586-023-06792-0},
  url = {https://doi.org/10.1038/s41586-023-06792-0}
}

@article{jamali2024automated,
  title = {Automated model building and protein identification in cryo-{EM} maps},
  author = {Jamali, Kiarash and K{\"a}ll, Lukas and Zhang, Rui and Brown, Alan and Kimanius, Dari and Scheres, Sjors H. W.},
  journal = {Nature},
  volume = {628},
  number = {8007},
  pages = {450--457},
  year = {2024},
  doi = {10.1038/s41586-024-07215-4},
  url = {https://doi.org/10.1038/s41586-024-07215-4}
}

@inproceedings{mallick2024taming,
  author    = {Mallick, Saswat Subhajyoti and Goel, Rahul and Kerbl, Bernhard and Vicente Carrasco, Francisco and Steinberger, Markus and De La Torre, Fernando},
  title     = {Taming {3DGS}: High-Quality Radiance Fields with Limited Resources},
  booktitle = {SIGGRAPH Asia 2024 Conference Papers},
  year      = {2024},
  url       = {https://arxiv.org/abs/2406.15643}
}

@inproceedings{wang2024adr,
  author    = {Wang, Xinzhe and Yi, Ran and Ma, Lizhuang},
  title     = {{AdR-Gaussian}: Accelerating Gaussian Splatting with Adaptive Radius},
  booktitle = {SIGGRAPH Asia 2024 Conference Papers},
  year      = {2024},
  url       = {https://doi.org/10.1145/3680528.3687675}
}

@article{gui2024balanced,
  author  = {Gui, Hao and Hu, Lin and Chen, Rui and Huang, Mingxiao and Yin, Yuxin and Yang, Jin and Wu, Yong and Liu, Chen and Sun, Zhongxu and Zhang, Xueyang and Zhan, Kun},
  title   = {Balanced {3DGS}: Gaussian-wise Parallelism Rendering with Fine-Grained Tiling},
  journal = {arXiv preprint arXiv:2412.17378},
  year    = {2024}
}

@inproceedings{hanson2025speedy,
  author    = {Hanson, Alex and Tu, Allen and Lin, Geng and Singla, Vasu and Zwicker, Matthias and Goldstein, Tom},
  title     = {{Speedy-Splat}: Fast {3D} Gaussian Splatting with Sparse Pixels and Sparse Primitives},
  booktitle = {Proceedings of the IEEE/CVF Conference on Computer Vision and Pattern Recognition},
  year      = {2025},
  url       = {https://arxiv.org/abs/2412.00578}
}

@inproceedings{yang2024deformable,
  author    = {Yang, Ziyi and Gao, Xinyu and Zhou, Wen and Jiao, Shaohui and Zhang, Yuqing and Jin, Xiaogang},
  title     = {Deformable {3D} Gaussians for High-Fidelity Monocular Dynamic Scene Reconstruction},
  booktitle = {Proceedings of the IEEE/CVF Conference on Computer Vision and Pattern Recognition},
  year      = {2024},
  url       = {https://arxiv.org/abs/2309.13101}
}

@inproceedings{wu20244dgs,
  author    = {Wu, Guanjun and Yi, Taoran and Fang, Jiemin and Xie, Lingxi and Zhang, Xiaopeng and Wei, Wei and Liu, Wenyu and Tian, Qi and Wang, Xinggang},
  title     = {{4D} Gaussian Splatting for Real-Time Dynamic Scene Rendering},
  booktitle = {Proceedings of the IEEE/CVF Conference on Computer Vision and Pattern Recognition},
  year      = {2024},
  url       = {https://arxiv.org/abs/2310.08528}
}

@article{qu2025gem,
  author  = {Qu, Huaizhi and Wang, Xiao and Zhang, Gengwei and Peng, Jie and Chen, Tianlong},
  title   = {{GEM}: {3D} Gaussian Splatting for Efficient and Accurate Cryo-EM Reconstruction},
  journal = {arXiv preprint arXiv:2509.25075},
  year    = {2025}
}

@article{he2025gaussianem,
  author  = {He, Bintao and Cheng, Yiran and Li, Hongjia and Gao, Xiang and Gao, Xin and Zhang, Fa and Han, Renmin},
  title   = {Resolving Compositional and Conformational Heterogeneity in Cryo-EM with Deformable {3D} Gaussian Representations},
  journal = {arXiv preprint arXiv:2512.21599},
  year    = {2025}
}

@article{chen2016training,
  author  = {Chen, Tianqi and Xu, Bing and Zhang, Chiyuan and Guestrin, Carlos},
  title   = {Training Deep Nets with Sublinear Memory Cost},
  journal = {arXiv preprint arXiv:1604.06174},
  year    = {2016},
}

@article{bendory2020single,
  author  = {Bendory, Tamir and Bartesaghi, Alberto and Singer, Amit},
  title   = {Single-Particle Cryo-Electron Microscopy: Mathematical Theory, Computational Challenges, and Opportunities},
  journal = {IEEE Signal Processing Magazine},
  volume  = {37},
  number  = {2},
  pages   = {58--76},
  year    = {2020},
  doi     = {10.1109/MSP.2019.2957822}
}

@article{mcinnes2018umap,
  author  = {McInnes, Leland and Healy, John and Melville, James},
  title   = {{UMAP}: Uniform Manifold Approximation and Projection for Dimension Reduction},
  journal = {arXiv preprint arXiv:1802.03426},
  year    = {2018}
}

@inproceedings{bae2024pergaussian,
  author    = {Bae, Jeongmin and Kim, Seoha and Yun, Youngsik and Lee, Hahyun
               and Bang, Gun and Uh, Youngjung},
  title     = {Per-{Gaussian} Embedding-Based Deformation for Deformable {3D} {Gaussian} Splatting},
  booktitle = {Computer Vision -- ECCV 2024},
  series    = {Lecture Notes in Computer Science},
  volume    = {15073},
  publisher = {Springer},
  year      = {2025},
  pages     = {321--335},
  doi       = {10.1007/978-3-031-72633-0_18},
}

\clearpage
\appendix

\etocdepthtag.toc{appendix}
\etocsettagdepth{mainmatter}{none}
\etocsettagdepth{appendix}{subsection}
\renewcommand{\contentsname}{Appendix contents}
\tableofcontents
\clearpage

\section{Gaussian parameterization}
\label{app:gaussian_parameterization}

\subsection{Base Gaussian attributes}

The base amplitude is parameterized as
$p_g^{\mathrm{base}}=\operatorname{sigmoid}(\theta_g)$. Without latent
conditioning, $p_g=p_g^{\mathrm{base}}$ and
$\boldsymbol\mu_g=\boldsymbol\mu_g^{\mathrm{base}}$, and the covariance is
\begin{equation}
  \boldsymbol\Sigma_g
  =Q_g\operatorname{diag}(\sigma_{g1}^2,\sigma_{g2}^2,\sigma_{g3}^2)Q_g^{\mathsf T},
  \qquad Q_g\in\mathrm{SO}(3),
  \label{eq:gaussian_covariance}
\end{equation}
where each effective principal-axis scale satisfies the Gaussian FWHM constraint
in Equation~\eqref{eq:fwhm_constraint}. We express scales in physical units
throughout, using $s_{gk}^{\mathrm{base}}>0$ for a learned base scale before applying
the constraint.

Gaussian positions are sampled from an isotropic Gaussian distribution
centered in the reconstruction box. We initialize all three base scales
to $2a$ and each rotation matrix $Q_g$ to the identity matrix $I_3$.

\subsection{Gaussian attribute decoders}
\label{app:decoder_implementation}

The position and amplitude decoders follow
Equation~\eqref{eq:state_position_amplitude}. The scale decoder produces log-scale
modulation $\Delta\boldsymbol\ell_g(\mathbf z)\in\mathbb R^3$ from
$[\mathbf f_g;\mathbf z]$ and modulates each base scale separately:
\begin{equation}
  \widetilde\sigma_{gk}(\mathbf z)
  =s_{gk}^{\mathrm{base}}
     \exp\!\left[\Delta\ell_{gk}(\mathbf z)\right],
  \qquad k=1,2,3.
  \label{eq:decoded_scales}
\end{equation}
Applying the Gaussian FWHM constraint to each principal-axis scale gives the
effective scales $\sigma_{gk}(\mathbf z)$ and the effective covariance of
Section~\ref{sec:heterogeneous_learning},
\begin{equation}
\begin{aligned}
  \boldsymbol\Sigma_g(\mathbf z)
  &=Q_g\operatorname{diag}\!\left(
      \sigma_{g1}(\mathbf z)^2,
      \sigma_{g2}(\mathbf z)^2,
      \sigma_{g3}(\mathbf z)^2\right)Q_g^{\mathsf T}, \\
  2\sqrt{2\ln2}\,\sigma_{gk}(\mathbf z)
  &\geq d,\qquad k=1,2,3,
\end{aligned}
  \label{eq:state_effective_scale}
\end{equation}
where $Q_g\in\mathrm{SO}(3)$ is the learned Gaussian rotation.

\paragraph{Decoder configuration.}
Each Gaussian attribute decoder takes $[\mathbf f_g;\mathbf z]$ as input
and uses one hidden layer of width $32$ with a ReLU activation. Latent
dimensions are specified in Section~\ref{sec:experiments}. Separate decoders
produce position modulation, amplitude modulation, and scale modulation; all three
are trained during heterogeneity estimation and per-state reconstruction. The position decoder predicts a
three-dimensional position modulation that is added directly to the base
Gaussian position. Amplitude modulation follows the sigmoid
parameterization in Equation~\eqref{eq:state_position_amplitude}.

\subsection{Gaussian relocation at a fixed number of Gaussians}
\label{app:gaussian_relocation}

Inspired by the relocation strategy of 3DGS-MCMC~\citep{kheradmand2024mcmc},
we redistribute existing Gaussians while retaining the predetermined number of
Gaussians $G$. Gaussians are selected for relocation when they
have low base amplitudes. For each
selected Gaussian, we sample a Gaussian from those not selected for relocation,
with replacement and with probabilities proportional to their base amplitudes. The relocated Gaussian copies
the base position, scales, and rotation of the sampled Gaussian. If
Gaussian $h$ is sampled $m_h$ times, its base amplitude before relocation is divided
equally among Gaussian $h$ and the $m_h$ relocated Gaussians, so that each receives
$p_h^{\mathrm{base}}/(m_h+1)$.

\section{Experimental settings}
\label{app:experimental_settings}

\subsection{Datasets and preprocessing}
\label{app:experimental_data}

Table~\ref{tab:cryobench_data} summarizes the CryoBench particle images used
in our experiments. Table~\ref{tab:experimental_data} summarizes the source
particle-image datasets; particle selection and preprocessing are described
below.

\begin{table*}[t]
\centering
\small
\caption{\textbf{CryoBench datasets used in the main comparison.} Dimensions and pixel
sizes describe the processed images; the signal-to-noise ratio identifies the noise condition of
the released dataset. For Spike-MD, reference states are the 46,789
distinct molecular-dynamics configurations reported by CryoBench.}
\label{tab:cryobench_data}
\setlength{\tabcolsep}{3pt}
\renewcommand{\arraystretch}{1.15}
\begin{tabular}{@{}lrrrrr@{}}
\toprule
Dataset & \shortstack[r]{Number of\\particle images}
& \shortstack[r]{Reference\\states}
& \shortstack[r]{Image dimensions\\(pixels)}
& \shortstack[r]{Pixel size\\(\AA/pixel)}
& \shortstack[r]{Signal-to-noise\\ratio} \\
\midrule
IgG-1D & 100,000 & 100 & $128\times128$ & 3.0 & 0.01 \\
IgG-RL & 100,000 & 100 & $128\times128$ & 3.0 & 0.01 \\
Ribosembly & 335,240 & 16 & $128\times128$ & 3.0 & 0.01 \\
Tomotwin-100 & 100,000 & 100 & $128\times128$ & 4.5 & 0.01 \\
Spike-MD & 100,000 & 46,789 & $256\times256$ & 1.5 & 0.1 \\
\bottomrule
\end{tabular}
\end{table*}

\paragraph{CryoBench data.}
We use the released CryoBench datasets~\citep{jeon2024cryobench} and their
supplied pose and CTF metadata, without additional particle selection.
IgG-1D, IgG-RL, and Tomotwin-100 each contain 1,000 particle images per state;
Ribosembly has unequal state populations. IgG-1D, IgG-RL, and Ribosembly are
Fourier-downsampled from $256\times256$ to $128\times128$, and Tomotwin-100
from $384\times384$ to $128\times128$. In each case, the physical field of
view is preserved. Spike-MD is used at its released image dimensions.
Its reference-state count follows the sampled molecular-dynamics
configurations rather than a partition into discrete structural classes.

\begin{table*}[t]
\centering
\small
\caption{\textbf{Experimental particle-image datasets.} Particle selection and
preprocessing are described in Appendix~\ref{app:experimental_data}. Pixel sizes refer to the deposited
particle images.}
\label{tab:experimental_data}
\setlength{\tabcolsep}{4pt}
\renewcommand{\arraystretch}{1.15}
\begin{tabular}{@{}llrrr@{}}
\toprule
Dataset & Sample & \shortstack[r]{Deposited\\particle images}
& \shortstack[r]{Image dimensions\\(pixels)}
& \shortstack[r]{Pixel size\\(\AA/pixel)} \\
\midrule
EMPIAR-10076~\citep{davis2016assembly} & Assembling 50S ribosomal subunit & 131,899 & $320\times320$ & 1.31 \\
EMPIAR-10180~\citep{plaschka2017spliceosome} & Pre-catalytic spliceosome & 327,490 & $320\times320$ & 1.699 \\
\bottomrule
\end{tabular}
\end{table*}

\paragraph{Experimental data.}
For the main EMPIAR-10076 and EMPIAR-10180 reconstructions, we follow the
particle selection and preprocessing procedures provided with
CryoDRGN~\citep{zhong2021cryodrgn}, using the corresponding pose and CTF
metadata. We Fourier-downsample the $320\times320$ images from EMPIAR-10076
and EMPIAR-10180 to $256\times256$, following
CryoDRGN~\citep{zhong2021cryodrgn}. This preserves the physical field of
view and yields pixel sizes of 1.6375 and 2.12375~\AA/pixel, respectively. EMPIAR-10076 contains
131,899 source images; its CryoDRGN-provided selection contains 87,327
images.
For EMPIAR-10180, the main reconstruction uses 139,720 particle images
from the CryoDRGN-filtered subset. We exclude corrupted particle 308902 from the 327,490-image
source stack, leaving 327,489 images for the separate full-stack
analysis in Figure~\ref{fig:empiar10180_full_dataset}, Fourier-downsampled
to $128\times128$ at 4.2475~\AA/pixel following
Zhong et~al.~\citep{zhong2021cryodrgn}.

\paragraph{CryoGaussian particle selection for EMPIAR-10076.}
For the separate filtering analysis in Figure~\ref{fig:empiar10076_filtering},
we follow the filtering strategy of Zhong et~al.~\citep{zhong2021cryodrgn} with one- and
four-dimensional models trained on all 131,899 particle images. The
four-dimensional model uses the same latent dimension $d_z=4$ as our
heterogeneous reconstruction. For the one-dimensional latents, we form a
Gaussian kernel density estimate and exclude the interval between the
two local minima adjacent to its highest peak, retaining 100,528 images.
For the four-dimensional latents, we fit a five-component Gaussian mixture
and exclude the component with the largest mean latent norm, retaining
105,211 images. The intersection of the two retained sets contains
97,651 images. We train the subsequent reconstruction on a random 90\%
of this intersection, comprising 87,885 images. Published class labels
are used only for post-hoc evaluation of the selection.

\subsection{Training settings}
\label{app:training_ablation_settings}

Heterogeneity estimation and per-state reconstruction use the same particle
images, pose and CTF metadata, and particle selections, and each use 30,000
iterations (Section~\ref{sec:experiments}).
We set the base learning rates to $2\times10^{-4}$ for Gaussian
positions; $10^{-3}$ for Gaussian scales, rotations, base-amplitude
parameters, and trainable particle-image latents; and $10^{-4}$ for
per-Gaussian latents and the position, amplitude, and scale decoders.
These rates are defined for a global batch size of one and scaled by
the square root of the global batch size, consistent with the learning-rate
scaling in Grendel and gsplat~\citep{zhao2025scaling,ye2025gsplat}.
The Gaussian position learning rate
decays to 1\% of its initial value, following
3DGS and X-Gaussian~\citep{kerbl2023gaussian,cai2024radiative}.
We use Adam~\citep{kingma2015adam} with momentum coefficients
$(0.9,0.999)$, $\epsilon=10^{-15}$, and zero global weight decay.
Inspired by
3DGS-MCMC~\citep{kheradmand2024mcmc}, we relocate Gaussians every
100 iterations starting at iteration 500 and stopping at iteration 25,000
(Appendix~\ref{app:gaussian_relocation}).
Latent dimensions are specified in Section~\ref{sec:experiments} and the
relative density threshold in Section~\ref{sec:gaussian_capacity}.
For IgG-1D, IgG-RL, Tomotwin-100, Spike-MD, EMPIAR-10076,
and EMPIAR-10180, global batch sizes are 256 for heterogeneity estimation
and 128 for per-state reconstruction. For the larger Ribosembly particle
stack, we use 512 and 256, respectively. CryoGaussian results are from one
training run. To assess variability across training seeds, heterogeneity
estimation and per-state reconstruction are additionally rerun with 10
training seeds for the clustering accuracy in
Table~\ref{tab:clustering_accuracy} and the seed-variability analysis in
Table~\ref{tab:seed_variability}.

\subsection{Training time and memory}
\label{app:computational_cost}

We run CryoGaussian on eight NVIDIA RTX PRO 6000 Blackwell GPUs. Table~\ref{tab:computational_cost} reports global batch
sizes, times per iteration, training times, forward throughput, and peak
allocated memory per GPU for heterogeneous reconstruction, together with
the speedups and memory reductions of Table~\ref{tab:rasterizer_speedup}.
Table~\ref{tab:rasterizer_speedup} isolates the two rasterizer
optimizations of Section~\ref{sec:efficient_optimization} and distributed
training with three implementations of the rasterizer that evaluate the
same image-formation model and differ only in how projections and
gradients are scheduled. The CryoSplat rasterizer~\citep{chen2026cryosplat}
scatters gradients pixel-wise with atomic updates and processes each tile
list in one pass,
the second accumulates gradients per Gaussian
(Appendix~\ref{app:gaussian_gradient_accumulation}), and the third
additionally partitions long tile lists into independent partial
projections (Appendix~\ref{app:partitioned_projection}) and is the
rasterizer of CryoGaussian. Speedups are ratios of mean time per
iteration, forward and backward passes and the optimizer update included,
over training runs with the global batch sizes of
Table~\ref{tab:computational_cost} and otherwise identical settings. The
two rasterizer speedups are measured on one GPU, so that they exclude
communication between GPUs, and are applied cumulatively, the first from
the CryoSplat rasterizer to the second and the second from the second to
the third, so that their product, 12.8--28.0 across the seven datasets, is
the speedup of the CryoGaussian rasterizer over the CryoSplat rasterizer. The distributed-training speedup is the ratio of the mean
time per iteration of the CryoGaussian rasterizer on one GPU to that on
eight GPUs at the same global batch, with each GPU processing one eighth
of the batch, and corresponds to a strong-scaling efficiency, the speedup
divided by the number of GPUs, of 64--81\%. Releasing the forward
workspaces before the backward pass
(Appendix~\ref{app:partitioned_projection}) lowers peak allocated memory
per GPU by 8.0--10.2\% at these global batch sizes, and
Table~\ref{tab:computational_cost} reports the peak with the workspaces
retained and this reduction. These optimizations make the numbers of Gaussians given by
Equation~\eqref{eq:gaussian_count} and the full box sizes of the
experimental datasets practical to optimize, with training times of
6.8--8.5 h on eight GPUs for EMPIAR-10076 and EMPIAR-10180
(Table~\ref{tab:computational_cost}).

\begin{table*}[t]
    \centering
    \caption{\textbf{Training time and memory of CryoGaussian on eight NVIDIA RTX PRO 6000 Blackwell GPUs.}
    We report global batch sizes. Estimated numbers of Gaussians at $\tau=0.02$
    are rounded to the nearest integer. Peak allocated memory per GPU is
    reported with the forward workspaces retained, and the memory saved is
    the reduction of this peak when the workspaces are released before the
    backward pass (Appendix~\ref{app:partitioned_projection}). The speedups of per-Gaussian
    gradient accumulation, projection scheduling, and eight GPUs over one GPU
    and the memory saved are those of
    Table~\ref{tab:rasterizer_speedup}, measured in the same way for the
    other configurations.}
    \label{tab:computational_cost}
    \footnotesize
    \setlength{\tabcolsep}{2.6pt}
    \renewcommand{\arraystretch}{1.15}
    \resizebox{\linewidth}{!}{%
    \begin{tabular}{@{}lrrrrrrrrrrr@{}}
        \toprule
        Dataset & Box & \shortstack{Estimated\\Gaussians}
        & \shortstack{Global\\batch}
        & \shortstack{Time per\\iteration\\(ms)}
        & \shortstack{Training\\time (h)}
        & \shortstack{Forward\\(particles/s)}
        & \shortstack{Peak\\allocated\\(GiB/GPU)}
        & \shortstack{Gradient\\accumu-\\lation}
        & \shortstack{Projection\\sched-\\uling}
        & \shortstack{8\\GPUs}
        & \shortstack{Memory\\saved\\(\%)} \\
        \midrule
        EMPIAR-10180 & 256 & 448,148 & 256 & 376.64 & 6.83 & 956.3 & 12.14 & 2.64$\times$ & 8.51$\times$ & 5.59$\times$ & 9.95 \\
        EMPIAR-10076 (CryoDRGN-filtered) & 256 & 513,705 & 256 & 520.19 & 8.50 & 841.1 & 14.14 & 2.35$\times$ & 5.43$\times$ & 6.18$\times$ & 9.63 \\
        EMPIAR-10076 (CryoGaussian-filtered) & 256 & 463,802 & 256 & 464.01 & 7.63 & 950.5 & 12.78 & 2.25$\times$ & 5.58$\times$ & 6.12$\times$ & 9.86 \\
        \addlinespace[3pt]
        IgG-1D (SNR 0.01) & 128 & 10,316 & 256 & 17.31 & 0.30 & 27,250.4 & 0.33 & 2.39$\times$ & 7.69$\times$ & 5.41$\times$ & 7.98 \\
        IgG-1D (SNR 0.005) & 128 & 10,736 & 256 & 17.94 & 0.31 & 25,501.7 & 0.34 & 2.44$\times$ & 7.47$\times$ & 5.75$\times$ & 8.28 \\
        IgG-1D (SNR 0.001) & 128 & 31,338 & 256 & 36.51 & 0.64 & 9,738.0 & 0.90 & 2.36$\times$ & 9.65$\times$ & 6.19$\times$ & 8.94 \\
        IgG-RL & 128 & 11,836 & 256 & 34.33 & 0.46 & 24,235.4 & 0.37 & 2.27$\times$ & 7.96$\times$ & 5.79$\times$ & 9.56 \\
        Ribosembly & 128 & 33,033 & 512 & 73.89 & 1.21 & 12,287.0 & 1.88 & 2.46$\times$ & 11.37$\times$ & 6.48$\times$ & 8.97 \\
        Tomotwin-100 & 128 & 57,818 & 256 & 72.90 & 1.22 & 4,942.5 & 1.65 & 2.62$\times$ & 10.23$\times$ & 5.14$\times$ & 9.74 \\
        Spike-MD & 256 & 115,877 & 256 & 216.49 & 2.79 & 3,680.6 & 3.22 & 2.38$\times$ & 6.68$\times$ & 6.16$\times$ & 10.17 \\
        \bottomrule
    \end{tabular}%
    }
\end{table*}

\subsection{Baseline configurations}
\label{app:baseline_configurations}

The FSC AUC comparisons include CryoDRGN, CryoDRGN-AI-fixed, OPUS-DSD,
3DFlex, 3DVA, RECOVAR, 3D Classification, CryoSPIRE, CryoHype,
SOLVAR~\citep{yadgar2026solvar}, and CryoKRAQEN. Dataset coverage follows
Tables~\ref{tab:cryobench_per_conformation_fsc}
and~\ref{tab:cryobench_per_image_fsc} under the CryoBench evaluation
framework~\citep{jeon2024cryobench}. Methods with publicly available
implementations are run following CryoBench~\citep{jeon2024cryobench}. For
CryoSPIRE~\citep{shekarforoush2025cryospire} and
CryoKRAQEN~\citep{gao2026cryokraqen}, the included results are taken
directly from their publications, as we were unable to find a publicly
available implementation at the time of writing. For CryoHype~\citep{gu2026cryohype},
the included result is taken directly from its publication. Its public
release provides the configuration for the Tomotwin-100 experiment only,
and its IgG-1D result is obtained with an additional smoothness loss for
which we were unable to find a specification in the publication or a
configuration in the release. Obtaining results for CryoHype on
IgG-1D and the other datasets would require configurations and an implementation of
this term that the publication does not specify, so we do not include
them. CryoHype and CryoKRAQEN are compared under the per-image
protocol only (Table~\ref{tab:cryobench_per_image_fsc}). Dashes denote
results not included for a given dataset and protocol.

\subsection{Reconstruction methods in our comparisons}
\label{app:extended_related_work}

Our CryoBench comparisons use the same supplied ground-truth poses as fixed
inputs~\citep{jeon2024cryobench}. The compared methods represent structural
variability through discrete
states, continuous subspaces, deformation fields, or neural representations.
3D classification estimates a finite set of
volumes~\citep{punjani2017cryosparc,jeon2024cryobench}, whereas 3DVA learns
a linear subspace~\citep{punjani20213dva}. RECOVAR combines regularized
covariance estimation with adaptive kernel regression for per-state
reconstruction~\citep{gilles2025recovar}; SOLVAR directly optimizes a
low-rank covariance representation and uses RECOVAR's reconstruction
procedure with its estimated principal components~\citep{yadgar2026solvar}.
3DFlex learns
latent-conditioned deformation fields that transport a canonical
density~\citep{punjani20233dflex}. CryoDRGN learns image embeddings with
an encoder and a latent-conditioned neural field~\citep{zhong2021cryodrgn},
while CryoDRGN-AI-fixed directly optimizes particle-image
latents~\citep{levy2025cryodrgnai,jeon2024cryobench}.
OPUS-DSD uses a convolutional encoder--decoder with a structural
disentanglement prior~\citep{luo2023opusdsd}. CryoHype uses a transformer
hypernetwork to condition implicit neural representation weights on each
particle image~\citep{gu2026cryohype}, and CryoKRAQEN combines quantized
embeddings, kernel-guided assignment, and a triplane
representation~\citep{gao2026cryokraqen}. Among Gaussian approaches,
CryoSPIRE jointly optimizes particle-image latents and Gaussian features
in a model initialized from a reconstructed density. Feature clustering
identifies structural parts and anchors for a hierarchical representation
of part-level motion, local deformation, and occupancy
variation~\citep{shekarforoush2025cryospire}. CryoGaussian models
compositional and conformational heterogeneity through latent-conditioned
Gaussian attributes, starting from randomly initialized Gaussians.
Particle-image and per-Gaussian latents jointly condition this
representation, and each structural state is then reconstructed
conditioned on the inferred particle-image latents with a representation
optimized from randomly initialized Gaussians.

\section{Evaluation protocols}
\label{app:experimental_evaluation}

The synthetic-data evaluation distinguishes heterogeneity estimation from
per-state reconstruction, following the evaluation framework of
Jeon et~al.~\citep{jeon2024cryobench}. The evaluation of reconstructions from
the experimental datasets (\mbox{EMPIAR-10076} and \mbox{EMPIAR-10180}),
which provide no ground truth, is described in
Appendix~\ref{app:experimental_heterogeneity}.

\subsection{Heterogeneity estimation}
\label{app:latent_evaluation}

Neighborhood similarity (percentage of matching neighbors, pMN) and
information imbalance compare the inferred latents with the ground-truth
heterogeneity embedding of each particle image, as defined in
CryoBench~\citep{jeon2024cryobench}. These comparisons are made on all
five datasets with the embedding that CryoBench defines for each of them,
the sine and cosine of the dihedral angle for IgG-1D, the center of mass and
the sine and cosine of the dihedral angle for IgG-RL, the size rank of the
atomic models for Tomotwin-100 and for neighborhood similarity on
Ribosembly, the voxel intensities of the reference densities for
information imbalance on Ribosembly, and the two collective variables of
the receptor-binding domain for Spike-MD. Neighborhood similarity is
reported as the mean and standard deviation over the evaluation subsets of
CryoBench as a function of the neighborhood radius, over the range of radii
shown for each dataset in the CryoBench paper. Following the CryoBench
protocol,
information imbalance is reported as the mean and standard deviation over
$k=1$, 3, 10, and 30 nearest neighbors ($k=1$, 3, and 10 for Tomotwin-100).
As in CryoBench, it is also computed between the
inferred latents and the ground-truth rotation of each particle image,
represented by its flattened rotation matrix, and its CTF parameters,
represented by the two defocus values and the sine and cosine of the
astigmatism angle after standardization. Clustering accuracy on Ribosembly
and Tomotwin-100 follows CryoBench, and for CryoGaussian it is reported as
the mean and standard deviation over 10 random seeds.

\subsection{Per-image and per-conformation FSC evaluation}
\label{app:reference_fsc}

For density maps $u$ and $v$, Fourier shell correlation is
defined as~\citep{jeon2024cryobench}
\begin{equation}
\operatorname{FSC}_{u,v}(\nu_j)=
\frac{\operatorname{Re}\!\left[
\sum_{\mathbf{k}\in S_j}\widehat{u}(\mathbf{k})
\widehat{v}(\mathbf{k})^{*}\right]}
{\sqrt{\left(\sum_{\mathbf{k}\in S_j}|\widehat{u}(\mathbf{k})|^2\right)
\left(\sum_{\mathbf{k}\in S_j}|\widehat{v}(\mathbf{k})|^2\right)}}.
\label{eq:evaluation_fsc}
\end{equation}
Here, $S_j$ is the Fourier shell at radial spatial frequency $\nu_j$;
hats denote Fourier transforms, and $^{*}$ denotes complex conjugation.
For masked comparisons, the evaluation mask is applied to both maps before
the Fourier transform. Following CryoBench's per-image evaluation
implementation~\citep{jeon2024cryobench}, we compute FSC AUC by
trapezoidal integration:
\begin{equation}
\operatorname{FSC}_{\mathrm{AUC}}
=\sum_{j=0}^{J-1}\frac{c_j+c_{j+1}}{2}
(\nu_{j+1}-\nu_j),
\label{eq:evaluation_fsc_auc}
\end{equation}
where $\nu_0,\ldots,\nu_J$ are the sampled frequencies in cycles/voxel
and $c_j=|\operatorname{FSC}_{u,v}(\nu_j)|$. We retain this frequency scale
without rescaling AUC to a unit interval; its theoretical maximum over
the full frequency range $[0,1/2]$ is $0.5$~\citep{jeon2024cryobench}.

Per-image FSC compares a density reconstructed at an evaluated particle's
inferred latent with that particle's reference density. It therefore
retains the correspondence between an image and its underlying state.
Per-conformation FSC instead selects one representative latent for each
evaluated reference state: the latent nearest to the mean of the observed latents
assigned to that state is decoded and compared with the corresponding
reference density. These two evaluations differ in their sampling and reference
pairing~\citep{jeon2024cryobench}.

For CryoGaussian, densities are evaluated from the final Gaussian density
representation $\rho_{\Theta_2}$. Per-image and per-conformation FSC and their
AUC summaries follow the CryoBench evaluation protocol and
implementation~\citep{jeon2024cryobench}, with
background masking unless otherwise stated.
The density panels in
Figures~\ref{fig:igg1d_reconstruction}--\ref{fig:spikemd_reconstruction}
use inferred latents for particle images; the FSC panels report the two evaluations
separately. Solid curves show the arithmetic mean, and shaded regions cover
the 5th--95th percentiles across the evaluated images or reference states.
Figure~\ref{fig:cryobench_cluster_densities} instead selects an observed
latent nearest to each $k$-means cluster center, using 20
clusters except for Ribosembly, which uses 16. Colors identify cluster
membership within each dataset.
For Spike-MD, we generate reference densities from the corresponding
molecular-dynamics structures using ChimeraX 1.6.1 \texttt{molmap},
following the dataset-generation protocol of Jeon et~al.~\citep{jeon2024cryobench}.
For this dataset, the reference states are the distinct
molecular-dynamics configurations, each shared by approximately two
particle images~\citep{jeon2024cryobench}, and the per-conformation
evaluation selects one representative latent per configuration as for the
other datasets.

\subsection{Evaluation on experimental data}
\label{app:experimental_heterogeneity}

\paragraph{Assembly-state selection and comparison.}
We train CryoGaussian on the full EMPIAR-10076 particle stack for
particle selection (Figure~\ref{fig:empiar10076_filtering};
Appendix~\ref{app:particle_selection}). Assembly-state comparisons use
both the resulting CryoGaussian-filtered subset and the CryoDRGN-filtered
subset. Published class assignments~\citep{davis2016assembly}
identify the particle images associated with each reference state. We decode
the observed latent nearest to the class-mean latent to obtain a representative
density, following the selection procedure of Zhong et~al.~\citep{zhong2021cryodrgn}.
Map-to-map FSC compares these densities with the corresponding deposited
reconstructions. The published assignments are used to select and evaluate
representative densities after training.
For the assembly intermediate with H68 density
but no resolved central protuberance (class C4)~\citep{zhong2021cryodrgn}, the
CryoDRGN-filtered reconstruction uses a single representative latent:
the observed particle latent nearest to the mean inferred latent of the
1,113 particles identified by the released CryoDRGN state indices.
We decode this latent using the same representative-latent convention
and density rendering pipeline as the other assembly classes.
We additionally perform homogeneous refinement in CryoSPARC. Panel (a) of
Figure~\ref{fig:empiar10076_c4_comparison} uses all 1,113 particles
identified by the released CryoDRGN state indices; panel (b) uses their
intersection with the 97,651-particle set retained by CryoGaussian
filtering, yielding 1,019 particles. This intersection uses the full
filtered set before the random 90\% training subsampling.
The identification of this intermediate does not require the released state
indices. On the UMAP projection of the inferred latents, colored by published
major class assignments, we manually outline, at the boundary of the class C
cluster, a group of particles that includes particles assigned to class E by
the published classification (panel (c)). The outlined regions contain 1,062
particles for the CryoDRGN-filtered subset and 954 for the full
CryoGaussian-filtered set. Panels (d) and (e) show the CryoGaussian
densities decoded at the observed latent nearest to the mean inferred latent
of the selected particles, together with the CryoSPARC refinement of these
particles. Gold-standard
FSC assesses consistency between the independent CryoSPARC half-maps,
while map-to-map FSC assesses agreement between the CryoGaussian and
CryoSPARC densities.

\paragraph{Spliceosome latent selection.}
For the main EMPIAR-10180 reconstruction using the CryoDRGN-filtered
subset, we apply principal component analysis to the inferred latents and
decode the five positions marked along each of the first two principal
components in Figure~\ref{fig:empiar10180_main}. Separately, we train
CryoGaussian on the full particle stack
(Figure~\ref{fig:empiar10180_full_dataset}), fit 20 $k$-means clusters to
the inferred latents, and subdivide the largest cluster into two groups.
Representative densities are decoded at the observed latents nearest to
the cluster centers.

\section{Additional reconstruction results}
\label{app:additional_results}

\subsection{CryoBench reconstruction}
\label{app:additional_cryobench}

\begin{figure*}[tp]
    \centering
    \includegraphics[width=0.9\linewidth]{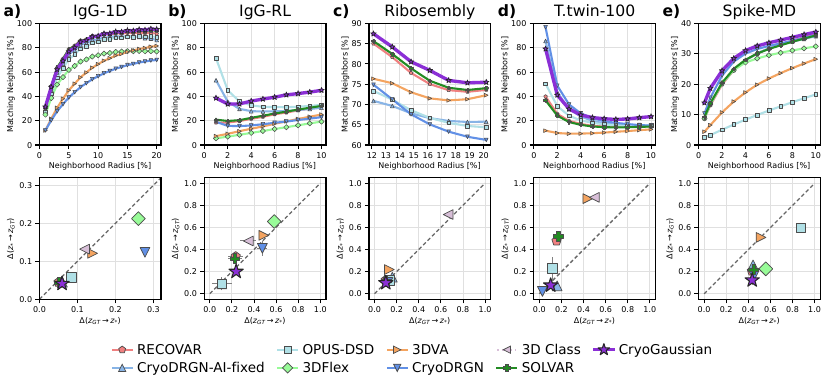}
    \caption{\textbf{Heterogeneity estimation on CryoBench.}
    Neighborhood similarity as a function of the neighborhood radius (top
    row) and information imbalance (bottom row) between inferred latents
    and the ground-truth heterogeneity embedding on \textbf{(a)} IgG-1D,
    \textbf{(b)} IgG-RL, \textbf{(c)} Ribosembly, \textbf{(d)}
    Tomotwin-100, and \textbf{(e)} Spike-MD, presented in the format of
    Figures~2d, 4d, 6d, 7d, and S24 of the CryoBench
    paper~\citep{jeon2024cryobench}. Stars
    denote CryoGaussian.}
    \label{fig:heterogeneity_latent_metrics}
\end{figure*}

\begin{figure*}[tp]
    \centering
    \includegraphics[width=0.9\linewidth]{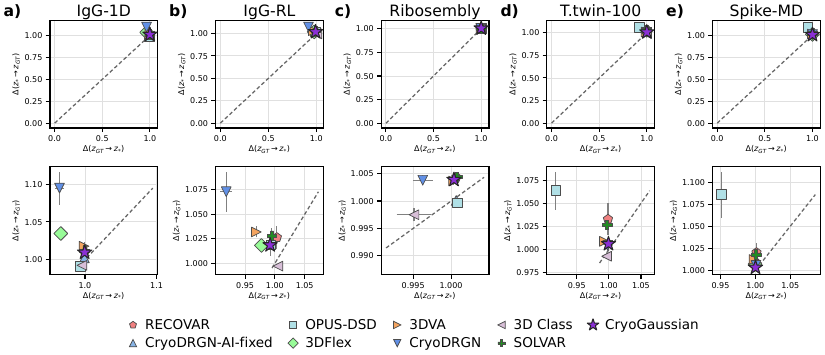}
    \caption{\textbf{Pose information imbalance on CryoBench.}
    Information imbalance between inferred latents and ground-truth
    rotations on \textbf{(a)} IgG-1D, \textbf{(b)} IgG-RL, \textbf{(c)}
    Ribosembly, \textbf{(d)} Tomotwin-100, and \textbf{(e)} Spike-MD, in
    full view (top row) and enlarged near $(1,1)$ (bottom row), presented in
    the format of Figure~S22 of the CryoBench
    paper~\citep{jeon2024cryobench}.}
    \label{fig:pose_information_imbalance}
\end{figure*}

\begin{figure*}[tp]
    \centering
    \includegraphics[width=0.9\linewidth]{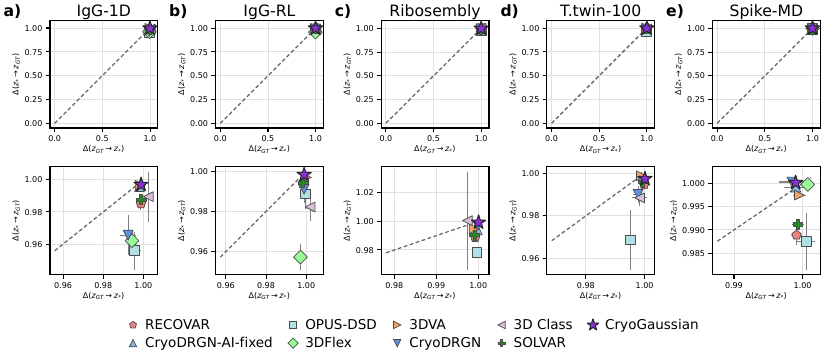}
    \caption{\textbf{CTF information imbalance on CryoBench.}
    Information imbalance between inferred latents and CTF parameters on
    \textbf{(a)} IgG-1D, \textbf{(b)} IgG-RL, \textbf{(c)} Ribosembly,
    \textbf{(d)} Tomotwin-100, and \textbf{(e)} Spike-MD, in full view
    (top row) and enlarged near $(1,1)$ (bottom row), presented in the
    format of Figure~S23 of the CryoBench paper~\citep{jeon2024cryobench}.}
    \label{fig:ctf_information_imbalance}
\end{figure*}

Table~\ref{tab:cryobench_per_conformation_fsc} complements the per-image
comparison by evaluating one representative latent for each reference state.
For CryoGaussian, mean per-conformation FSC AUC is 0.402 on IgG-1D, 0.387
on IgG-RL, 0.432 on Ribosembly, 0.336 on Tomotwin-100, and 0.365 on
Spike-MD.
Table~\ref{tab:clustering_accuracy} reports clustering accuracy of the
inferred latents on Ribosembly and Tomotwin-100, and
Table~\ref{tab:seed_variability} reports the variability of the mean FSC
AUC across training seeds.
Figure~\ref{fig:heterogeneity_latent_metrics} reports neighborhood
similarity and information imbalance between the inferred latents and the
ground-truth heterogeneity embedding on all five datasets for CryoGaussian
and the compared methods, computed and presented following CryoBench
(Appendix~\ref{app:latent_evaluation}). In CryoBench, neighborhood
similarity at the smallest radius reflects how well images of the same
structure are grouped and at larger radii how the relations between
structures are preserved, and information imbalance near $(0,0)$ indicates
equivalent information, near $(1,1)$ orthogonal information, and below the
diagonal that the inferred latents predict the ground-truth neighborhoods
more closely than the reverse~\citep{jeon2024cryobench}. The neighborhood
similarity of CryoGaussian is comparable to or higher than that of the
compared methods over most of the evaluated radii and the highest at the
largest radius on all five datasets, where it is 95.1\% on IgG-1D, 45.0\%
on IgG-RL, 75.5\% on Ribosembly, 23.4\% on Tomotwin-100, and 37.2\% on
Spike-MD. Its information imbalance is $(0.059, 0.041)$ on IgG-1D, $(0.238, 0.201)$ on IgG-RL,
$(0.101, 0.098)$ on Ribosembly, $(0.102, 0.076)$ on Tomotwin-100, and
$(0.435, 0.123)$ on Spike-MD, in the order
$(\Delta(z_{\mathrm{GT}}\rightarrow z_{*}),
\Delta(z_{*}\rightarrow z_{\mathrm{GT}}))$, close to $(0,0)$ on IgG-1D,
Ribosembly, and Tomotwin-100 and below the diagonal on IgG-RL and Spike-MD.
Figures~\ref{fig:pose_information_imbalance}
and~\ref{fig:ctf_information_imbalance} report the information imbalance
between the inferred latents and the ground-truth rotation and CTF
parameters of each particle image for CryoGaussian and the compared
methods, computed with the pose and CTF embeddings, neighborhood sizes, and
evaluation subset of CryoBench (Appendix~\ref{app:latent_evaluation}) and
presented in the format of Figures~S22 and S23 of the CryoBench
paper~\citep{jeon2024cryobench}.
$\Delta(z_{\mathrm{GT}}\rightarrow z_{*})$ measures how well neighborhoods
in the ground-truth rotation or CTF parameters predict neighborhoods in the
latent space, $\Delta(z_{*}\rightarrow z_{\mathrm{GT}})$ the reverse, and
values near 1 in both directions correspond to orthogonal information as
defined in CryoBench, where neither predicts the other. For CryoGaussian,
the information imbalance with the ground-truth rotations is
$(0.9995, 1.0097)$ on IgG-1D, $(0.9919, 1.0185)$ on IgG-RL,
$(1.0004, 1.0039)$ on Ribosembly, $(0.9993, 1.0061)$ on Tomotwin-100, and
$(1.0000, 1.0029)$ on Spike-MD, and with the CTF parameters
$(0.9988, 0.9970)$, $(0.9989, 0.9982)$, $(1.0001, 0.9993)$,
$(1.0001, 0.9973)$, and $(0.9990, 1.0001)$ on the same datasets, in the
order $(\Delta(z_{\mathrm{GT}}\rightarrow z_{*}),
\Delta(z_{*}\rightarrow z_{\mathrm{GT}}))$. As in CryoBench, each value is
the mean of the information imbalance over the neighborhood sizes $k$, and
its standard deviation over these sizes, shown as error bars in the
figures, is at most 0.004 for CryoGaussian. The values for the compared
methods lie between 0.92 and 1.09 for rotations and between 0.96 and 1.00
for CTF parameters, so that the latents of all methods are orthogonal to
rotation and CTF parameters in this sense, as observed in CryoBench.

\begin{figure*}[p]
    \centering
    \includegraphics[width=\linewidth, height=\dimexpr\textheight-11\baselineskip\relax, keepaspectratio]{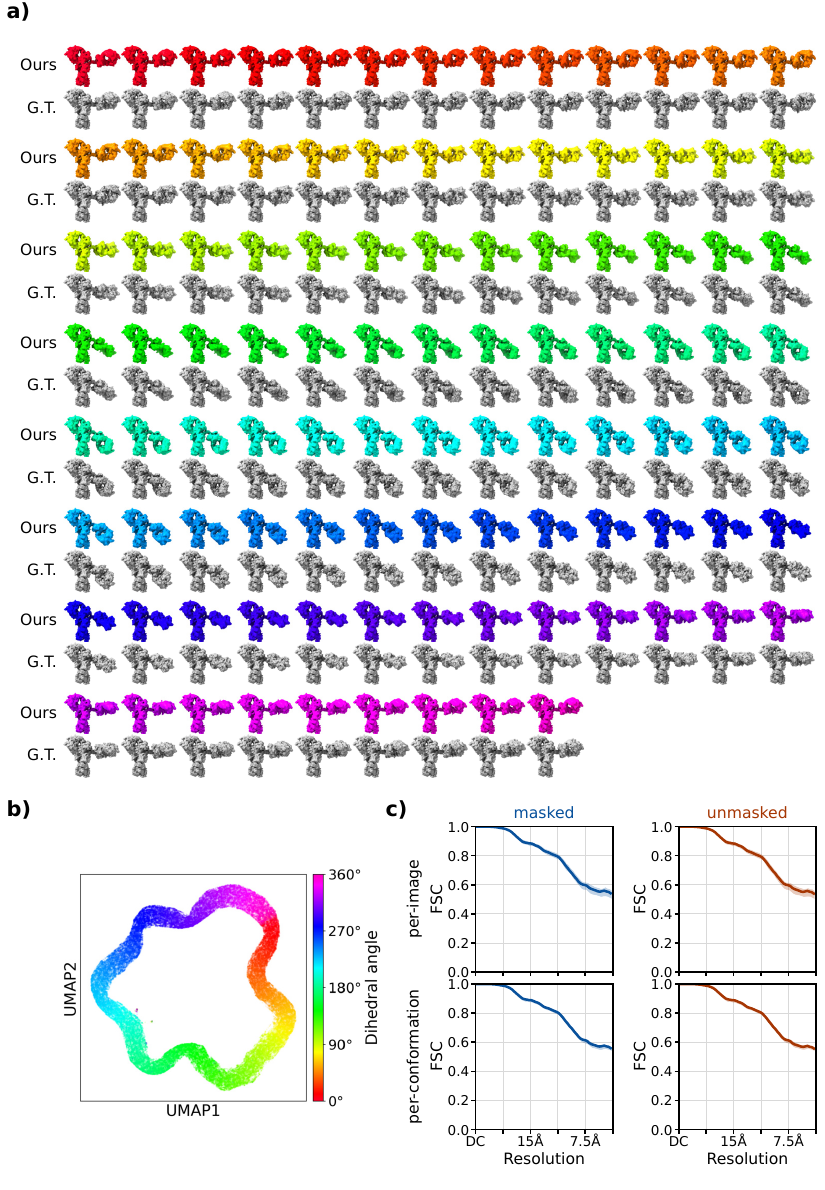}
    \caption{\textbf{IgG-1D reconstruction at SNR 0.01.} \textbf{(a)} CryoGaussian densities
reconstructed at the inferred latents for 100 particle images (Ours), each paired with
its reference density (G.T., gray). Reconstructed densities are colored by the
ground-truth dihedral angle. \textbf{(b)} UMAP projection of inferred latents with the
same color scale. \textbf{(c)} Per-image FSC (top) and per-conformation FSC (bottom),
evaluated with background masking (left) and without masking (right). Solid curves
show the mean, and shaded regions cover the 5th--95th percentiles across 100 evaluated
particle images or 100 reference states, respectively.}
    \label{fig:igg1d_reconstruction}
\end{figure*}

\begin{figure*}[p]
    \centering
    \includegraphics[width=\linewidth, height=\dimexpr\textheight-11\baselineskip\relax, keepaspectratio]{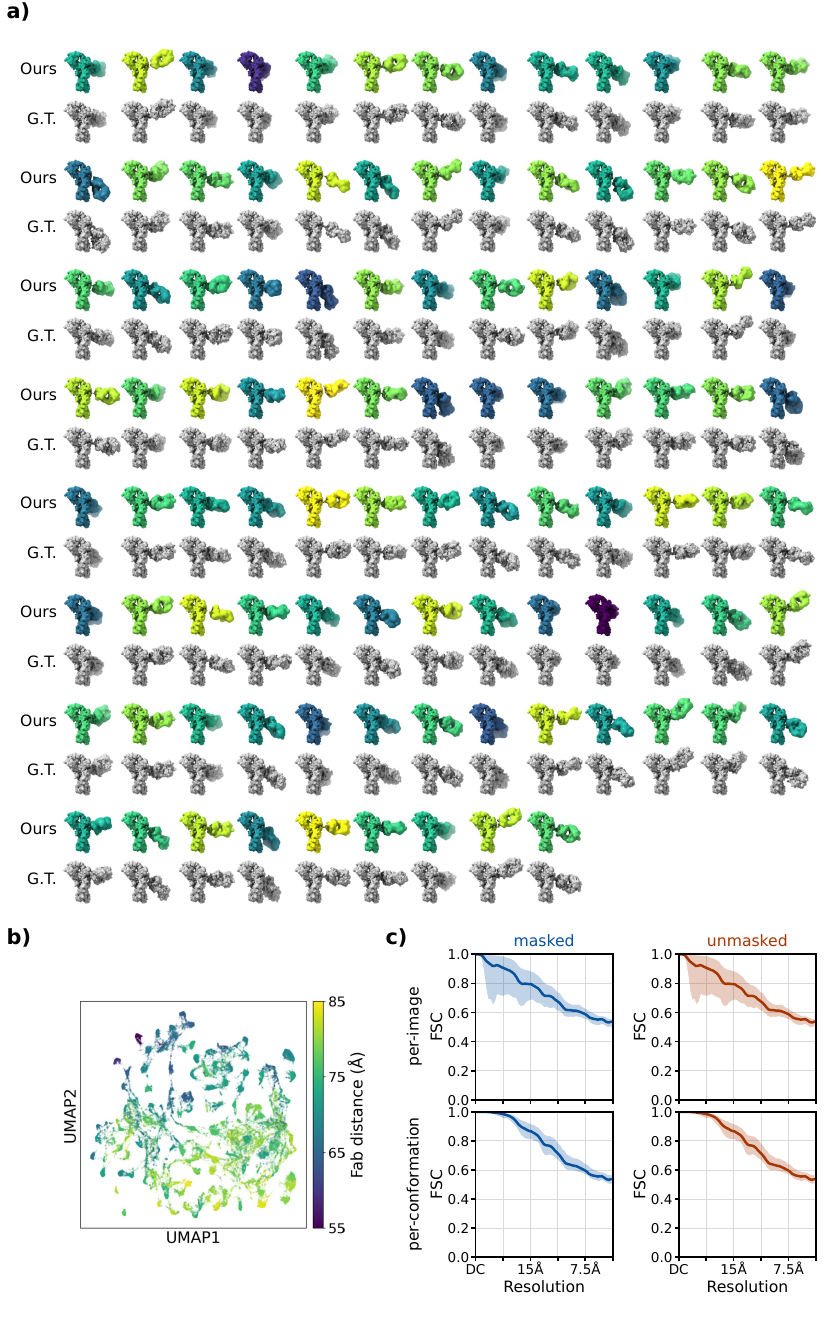}
    \caption{\textbf{IgG-RL reconstruction.} \textbf{(a)} CryoGaussian densities reconstructed at
the inferred latents for 100 particle images (Ours), paired with their reference
densities (G.T., gray). Colors indicate the distance from the Fab domain to the center
of mass of the rest of the structure. \textbf{(b)} UMAP projection of inferred latents
with the same color scale. \textbf{(c)} Per-image FSC (top) and per-conformation FSC
(bottom), with background masking (left) and without masking (right). Solid curves
show the mean, and shaded regions cover the 5th--95th percentiles across 100 evaluated
particle images or 100 reference states, respectively.}
    \label{fig:iggrl_reconstruction}
\end{figure*}

\begin{figure*}[p]
    \centering
    \includegraphics[width=\linewidth, height=\dimexpr\textheight-12\baselineskip\relax, keepaspectratio]{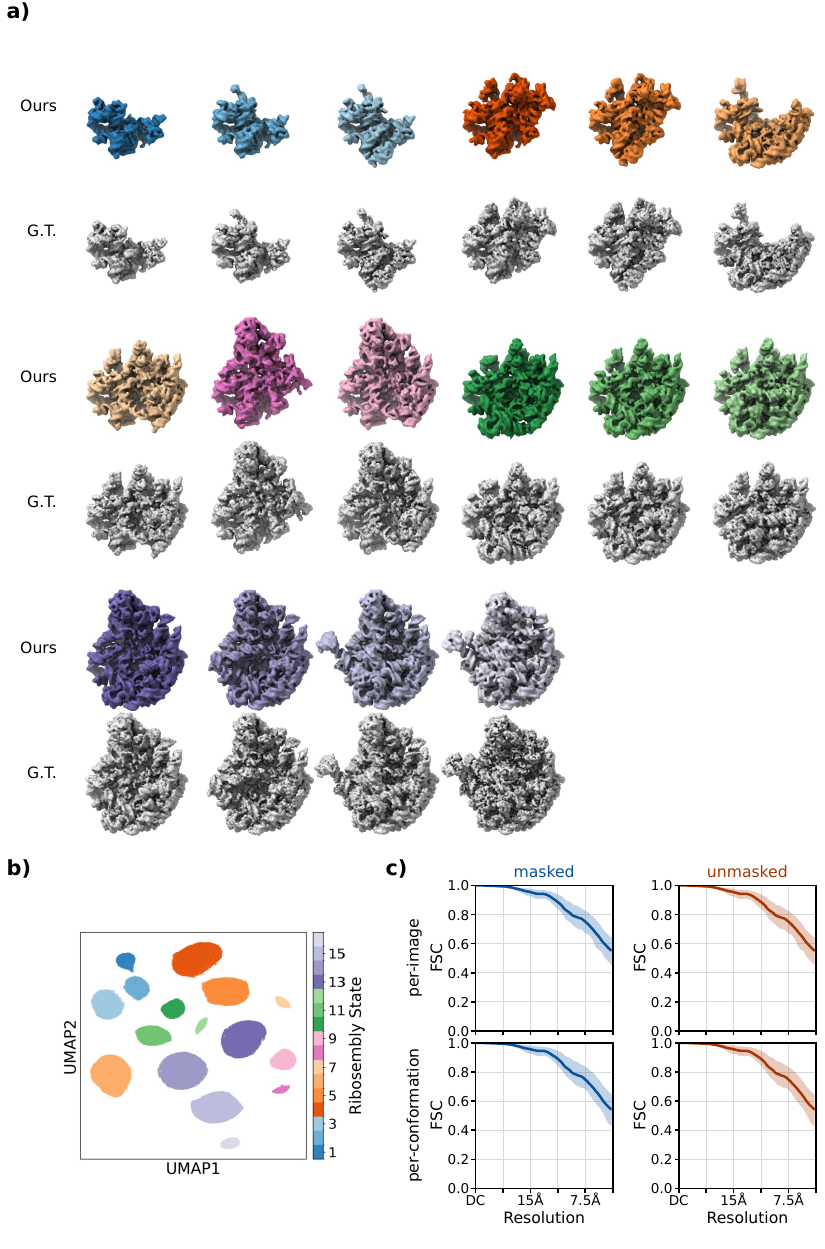}
    \caption{\textbf{Ribosembly reconstruction.} \textbf{(a)} CryoGaussian densities reconstructed
at inferred latents for particle images from all 16 reference assembly states
(Ours), paired with their reference densities (G.T., gray). \textbf{(b)} UMAP
projection of inferred latents. Colors in (a) and (b) identify reference assembly
states. \textbf{(c)} Per-image FSC
(top) and per-conformation FSC (bottom), with background masking (left) and without
masking (right). Solid curves show the mean, and shaded regions cover the 5th--95th
percentiles across 101 evaluated particle images in the top row and 16 reference
states in the bottom row.}
    \label{fig:ribosembly_reconstruction}
\end{figure*}

\begin{figure*}[p]
    \centering
    \includegraphics[width=\linewidth, height=\dimexpr\textheight-11\baselineskip\relax, keepaspectratio]{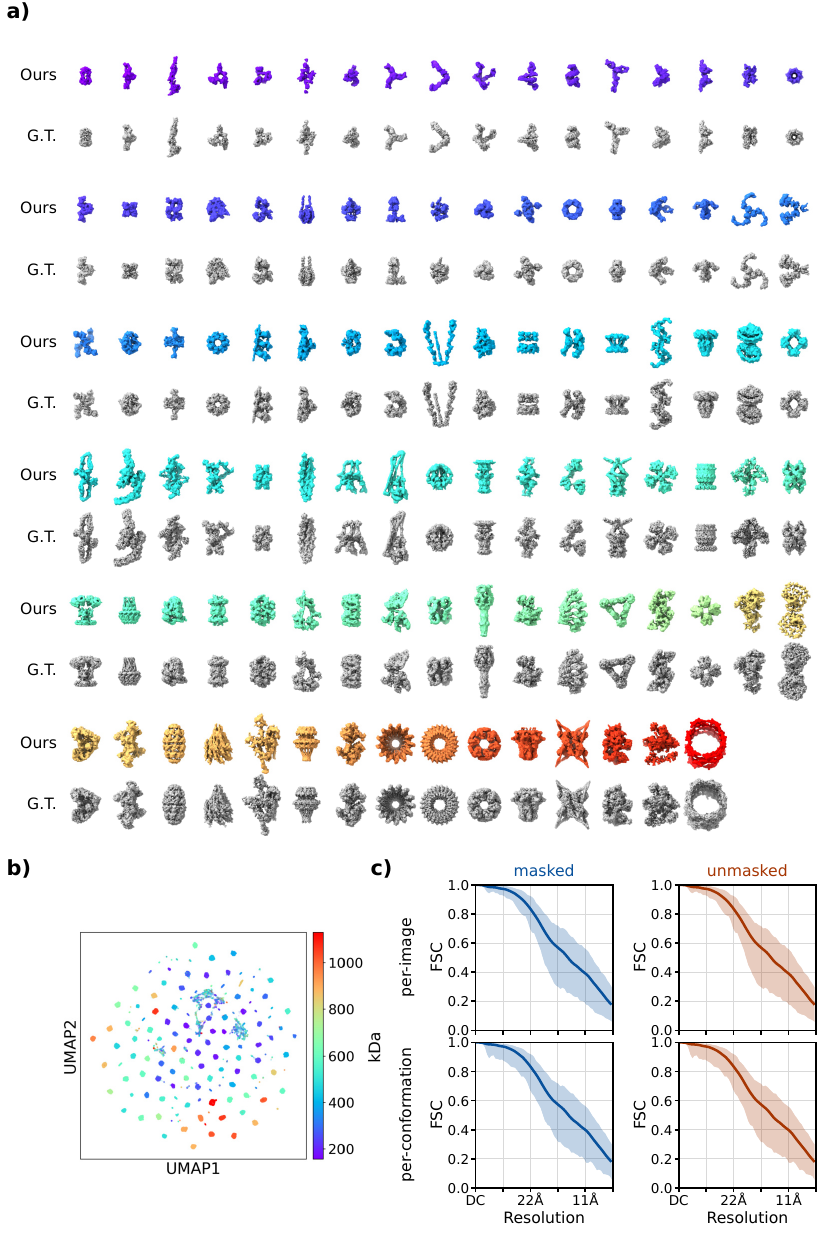}
    \caption{\textbf{Tomotwin-100 reconstruction.} \textbf{(a)} Per-image CryoGaussian densities
(Ours) and paired reference densities (G.T., gray) for 100 distinct complexes.
\textbf{(b)} UMAP projection of inferred latents;
colors in (a) and (b) indicate molecular mass in kDa. \textbf{(c)} Per-image FSC (top)
and per-conformation FSC (bottom), with background masking (left) and without masking
(right). Solid curves show the mean, and shaded regions cover the 5th--95th percentiles
across 100 evaluated particle images or 100 reference complexes, respectively.}
    \label{fig:tomotwin_reconstruction}
\end{figure*}

\begin{figure*}[p]
    \centering
    \includegraphics[width=\linewidth, height=\dimexpr\textheight-11\baselineskip\relax, keepaspectratio]{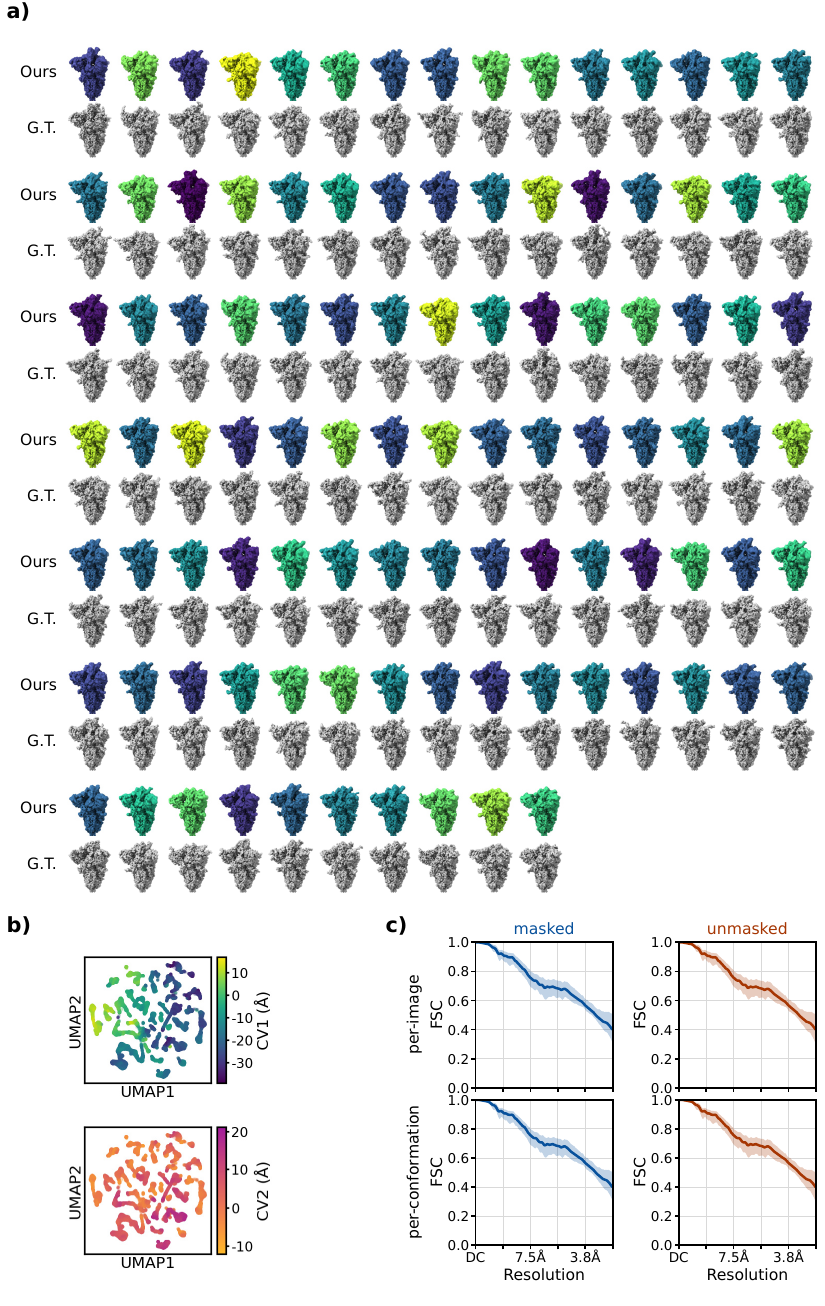}
    \caption{\textbf{Spike-MD reconstruction.} \textbf{(a)} CryoGaussian densities reconstructed at
the inferred latents for 100 selected particle images (Ours), paired with their
reference densities (G.T., gray) and colored by collective variable CV1. \textbf{(b)}
UMAP projections of inferred latents, colored by CV1 (top) and CV2 (bottom), with
collective-variable definitions from Wiecz{\'o}r et~al.~\citep{wieczor2023spike}. \textbf{(c)} Per-image FSC (top)
and per-conformation FSC (bottom), with background masking (left) and without
masking (right). Solid curves show the mean, and shaded regions cover the
5th--95th percentiles across 100 evaluated particle images or their reference
configurations, respectively.}
    \label{fig:spikemd_reconstruction}
\end{figure*}

\begin{figure*}[p]
    \centering
    \includegraphics[width=\linewidth, height=\dimexpr\textheight-10\baselineskip\relax, keepaspectratio]{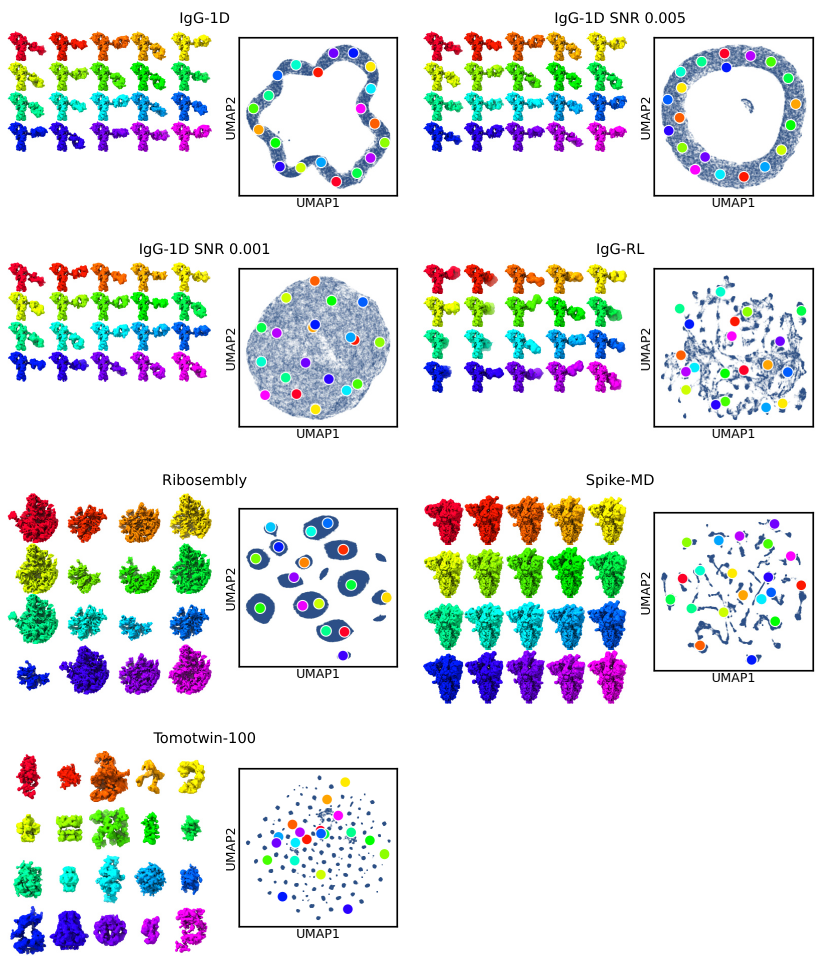}
    \caption{\textbf{Representative densities selected by latent clustering.} For each dataset and
noise condition, inferred latents are partitioned by $k$-means into 20 clusters, or 16
for Ribosembly. The observed latent nearest to each cluster center is decoded to
obtain a representative density. Each dataset block shows these densities alongside
the UMAP projection, with colored markers locating the selected latents. Matching
colors link each density to its marker and indicate cluster identity within that
dataset.}
    \label{fig:cryobench_cluster_densities}
\end{figure*}

Figures~\ref{fig:igg1d_reconstruction}--\ref{fig:spikemd_reconstruction}
pair per-image reconstructed densities with their corresponding reference
densities and show masked and unmasked FSC curves. The paired displays
relate per-state reconstruction to the reference heterogeneity represented
by the inferred latents. Figure~\ref{fig:cryobench_cluster_densities}
provides a complementary selection based on latent clustering, with each
representative density decoded at an observed latent nearest to a cluster
center. Lower-SNR IgG-1D results are discussed separately in
Appendix~\ref{app:noise_sensitivity}.

\begin{table*}[tp]
    \centering
    \caption{\textbf{Clustering accuracy on Ribosembly and Tomotwin-100.}
Adjusted Rand index (ARI) and adjusted mutual information (AMI) between
ground-truth labels and $k$-means cluster assignments of the inferred
particle-image latents, with $k$ equal to the number of ground-truth states
(16 for Ribosembly and 100 for Tomotwin-100). Higher is better. Bold and
underline mark the best and second best per column. The compared methods
are those of Tables~\ref{tab:cryobench_per_conformation_fsc}
and~\ref{tab:cryobench_per_image_fsc} with publicly available
implementations. CryoGaussian values are the mean (standard deviation) over
10 random seeds.}
    \label{tab:clustering_accuracy}
    \setlength{\tabcolsep}{4pt}
    \setlength{\aboverulesep}{0pt}
    \setlength{\belowrulesep}{0pt}
    \renewcommand{\arraystretch}{1.3}
    \small
    {%
        \begin{tabular}{@{}lcccc@{}}
            \toprule
            Method
            & \multicolumn{2}{c}{Ribosembly}
            & \multicolumn{2}{c}{Tomotwin-100} \\
            \cmidrule(lr){2-3}
            \cmidrule(lr){4-5}
            & ARI & AMI & ARI & AMI \\
            \midrule
            CryoDRGN & 0.873 & 0.935 & \textbf{0.956} & \textbf{0.983} \\
            CryoDRGN-AI-fixed & 0.624 & 0.771 & 0.791 & 0.906 \\
            OPUS-DSD & 0.891 & 0.934 & 0.500 & 0.781 \\
            3DVA & 0.666 & 0.823 & 0.058 & 0.335 \\
            RECOVAR & 0.968 & 0.976 & 0.315 & 0.649 \\
            SOLVAR & \underline{0.971} & \underline{0.979} & 0.280 & 0.625 \\
            \midrule
            \rowcolor{gray!15}
            CryoGaussian (Ours)
            & \textbf{0.974 (0.004)} & \textbf{0.982 (0.001)}
            & \underline{0.823 (0.005)} & \underline{0.926 (0.001)} \\
            \bottomrule
        \end{tabular}%
    }
\end{table*}

\begin{table*}[tp]
    \centering
    \caption{\textbf{Variability of CryoGaussian across training seeds on CryoBench.}
Mean $\pm$ standard deviation across 10 training seeds of the mean FSC AUC
with background masking, with the range across seeds in parentheses. Seeds
vary the random initialization of the Gaussians and latents and the order
of particle images, and heterogeneity estimation and per-state
reconstruction are both rerun for each seed.}
    \label{tab:seed_variability}
    \setlength{\tabcolsep}{6pt}
    \setlength{\aboverulesep}{0pt}
    \setlength{\belowrulesep}{0pt}
    \renewcommand{\arraystretch}{1.3}
    \small
    {%
        \begin{tabular}{@{}lcc@{}}
            \toprule
            Dataset & Per-image mean FSC AUC & Per-conformation mean FSC AUC \\
            \midrule
            IgG-1D & 0.3976 $\pm$ 0.0006 (0.3962--0.3982) & 0.4024 $\pm$ 0.0004 (0.4019--0.4032) \\
            IgG-RL & 0.3673 $\pm$ 0.0011 (0.3656--0.3685) & 0.3879 $\pm$ 0.0009 (0.3861--0.3892) \\
            Ribosembly & 0.4326 $\pm$ 0.0007 (0.4321--0.4339) & 0.4328 $\pm$ 0.0008 (0.4313--0.4338) \\
            Tomotwin-100 & 0.3366 $\pm$ 0.0012 (0.3353--0.3387) & 0.3373 $\pm$ 0.0012 (0.3359--0.3392) \\
            Spike-MD & 0.3647 $\pm$ 0.0008 (0.3634--0.3654) & 0.3651 $\pm$ 0.0008 (0.3640--0.3669) \\
            \bottomrule
        \end{tabular}%
    }
\end{table*}

\subsection{Heterogeneous reconstruction on experimental data}
\label{app:additional_experimental_heterogeneity}

\begin{figure*}[p]
    \centering
    \includegraphics[width=\linewidth, height=\dimexpr\textheight-12\baselineskip\relax, keepaspectratio]{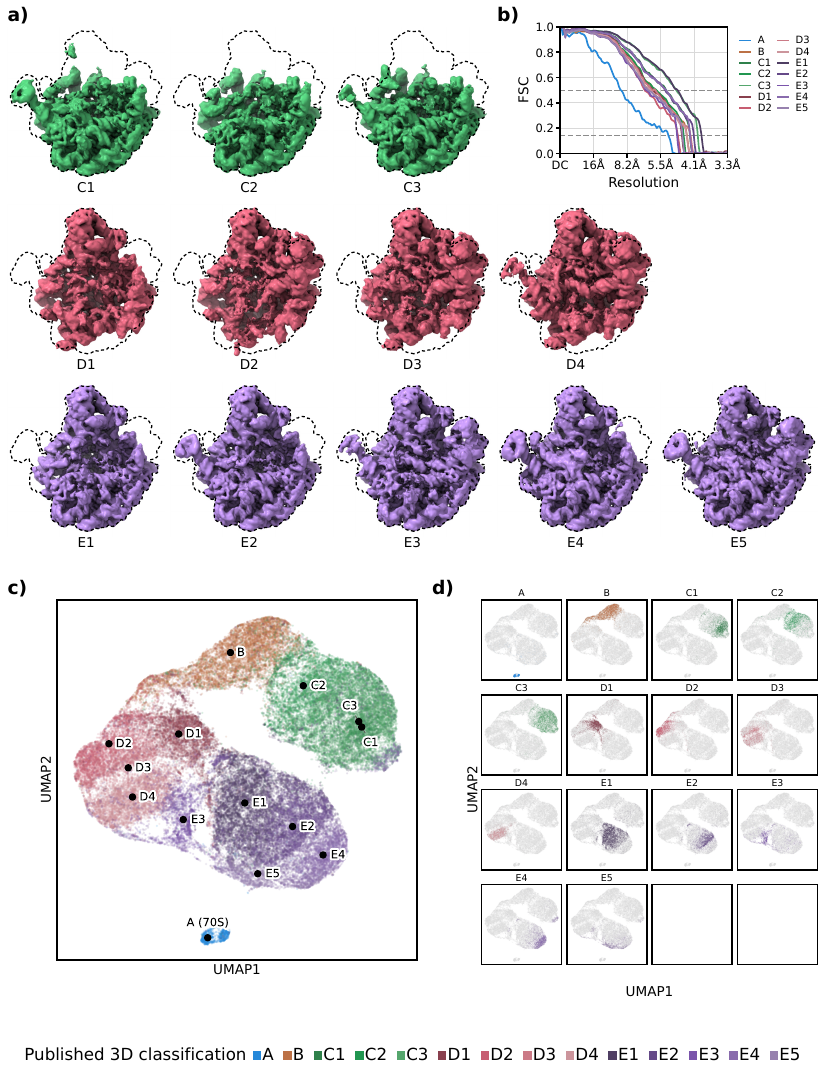}
    \caption{\textbf{Assembly-state reconstruction using the CryoDRGN particle selection for
EMPIAR-10076.} The model is trained on 87,327 particle images. \textbf{(a)}
Reconstructed densities for the twelve minor classes C1--C3, D1--D4, and E1--E5, with
dashed lines tracing the outline of the mature 50S subunit. \textbf{(b)} Map-to-map FSC
between CryoGaussian densities and the corresponding deposited
reconstructions~\citep{davis2016assembly}, for class A and the thirteen assembly
classes B--E5. Horizontal dashed lines indicate FSC values of 0.5 and 0.143.
\textbf{(c)} UMAP projection of inferred latents colored by published class
assignments. \textbf{(d)} The same projection with each class highlighted separately.
Colors are shared across panels (b)--(d). Representative-latent selection and
map comparison are described in
Appendix~\ref{app:experimental_heterogeneity}.}
    \label{fig:empiar10076_supplied_states}
\end{figure*}

\paragraph{Assembling 50S ribosomal subunit.}
Figure~\ref{fig:empiar10076_supplied_states} shows the reconstructed minor
assembly states and their map-to-map FSC against deposited densities using
the CryoDRGN-filtered particle subset. Representative latents are selected
as described in Appendix~\ref{app:experimental_heterogeneity}.
Figure~\ref{fig:empiar10076_c4_comparison} compares the intermediate with H68
density but no resolved central protuberance with homogeneous CryoSPARC
reconstructions using the released CryoDRGN state selection (a), its
intersection with the CryoGaussian-filtered particle set (b), and particles
selected in the CryoGaussian latent space (c--e). The intermediate is
identified in this latent space without the released state indices:
refinement of the selected particles also shows H68 density and no resolved
central protuberance.
The separate particle-selection analysis is described in
Appendix~\ref{app:particle_selection}.

\begin{figure*}[p]
    \centering
    \includegraphics[width=\linewidth, height=\dimexpr\textheight-14\baselineskip\relax, keepaspectratio]{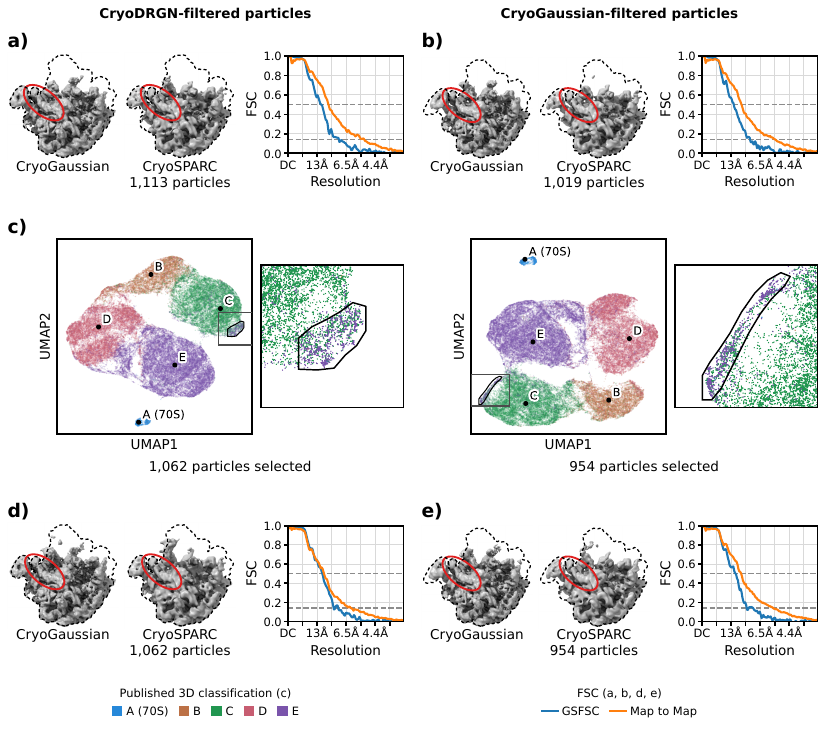}
    \caption{\textbf{Identification and homogeneous refinement of the 50S
assembly intermediate C4.} The left and right columns use the
CryoDRGN-filtered and CryoGaussian-filtered particles, respectively.
\textbf{(a, b)} CryoSPARC refinement uses the 1,113 particles of the
released C4 particle selection~\citep{zhong2021cryodrgn} (a) and their
intersection with the CryoGaussian-filtered particle set (1,019 particles; b).
\textbf{(c)} Identification of this intermediate in the CryoGaussian latent
space, without the released selection. UMAP projections of the inferred
latents are colored by published major class assignments; black points mark
class centers. Gray boxes mark the regions enlarged on the right, and black
polygons outline the selected particles. \textbf{(d, e)} CryoSPARC
refinement uses the 1,062 (d) and 954 (e) particles selected in (c).
Panels (a), (b), (d), and (e) each show the CryoGaussian density, the
CryoSPARC density obtained by homogeneous refinement of the corresponding
selected particle images, and FSC curves. Blue curves show gold-standard FSC
between independent CryoSPARC half-maps; orange curves show map-to-map
FSC between the displayed CryoGaussian and CryoSPARC densities.
Horizontal dashed lines indicate FSC values of 0.5 and 0.143. Dashed
lines trace the outline of the mature 50S subunit, and red ellipses mark rRNA
helix 68 (H68).}
    \label{fig:empiar10076_c4_comparison}
\end{figure*}

\paragraph{Pre-catalytic spliceosome.}
The principal-component traversals in Figure~\ref{fig:empiar10180_main}
describe conformational variation within the CryoDRGN-filtered subset.
Figure~\ref{fig:empiar10180_full_dataset} separately examines the full
particle stack using the preprocessing in Appendix~\ref{app:experimental_data}.
Its cluster-selected
densities include differences in SF3b density and additional density near
the U2 core.

\begin{figure*}[p]
    \centering
    \includegraphics[width=\linewidth, height=\dimexpr\textheight-15\baselineskip\relax, keepaspectratio]{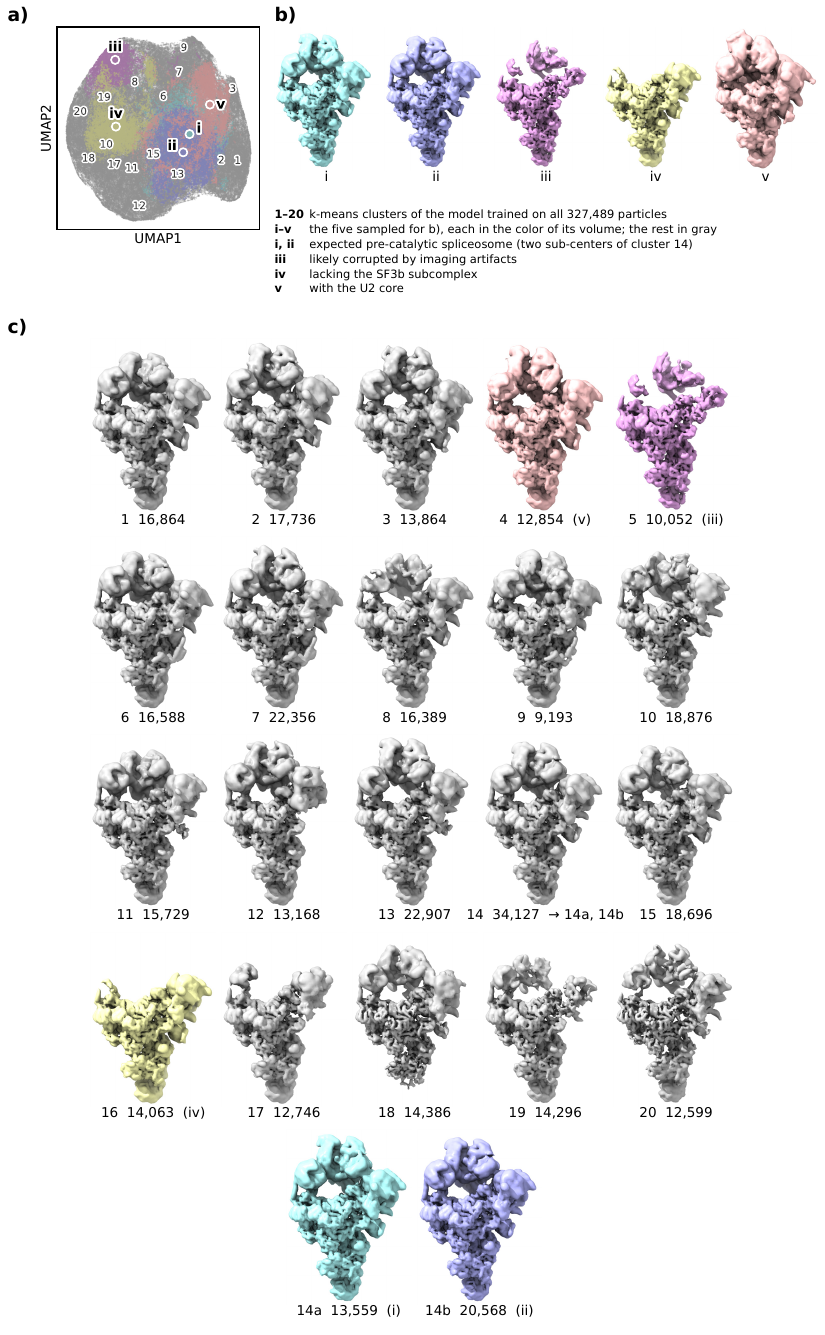}
    \caption{\textbf{Latent-space analysis of the full EMPIAR-10180 particle stack.} A
four-dimensional model is trained on 327,489 particle images at $128\times128$ pixels.
\textbf{(a)} UMAP projection of inferred latents, with 20 $k$-means clusters numbered
1--20. Five selected groups are colored to match the reconstructed densities in (b);
the other groups are gray. \textbf{(b)} Representative densities from two
subdivisions of the largest cluster (i, ii), a cluster with poorly resolved density (iii), a
cluster lacking density in the SF3b region (iv), and a cluster with additional density
near the U2 core (v). \textbf{(c)} Representative densities for all 20 clusters and
the two subdivisions 14a and 14b, with particle counts below each density.
Representative-latent selection is described in
Appendix~\ref{app:experimental_heterogeneity}. This full-stack analysis is separate
from the CryoDRGN-filtered particle subset used in
Figure~\ref{fig:empiar10180_main}.}
    \label{fig:empiar10180_full_dataset}
\end{figure*}

\clearpage
\section{Ablations and robustness}
\label{app:sensitivity_robustness}

\subsection{Number of Gaussians and relative density threshold}
\label{app:count_sensitivity}

\begin{figure*}[tp]
    \centering
    \includegraphics[width=0.75\linewidth]{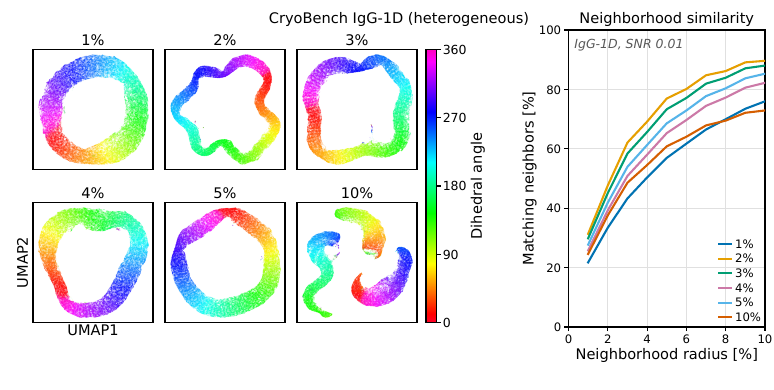}
    \caption{\textbf{Sensitivity to the relative density threshold on IgG-1D.}
    Heterogeneity estimation on IgG-1D at SNR 0.01 with
    $\tau\in\{0.01,0.02,0.03,0.04,0.05,0.10\}$ for estimating the number of
    Gaussians.
    UMAP projections of inferred latents, colored by the ground-truth
    dihedral angle, are shown on the left; corresponding
    neighborhood-similarity curves are shown on the right, following
    Appendix~\ref{app:latent_evaluation}.
    The default threshold $\tau=0.02$ yields the highest neighborhood
    similarity among the tested values. The UMAP projections retain
    an ordered loop for thresholds from 0.01 to 0.05, whereas
    $\tau=0.10$ produces a fragmented projection and lower
    neighborhood similarity.}
    \label{fig:gaussian_count_sensitivity}
\end{figure*}

The number of Gaussians is determined by the estimated spatial support
through Equation~\eqref{eq:gaussian_count}, and the support by the
backprojection and the relative density threshold $\tau$ through
Equation~\eqref{eq:estimated_support}. We vary $\tau$ over
$\{0.01,0.02,0.03,0.04,0.05,0.10\}$ and let the number of Gaussians follow.
Raising $\tau$ removes the voxels of lowest relative density from the
estimated support (Equation~\eqref{eq:threshold_count_difference}), and
across the five CryoBench datasets this range of $\tau$ changes the number
of Gaussians from 0.33 to 1.74 times its value at the default threshold.
Figure~\ref{fig:gaussian_count_sensitivity} evaluates heterogeneity
estimation on IgG-1D at SNR 0.01, and Table~\ref{tab:tau_sensitivity}
evaluates per-image and per-conformation FSC AUC on all five datasets as in
Tables~\ref{tab:cryobench_per_conformation_fsc}
and~\ref{tab:cryobench_per_image_fsc}.

On IgG-1D, the default $\tau=0.02$ gives the highest neighborhood
similarity, and thresholds of 0.01--0.05 retain conformational ordering in
the UMAP projections. Increasing $\tau$ to 0.10 reduces the estimated
support and the number of Gaussians, yielding a fragmented projection and
lower neighborhood similarity. Conversely, lowering $\tau$ to 0.01
increases the number of Gaussians without improving neighborhood similarity
over the default.

On all five datasets and under both protocols, FSC AUC is highest at
$\tau=0.02$ and decreases with every further increase in $\tau$, by
0.033--0.067 in mean FSC AUC between $\tau=0.02$ and $\tau=0.10$, with the
largest decrease on Tomotwin-100, where the number of Gaussians falls from
57,818 to 19,611. Lowering $\tau$ to 0.01 enlarges the estimated support and
the number of Gaussians without raising FSC AUC above the default on any
dataset. Reconstruction accuracy is thus highest at the default threshold,
decreases as the estimated support is restricted to denser regions of the
backprojection, and does not increase when weaker density is included.

\begin{table*}[tp]
    \centering
    \caption{\textbf{Sensitivity to the relative density threshold $\tau$ on CryoBench.}
$\tau$ selects the estimated spatial support of the density in the
CTF-weighted backprojection, and the number of Gaussians follows from the
space--bandwidth product of this support
(Section~\ref{sec:gaussian_capacity}). For each $\tau$, the number of
Gaussians and the per-image and per-conformation FSC AUC with background
masking are reported as in Tables~\ref{tab:cryobench_per_conformation_fsc}
and~\ref{tab:cryobench_per_image_fsc}, which correspond to the
default $\tau=0.02$. Bold marks the best per row.}
    \label{tab:tau_sensitivity}
    \setlength{\tabcolsep}{3pt}
    \setlength{\aboverulesep}{0pt}
    \setlength{\belowrulesep}{0pt}
    \renewcommand{\arraystretch}{1.22}
    \small
    \resizebox{\linewidth}{!}{%
    \begin{tabular}{@{}ll*{6}{c}@{}}
        \toprule
        & & \multicolumn{6}{c}{$\tau$} \\
        \cmidrule(lr){3-8}
        Dataset & & 0.01 & 0.02 & 0.03 & 0.04 & 0.05 & 0.10 \\
        \midrule
        IgG-1D & Gaussians & 16,111 & 10,316 & 9,231 & 8,359 & 7,598 & 4,542 \\
        & Per-image, mean (std) & 0.385 (0.007) & \textbf{0.397 (0.004)} & 0.390 (0.006) & 0.384 (0.009) & 0.376 (0.013) & 0.361 (0.021) \\
        & Per-image, median & 0.386 & \textbf{0.398} & 0.391 & 0.385 & 0.379 & 0.366 \\
        & Per-conformation, mean (std) & 0.392 (0.002) & \textbf{0.402 (0.001)} & 0.397 (0.001) & 0.391 (0.002) & 0.384 (0.003) & 0.369 (0.004) \\
        & Per-conformation, median & 0.392 & \textbf{0.402} & 0.397 & 0.391 & 0.385 & 0.371 \\
        \addlinespace[2pt]
        IgG-RL & Gaussians & 20,612 & 11,836 & 8,996 & 7,361 & 6,285 & 3,935 \\
        & Per-image, mean (std) & 0.359 (0.026) & \textbf{0.366 (0.020)} & 0.355 (0.025) & 0.352 (0.028) & 0.343 (0.033) & 0.325 (0.044) \\
        & Per-image, median & 0.364 & \textbf{0.370} & 0.360 & 0.357 & 0.351 & 0.336 \\
        & Per-conformation, mean (std) & 0.381 (0.012) & \textbf{0.387 (0.010)} & 0.379 (0.011) & 0.375 (0.014) & 0.368 (0.016) & 0.351 (0.021) \\
        & Per-conformation, median & 0.382 & \textbf{0.387} & 0.381 & 0.376 & 0.368 & 0.354 \\
        \addlinespace[2pt]
        Ribosembly & Gaussians & 37,387 & 33,033 & 31,647 & 30,420 & 29,285 & 24,076 \\
        & Per-image, mean (std) & 0.429 (0.018) & \textbf{0.432 (0.014)} & 0.425 (0.019) & 0.419 (0.023) & 0.409 (0.028) & 0.390 (0.039) \\
        & Per-image, median & 0.429 & \textbf{0.432} & 0.426 & 0.421 & 0.414 & 0.394 \\
        & Per-conformation, mean (std) & 0.430 (0.018) & \textbf{0.432 (0.016)} & 0.428 (0.018) & 0.423 (0.021) & 0.411 (0.026) & 0.394 (0.038) \\
        & Per-conformation, median & 0.431 & \textbf{0.432} & 0.428 & 0.424 & 0.415 & 0.396 \\
        \addlinespace[2pt]
        Tomotwin-100 & Gaussians & 87,988 & 57,818 & 44,938 & 37,322 & 32,124 & 19,611 \\
        & Per-image, mean (std) & 0.331 (0.040) & \textbf{0.335 (0.035)} & 0.321 (0.043) & 0.311 (0.051) & 0.300 (0.061) & 0.268 (0.082) \\
        & Per-image, median & 0.337 & \textbf{0.341} & 0.329 & 0.320 & 0.312 & 0.287 \\
        & Per-conformation, mean (std) & 0.333 (0.038) & \textbf{0.336 (0.034)} & 0.325 (0.042) & 0.315 (0.048) & 0.305 (0.057) & 0.273 (0.075) \\
        & Per-conformation, median & 0.340 & \textbf{0.343} & 0.333 & 0.325 & 0.318 & 0.293 \\
        \addlinespace[2pt]
        Spike-MD & Gaussians & 142,530 & 115,877 & 109,102 & 103,766 & 99,135 & 81,890 \\
        & Per-image, mean (std) & 0.355 (0.015) & \textbf{0.364 (0.012)} & 0.356 (0.015) & 0.348 (0.019) & 0.339 (0.023) & 0.317 (0.031) \\
        & Per-image, median & 0.358 & \textbf{0.366} & 0.358 & 0.352 & 0.343 & 0.325 \\
        & Per-conformation, mean (std) & 0.357 (0.013) & \textbf{0.365 (0.012)} & 0.358 (0.014) & 0.350 (0.016) & 0.343 (0.020) & 0.323 (0.027) \\
        & Per-conformation, median & 0.359 & \textbf{0.367} & 0.360 & 0.353 & 0.347 & 0.329 \\
        \bottomrule
    \end{tabular}}
\end{table*}

\subsection{Gaussian FWHM constraint}
\label{app:fwhm_ablation}

\begin{figure*}[tp]
    \centering
    \includegraphics[width=1\linewidth]{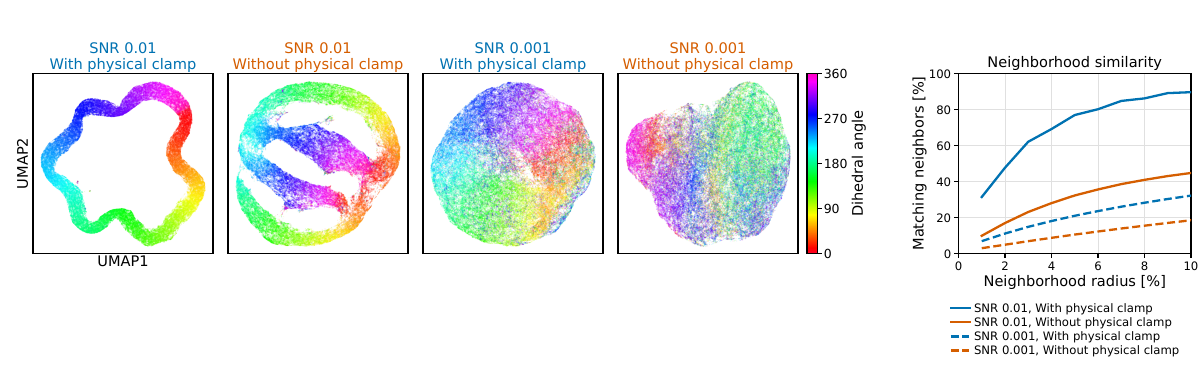}
    \caption{\textbf{Ablation of the Gaussian FWHM constraint on IgG-1D.}
    Heterogeneity estimation is evaluated at SNR 0.01 and 0.001 with
    and without the constraint in Equation~\eqref{eq:fwhm_constraint},
    denoted ``physical clamp'' in the figure.
    \textbf{Left:} UMAP projections of inferred latents, colored by the
    ground-truth dihedral angle. \textbf{Right:} Neighborhood similarity,
    measured as the percentage of matching neighbors (pMN) between the
    inferred latents and the ground-truth heterogeneity embedding,
    following Appendix~\ref{app:latent_evaluation}.
    The constraint yields higher pMN across the displayed neighborhood
    radii at both noise levels. Colors denote the constraint setting,
    and line styles denote SNR.}
    \label{fig:fwhm_ablation}
\end{figure*}

We compare heterogeneity estimation on IgG-1D with and without the
Gaussian FWHM constraint at SNR 0.01 and 0.001
(Figure~\ref{fig:fwhm_ablation}). The constraint in
Equation~\eqref{eq:fwhm_constraint} sets a lower bound of $2a$ on Gaussian
FWHM, preventing Gaussian widths from decreasing below this sampling scale.
At SNR 0.01, the inferred latents retain an ordered loop in the UMAP
projection with the constraint, whereas removing it produces folded,
overlapping branches. At SNR 0.001, both projections become diffuse.
Neighborhood similarity, evaluated in the inferred latent space against
the ground-truth heterogeneity embedding, is higher with the constraint
across the displayed neighborhood radii at both noise levels. These
comparisons support constraining Gaussian FWHM when learning conformational
heterogeneity from noisy particle images.

Table~\ref{tab:fwhm_ablation} evaluates per-image and per-conformation FSC
AUC on all five CryoBench datasets with and without the constraint, as in
Tables~\ref{tab:cryobench_per_conformation_fsc}
and~\ref{tab:cryobench_per_image_fsc}. On all five datasets and under both
protocols, removing the constraint decreases FSC AUC, by 0.016--0.030 in
mean and 0.016--0.028 in median FSC AUC, and increases its standard
deviation across evaluated particle images and across reference states.
The constraint raises reconstruction accuracy on every dataset in addition
to neighborhood similarity on IgG-1D.

\begin{table*}[tp]
    \centering
    \caption{\textbf{Ablation of the Gaussian FWHM constraint on CryoBench.}
Mean (standard deviation) and median FSC AUC with background masking across
evaluated particle images (per-image) and reference states
(per-conformation) with and without the constraint in
Equation~\eqref{eq:fwhm_constraint}. The results with the constraint
correspond to Tables~\ref{tab:cryobench_per_conformation_fsc}
and~\ref{tab:cryobench_per_image_fsc}. Higher is better. Bold marks the
best per dataset and column.}
    \label{tab:fwhm_ablation}
    \setlength{\tabcolsep}{6pt}
    \setlength{\aboverulesep}{0pt}
    \setlength{\belowrulesep}{0pt}
    \renewcommand{\arraystretch}{1.3}
    \small
    {%
        \begin{tabular}{@{}llcccc@{}}
            \toprule
            Dataset & FWHM constraint
            & \multicolumn{2}{c}{Per-image}
            & \multicolumn{2}{c}{Per-conformation} \\
            \cmidrule(lr){3-4}
            \cmidrule(lr){5-6}
            & & Mean (std) & Median & Mean (std) & Median \\
            \midrule
            IgG-1D & With & \textbf{0.397 (0.004)} & \textbf{0.398} & \textbf{0.402 (0.001)} & \textbf{0.402} \\
            & Without & 0.372 (0.013) & 0.374 & 0.386 (0.005) & 0.386 \\
            \midrule
            IgG-RL & With & \textbf{0.366 (0.020)} & \textbf{0.370} & \textbf{0.387 (0.010)} & \textbf{0.387} \\
            & Without & 0.342 (0.033) & 0.347 & 0.370 (0.018) & 0.371 \\
            \midrule
            Ribosembly & With & \textbf{0.432 (0.014)} & \textbf{0.432} & \textbf{0.432 (0.016)} & \textbf{0.432} \\
            & Without & 0.402 (0.025) & 0.404 & 0.406 (0.027) & 0.407 \\
            \midrule
            Tomotwin-100 & With & \textbf{0.335 (0.035)} & \textbf{0.341} & \textbf{0.336 (0.034)} & \textbf{0.343} \\
            & Without & 0.306 (0.047) & 0.314 & 0.313 (0.045) & 0.321 \\
            \midrule
            Spike-MD & With & \textbf{0.364 (0.012)} & \textbf{0.366} & \textbf{0.365 (0.012)} & \textbf{0.367} \\
            & Without & 0.343 (0.019) & 0.345 & 0.344 (0.019) & 0.347 \\
            \bottomrule
        \end{tabular}%
    }
\end{table*}

\subsection{Gaussian initialization}
\label{app:initialization_sensitivity}

\begin{figure*}[tp]
    \centering
    \includegraphics[width=1\linewidth]{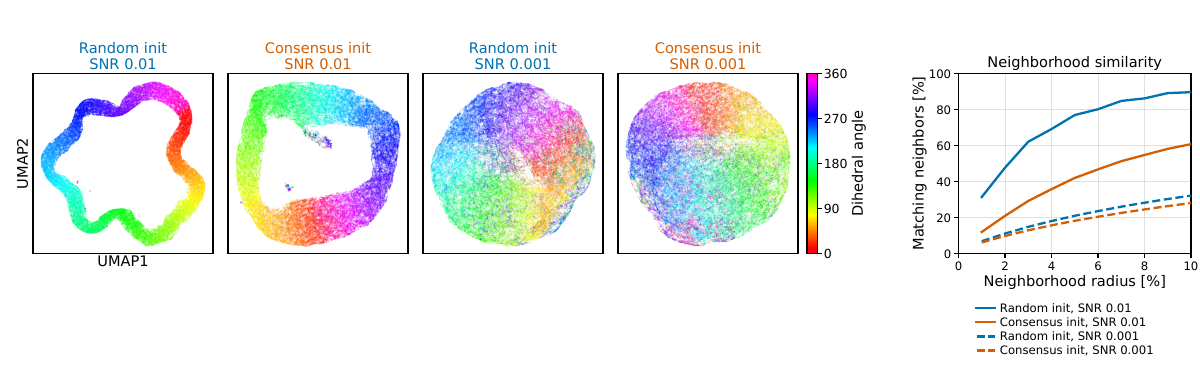}
    \caption{\textbf{Ablation of Gaussian initialization on IgG-1D.}
    Heterogeneity estimation is evaluated at SNR 0.01 and 0.001 using random initialization
    or Gaussian positions sampled from a consensus density optimized without
    latents on all particle images. Consensus initialization uses
    density-weighted sampling without replacement over thresholded voxels,
    as detailed in Appendix~\ref{app:initialization_sensitivity}.
    \textbf{Left:} UMAP projections of inferred latents, colored by the
    ground-truth dihedral angle. \textbf{Right:} Neighborhood similarity,
    measured as the percentage of matching neighbors (pMN) between the
    inferred latents and the ground-truth heterogeneity embedding, plotted
    against neighborhood radius as a percentage of the 20,000-image
    evaluation subset (Appendix~\ref{app:latent_evaluation}).
    Higher pMN indicates better preservation of reference conformational
    neighborhoods. Colors denote Gaussian initialization, and line styles
    denote SNR. Random initialization yields higher pMN across the
    evaluated neighborhood radii at both noise levels.}
    \label{fig:initialization_sensitivity}
\end{figure*}

IgG-1D provides a ground-truth one-dimensional conformational motion for evaluating
whether inferred latents preserve reference conformational neighborhoods.
We compare random and consensus initialization for heterogeneity estimation at
SNR 0.01 and 0.001, giving four runs in total (Figure~\ref{fig:initialization_sensitivity}).

Random initialization follows Appendix~\ref{app:gaussian_parameterization}.
For consensus initialization, we first
obtain a consensus density by optimizing the Gaussian representation without
latents on all particle images for 30,000 iterations, with the learning
rates, data-derived constraints, and relocation schedule of heterogeneity
estimation in Appendix~\ref{app:training_ablation_settings} otherwise unchanged, which
yields a single density fitted to all particle images irrespective of their
structural states. We initialize Gaussian positions by density-weighted sampling
without replacement over thresholded voxels. Only the
positions are initialized from this density; the other Gaussian attributes
use the same initialization as in the random-initialization runs. The positions
are optimized during training.

Each run jointly learns latents for the particle images and the
Gaussian density representation. We compute neighborhood similarity in the
inferred latent space following Appendix~\ref{app:latent_evaluation}; UMAP
projections visualize the latents colored by the ground-truth dihedral angle.

Averaged over 20 neighborhood radii (1--20\% of the evaluation subset),
pMN is 82.73\% for random initialization and 55.84\% for consensus
initialization at SNR 0.01.
At SNR 0.001, the corresponding values are 30.89\% and 27.10\%.
Random initialization yields higher pMN across all evaluated neighborhood
radii at both noise levels.

\subsection{Per-state reconstruction strategy}
\label{app:reconstruction_strategy}

Table~\ref{tab:reconstruction_strategy} compares three ways of obtaining
per-state densities on the five CryoBench datasets at the same total number
of iterations and otherwise identical settings. Heterogeneity estimation
alone optimizes $\Theta_1$ and $Z$ jointly for the combined 60,000
iterations of heterogeneity estimation and per-state reconstruction and
evaluates the densities of $\Theta_1$ at the inferred latents. Per-state
reconstruction continued from $\Theta_1$ continues the optimization of
$\Theta_1$ for the per-state reconstruction iterations conditioned on the
inferred latents $Z^\star$. Per-state reconstruction from randomly
initialized $\Theta_2$ conditioned on $Z^\star$ follows
Section~\ref{sec:heterogeneous_learning} and corresponds to the results in
Tables~\ref{tab:cryobench_per_conformation_fsc}
and~\ref{tab:cryobench_per_image_fsc}. The two per-state
reconstruction strategies use the same heterogeneity estimation and its
inferred latents $Z^\star$, so that their difference reflects per-state
reconstruction alone. Per-state reconstruction from randomly initialized
$\Theta_2$ attains the highest mean FSC AUC on all five datasets under both
protocols, by 0.008--0.016 over heterogeneity estimation alone and
0.005--0.010 over per-state reconstruction continued from $\Theta_1$. On
every dataset and under both protocols, the increase in mean FSC AUC from
continued optimization to reconstruction from randomly initialized
$\Theta_2$ exceeds the increase from heterogeneity estimation alone to
continued optimization, so that the larger part of the improvement at the
same total number of iterations arises from reconstruction from random
initialization rather than from further optimization of $\Theta_1$.

\begin{table*}[tp]
    \centering
    \caption{\textbf{Per-state reconstruction strategies on CryoBench.}
Mean (standard deviation) and median FSC AUC with background masking across
evaluated particle images (per-image) and reference states
(per-conformation) for three ways of obtaining per-state densities at the
same total number of iterations. Heterogeneity estimation alone optimizes
$\Theta_1$ and the latents jointly and evaluates $\Theta_1$ at the inferred
latents. Per-state reconstruction continued from $\Theta_1$ continues
optimizing $\Theta_1$ conditioned on the inferred latents. Per-state
reconstruction from randomly initialized $\Theta_2$ follows
Section~\ref{sec:heterogeneous_learning} and corresponds to
Tables~\ref{tab:cryobench_per_conformation_fsc}
and~\ref{tab:cryobench_per_image_fsc}. Higher is better. Bold marks
the best per dataset and column, including ties at the displayed
precision.}
    \label{tab:reconstruction_strategy}
    \setlength{\tabcolsep}{6pt}
    \setlength{\aboverulesep}{0pt}
    \setlength{\belowrulesep}{0pt}
    \renewcommand{\arraystretch}{1.3}
    \small
    \resizebox{\linewidth}{!}{%
        \begin{tabular}{@{}llcccc@{}}
            \toprule
            Dataset & Strategy
            & \multicolumn{2}{c}{Per-image}
            & \multicolumn{2}{c}{Per-conformation} \\
            \cmidrule(lr){3-4}
            \cmidrule(lr){5-6}
            & & Mean (std) & Median & Mean (std) & Median \\
            \midrule
            IgG-1D & Heterogeneity estimation alone & 0.381 (0.008) & 0.381 & 0.390 (0.004) & 0.390 \\
            & Per-state reconstruction continued from $\Theta_1$ & 0.387 (0.007) & 0.389 & 0.395 (0.003) & 0.394 \\
            & Per-state reconstruction from randomly initialized $\Theta_2$ & \textbf{0.397 (0.004)} & \textbf{0.398} & \textbf{0.402 (0.001)} & \textbf{0.402} \\
            \midrule
            IgG-RL & Heterogeneity estimation alone & 0.358 (0.027) & 0.362 & 0.379 (0.020) & 0.378 \\
            & Per-state reconstruction continued from $\Theta_1$ & 0.361 (0.024) & 0.366 & 0.382 (0.018) & 0.381 \\
            & Per-state reconstruction from randomly initialized $\Theta_2$ & \textbf{0.366 (0.020)} & \textbf{0.370} & \textbf{0.387 (0.010)} & \textbf{0.387} \\
            \midrule
            Ribosembly & Heterogeneity estimation alone & 0.421 (0.023) & 0.421 & 0.422 (0.026) & 0.422 \\
            & Per-state reconstruction continued from $\Theta_1$ & 0.426 (0.021) & 0.427 & 0.426 (0.022) & 0.423 \\
            & Per-state reconstruction from randomly initialized $\Theta_2$ & \textbf{0.432 (0.014)} & \textbf{0.432} & \textbf{0.432 (0.016)} & \textbf{0.432} \\
            \midrule
            Tomotwin-100 & Heterogeneity estimation alone & 0.320 (0.041) & 0.324 & 0.322 (0.048) & 0.329 \\
            & Per-state reconstruction continued from $\Theta_1$ & 0.326 (0.037) & 0.332 & 0.328 (0.038) & 0.338 \\
            & Per-state reconstruction from randomly initialized $\Theta_2$ & \textbf{0.335 (0.035)} & \textbf{0.341} & \textbf{0.336 (0.034)} & \textbf{0.343} \\
            \midrule
            Spike-MD & Heterogeneity estimation alone & 0.351 (0.026) & 0.354 & 0.353 (0.029) & 0.358 \\
            & Per-state reconstruction continued from $\Theta_1$ & 0.356 (0.022) & 0.359 & 0.358 (0.027) & 0.361 \\
            & Per-state reconstruction from randomly initialized $\Theta_2$ & \textbf{0.364 (0.012)} & \textbf{0.366} & \textbf{0.365 (0.012)} & \textbf{0.367} \\
            \bottomrule
        \end{tabular}%
    }
\end{table*}

\subsection{Combinations of Gaussian attribute decoders}
\label{app:decoder_ablation}

Tables~\ref{tab:decoder_ablation_per_conformation}
and~\ref{tab:decoder_ablation_per_image} evaluate all 15 combinations of
the position, amplitude, scale, and rotation decoders, from one decoder
to all four, on the five CryoBench datasets with the settings of
Section~\ref{sec:experiments} otherwise unchanged. The combination of the
position, amplitude, and scale decoders is the configuration of
CryoGaussian (Section~\ref{sec:heterogeneous_learning} and
Appendix~\ref{app:decoder_implementation}), in which the Gaussian rotation
$Q_g$ has no decoder. The rotation decoder has the same architecture as the other
decoders and produces an axis--angle vector
$\boldsymbol\omega_g(\mathbf z)\in\mathbb R^3$ from $[\mathbf f_g;\mathbf z]$,
so that $\exp([\boldsymbol\omega_g(\mathbf z)]_\times)\,Q_g$ replaces $Q_g$
in Equation~\eqref{eq:state_effective_scale}, where $[\cdot]_\times$ denotes
the skew-symmetric matrix of a vector. Its output layer is initialized to
zero, so that the decoded rotation equals $Q_g$ at initialization, and $Q_g$
is still learned per Gaussian. In every combination, the attributes without a
decoder keep their base values, so that a Gaussian without the position
decoder stays at $\boldsymbol\mu_g^{\mathrm{base}}$, one without the
amplitude decoder keeps $p_g^{\mathrm{base}}$, one without the scale decoder
keeps its base scales under the FWHM constraint, and one without the
rotation decoder keeps $Q_g$ (Appendix~\ref{app:gaussian_parameterization}).

The combination of the position, amplitude, and scale decoders attains
the highest mean and median FSC AUC and the lowest standard deviation on all
five datasets under both protocols, by 0.007--0.010 in mean FSC AUC over the
best combination of two decoders and 0.044--0.081 over the best single
decoder. Removing one decoder from this combination decreases mean FSC AUC on every
dataset under both protocols, by 0.013--0.027 for the amplitude decoder,
which is the largest decrease on every dataset, by 0.007--0.020 for the
position decoder, and by 0.008--0.015 for the scale decoder. Removing the
position decoder decreases mean FSC AUC by 0.012--0.020 on the
conformational datasets IgG-1D, IgG-RL, and Spike-MD and by 0.007--0.010 on
the compositional datasets Ribosembly and Tomotwin-100, and removing the
scale decoder by 0.008--0.009 on the conformational and 0.011--0.015 on the
compositional datasets. The same ordering holds with one and two decoders,
where the amplitude decoder alone attains the highest mean FSC AUC on every
dataset, the combination of the position and amplitude decoders is the
highest combination of two decoders on the conformational datasets, and the
combination of the amplitude and scale decoders is the highest on the
compositional datasets, in
agreement with the roles of the three attributes in
Section~\ref{sec:heterogeneous_learning}, where position modulation moves Gaussians
between states and amplitude and scale modulation change their presence and
shape. Adding the rotation decoder decreases mean FSC AUC in every
combination on every dataset under both protocols, by 0.001--0.006 when
added to one or two decoders and by 0.006--0.012 when added to the
combination of the position, amplitude, and scale decoders, and the rotation decoder alone
attains the lowest mean FSC AUC on every dataset. The configuration of
CryoGaussian is accordingly the combination of the position, amplitude,
and scale decoders, with no rotation decoder.

\begin{table*}[tp]
    \centering
    \caption{\textbf{Combinations of Gaussian attribute decoders on CryoBench, per-conformation FSC AUC.}
Mean (standard deviation) and median FSC AUC with background masking across
evaluated reference states for all 15 combinations of the position, amplitude,
scale, and rotation decoders, grouped by the number of decoders
(Appendix~\ref{app:decoder_ablation}). Each row lists the decoders used,
and attributes without a decoder keep their base values
(Section~\ref{sec:heterogeneous_learning} and
Appendix~\ref{app:gaussian_parameterization}). The combination of the
position, amplitude, and scale decoders is the configuration of
CryoGaussian and corresponds to Table~\ref{tab:cryobench_per_conformation_fsc}. Higher is better.
Bold marks the best per column.}
    \label{tab:decoder_ablation_per_conformation}
    \setlength{\tabcolsep}{3pt}
    \setlength{\aboverulesep}{0pt}
    \setlength{\belowrulesep}{0pt}
    \renewcommand{\arraystretch}{1.3}
    \resizebox{\linewidth}{!}{%
        \begin{tabular}{@{}l*{5}{cc}@{}}
            \toprule
            Decoders
            & \multicolumn{2}{c}{IgG-1D}
            & \multicolumn{2}{c}{IgG-RL}
            & \multicolumn{2}{c}{Ribosembly}
            & \multicolumn{2}{c}{Tomotwin-100}
            & \multicolumn{2}{c}{Spike-MD} \\
            \cmidrule(lr){2-3}\cmidrule(lr){4-5}\cmidrule(lr){6-7}\cmidrule(lr){8-9}\cmidrule(lr){10-11}
            & Mean (std) & Median & Mean (std) & Median & Mean (std) & Median & Mean (std) & Median & Mean (std) & Median \\
            \midrule
            Position & 0.343 (0.010) & 0.344 & 0.309 (0.024) & 0.306 & 0.305 (0.047) & 0.308 & 0.208 (0.071) & 0.204 & 0.287 (0.028) & 0.282 \\
            Amplitude & 0.358 (0.009) & 0.356 & 0.327 (0.020) & 0.327 & 0.360 (0.042) & 0.362 & 0.259 (0.054) & 0.266 & 0.308 (0.023) & 0.307 \\
            Scale & 0.323 (0.017) & 0.320 & 0.286 (0.033) & 0.286 & 0.335 (0.052) & 0.334 & 0.237 (0.068) & 0.232 & 0.269 (0.029) & 0.271 \\
            Rotation & 0.314 (0.018) & 0.312 & 0.278 (0.036) & 0.280 & 0.299 (0.048) & 0.300 & 0.201 (0.071) & 0.195 & 0.265 (0.033) & 0.261 \\
            \midrule
            Position and amplitude & 0.394 (0.003) & 0.394 & 0.379 (0.015) & 0.378 & 0.421 (0.022) & 0.420 & 0.321 (0.045) & 0.320 & 0.356 (0.018) & 0.356 \\
            Amplitude and scale & 0.382 (0.005) & 0.381 & 0.375 (0.020) & 0.377 & 0.425 (0.018) & 0.425 & 0.326 (0.037) & 0.325 & 0.347 (0.022) & 0.347 \\
            Position and scale & 0.377 (0.007) & 0.378 & 0.374 (0.021) & 0.372 & 0.418 (0.028) & 0.417 & 0.319 (0.043) & 0.318 & 0.342 (0.019) & 0.342 \\
            Amplitude and rotation & 0.356 (0.013) & 0.353 & 0.323 (0.024) & 0.323 & 0.358 (0.046) & 0.360 & 0.257 (0.057) & 0.263 & 0.306 (0.027) & 0.304 \\
            Position and rotation & 0.340 (0.014) & 0.340 & 0.307 (0.028) & 0.304 & 0.301 (0.047) & 0.306 & 0.206 (0.074) & 0.201 & 0.285 (0.031) & 0.282 \\
            Scale and rotation & 0.318 (0.021) & 0.319 & 0.283 (0.035) & 0.284 & 0.333 (0.053) & 0.331 & 0.235 (0.071) & 0.229 & 0.267 (0.033) & 0.267 \\
            \midrule
            Position, amplitude, and scale (CryoGaussian) & \textbf{0.402 (0.001)} & \textbf{0.402} & \textbf{0.387 (0.010)} & \textbf{0.387} & \textbf{0.432 (0.016)} & \textbf{0.432} & \textbf{0.336 (0.034)} & \textbf{0.343} & \textbf{0.365 (0.012)} & \textbf{0.367} \\
            Position, amplitude, and rotation & 0.391 (0.004) & 0.391 & 0.377 (0.017) & 0.373 & 0.419 (0.025) & 0.419 & 0.317 (0.046) & 0.319 & 0.352 (0.019) & 0.352 \\
            Amplitude, scale, and rotation & 0.377 (0.006) & 0.377 & 0.370 (0.020) & 0.372 & 0.423 (0.022) & 0.423 & 0.324 (0.039) & 0.321 & 0.344 (0.025) & 0.344 \\
            Position, scale, and rotation & 0.374 (0.008) & 0.375 & 0.368 (0.022) & 0.366 & 0.417 (0.029) & 0.416 & 0.314 (0.043) & 0.314 & 0.341 (0.022) & 0.340 \\
            \midrule
            Position, amplitude, scale, and rotation & 0.394 (0.003) & 0.393 & 0.379 (0.015) & 0.381 & 0.426 (0.019) & 0.428 & 0.327 (0.036) & 0.325 & 0.353 (0.018) & 0.353 \\
            \bottomrule
        \end{tabular}%
    }
\end{table*}

\begin{table*}[tp]
    \centering
    \caption{\textbf{Combinations of Gaussian attribute decoders on CryoBench, per-image FSC AUC.}
Mean (standard deviation) and median FSC AUC with background masking across
evaluated particle images for all 15 combinations of the position, amplitude,
scale, and rotation decoders, grouped by the number of decoders
(Appendix~\ref{app:decoder_ablation}). Each row lists the decoders used,
and attributes without a decoder keep their base values
(Section~\ref{sec:heterogeneous_learning} and
Appendix~\ref{app:gaussian_parameterization}). The combination of the
position, amplitude, and scale decoders is the configuration of
CryoGaussian and corresponds to Table~\ref{tab:cryobench_per_image_fsc}. Higher is better.
Bold marks the best per column.}
    \label{tab:decoder_ablation_per_image}
    \setlength{\tabcolsep}{3pt}
    \setlength{\aboverulesep}{0pt}
    \setlength{\belowrulesep}{0pt}
    \renewcommand{\arraystretch}{1.3}
    \resizebox{\linewidth}{!}{%
        \begin{tabular}{@{}l*{5}{cc}@{}}
            \toprule
            Decoders
            & \multicolumn{2}{c}{IgG-1D}
            & \multicolumn{2}{c}{IgG-RL}
            & \multicolumn{2}{c}{Ribosembly}
            & \multicolumn{2}{c}{Tomotwin-100}
            & \multicolumn{2}{c}{Spike-MD} \\
            \cmidrule(lr){2-3}\cmidrule(lr){4-5}\cmidrule(lr){6-7}\cmidrule(lr){8-9}\cmidrule(lr){10-11}
            & Mean (std) & Median & Mean (std) & Median & Mean (std) & Median & Mean (std) & Median & Mean (std) & Median \\
            \midrule
            Position & 0.334 (0.017) & 0.333 & 0.287 (0.041) & 0.290 & 0.305 (0.052) & 0.306 & 0.207 (0.076) & 0.202 & 0.287 (0.028) & 0.283 \\
            Amplitude & 0.348 (0.015) & 0.351 & 0.303 (0.035) & 0.303 & 0.356 (0.041) & 0.353 & 0.254 (0.056) & 0.252 & 0.308 (0.022) & 0.307 \\
            Scale & 0.313 (0.024) & 0.320 & 0.263 (0.050) & 0.267 & 0.332 (0.055) & 0.327 & 0.235 (0.063) & 0.234 & 0.269 (0.030) & 0.270 \\
            Rotation & 0.305 (0.026) & 0.311 & 0.255 (0.052) & 0.260 & 0.299 (0.051) & 0.297 & 0.198 (0.076) & 0.193 & 0.264 (0.032) & 0.261 \\
            \midrule
            Position and amplitude & 0.389 (0.008) & 0.387 & 0.357 (0.026) & 0.355 & 0.420 (0.022) & 0.421 & 0.322 (0.042) & 0.321 & 0.356 (0.017) & 0.355 \\
            Amplitude and scale & 0.377 (0.012) & 0.378 & 0.352 (0.027) & 0.352 & 0.424 (0.017) & 0.424 & 0.325 (0.039) & 0.326 & 0.345 (0.022) & 0.346 \\
            Position and scale & 0.370 (0.013) & 0.371 & 0.350 (0.034) & 0.350 & 0.416 (0.023) & 0.415 & 0.319 (0.043) & 0.320 & 0.342 (0.019) & 0.342 \\
            Amplitude and rotation & 0.344 (0.017) & 0.348 & 0.301 (0.037) & 0.299 & 0.352 (0.041) & 0.348 & 0.250 (0.058) & 0.249 & 0.306 (0.027) & 0.305 \\
            Position and rotation & 0.333 (0.021) & 0.331 & 0.283 (0.041) & 0.289 & 0.302 (0.056) & 0.303 & 0.204 (0.079) & 0.198 & 0.285 (0.031) & 0.282 \\
            Scale and rotation & 0.310 (0.029) & 0.315 & 0.259 (0.051) & 0.266 & 0.327 (0.059) & 0.323 & 0.232 (0.066) & 0.232 & 0.267 (0.033) & 0.267 \\
            \midrule
            Position, amplitude, and scale (CryoGaussian) & \textbf{0.397 (0.004)} & \textbf{0.398} & \textbf{0.366 (0.020)} & \textbf{0.370} & \textbf{0.432 (0.014)} & \textbf{0.432} & \textbf{0.335 (0.035)} & \textbf{0.341} & \textbf{0.364 (0.012)} & \textbf{0.366} \\
            Position, amplitude, and rotation & 0.387 (0.010) & 0.384 & 0.356 (0.026) & 0.351 & 0.416 (0.024) & 0.419 & 0.318 (0.045) & 0.316 & 0.352 (0.019) & 0.352 \\
            Amplitude, scale, and rotation & 0.373 (0.012) & 0.375 & 0.349 (0.029) & 0.348 & 0.420 (0.019) & 0.420 & 0.321 (0.041) & 0.324 & 0.344 (0.025) & 0.343 \\
            Position, scale, and rotation & 0.368 (0.015) & 0.367 & 0.346 (0.035) & 0.346 & 0.413 (0.023) & 0.414 & 0.315 (0.047) & 0.315 & 0.341 (0.023) & 0.340 \\
            \midrule
            Position, amplitude, scale, and rotation & 0.390 (0.009) & 0.391 & 0.358 (0.023) & 0.356 & 0.425 (0.019) & 0.428 & 0.324 (0.042) & 0.322 & 0.354 (0.018) & 0.353 \\
            \bottomrule
        \end{tabular}%
    }
\end{table*}

\subsection{Latent dimensions}
\label{app:latent_dimensions}

Table~\ref{tab:latent_dimensions} varies the particle-image latent dimension
$d_z$ at $d_f=24$ and the per-Gaussian latent dimension $d_f$ at $d_z=4$ on
the five CryoBench datasets with the settings of Section~\ref{sec:experiments}
otherwise unchanged. The configuration of CryoGaussian, $d_z=4$ and
$d_f=24$, attains the highest mean and median FSC AUC and the lowest standard
deviation on all five datasets under both protocols. Decreasing $d_z$ to 2
decreases mean FSC AUC by 0.016--0.025 on the conformational datasets IgG-1D,
IgG-RL, and Spike-MD and by 0.008--0.012 on the compositional datasets
Ribosembly and Tomotwin-100, whereas increasing $d_z$ to 8 or 10 decreases it
by 0.006--0.017 on the conformational and by 0.013--0.027 on the compositional
datasets. Decreasing $d_f$ to 16 decreases mean FSC AUC by 0.007--0.010 on
the conformational and by 0.021--0.023 on the compositional datasets, whereas
increasing $d_f$ to 32 decreases it by 0.013--0.018 on the conformational and
by 0.007--0.009 on the compositional datasets. The conformational datasets
are more sensitive to a smaller particle-image latent and a larger
per-Gaussian latent, and the compositional datasets to a larger
particle-image latent and a smaller per-Gaussian latent.

\begin{table*}[tp]
    \centering
    \caption{\textbf{Latent dimensions on CryoBench.}
Mean (standard deviation) and median FSC AUC with background masking across
evaluated particle images (per-image) and reference states
(per-conformation) for particle-image latent dimension $d_z$ and
per-Gaussian latent dimension $d_f$, with the settings of
Section~\ref{sec:experiments} otherwise unchanged. $d_z=4$ and $d_f=24$ is
the configuration of CryoGaussian and corresponds to
Tables~\ref{tab:cryobench_per_conformation_fsc}
and~\ref{tab:cryobench_per_image_fsc}. Higher is better. Bold marks the
best per column.}
    \label{tab:latent_dimensions}
    \setlength{\tabcolsep}{3pt}
    \setlength{\aboverulesep}{0pt}
    \setlength{\belowrulesep}{0pt}
    \renewcommand{\arraystretch}{1.3}
    \small
    \resizebox{\linewidth}{!}{%
        \begin{tabular}{@{}cc*{5}{cc}@{}}
            \toprule
            $d_z$ & $d_f$
            & \multicolumn{2}{c}{IgG-1D}
            & \multicolumn{2}{c}{IgG-RL}
            & \multicolumn{2}{c}{Ribosembly}
            & \multicolumn{2}{c}{Tomotwin-100}
            & \multicolumn{2}{c}{Spike-MD} \\
            \cmidrule(lr){3-4}\cmidrule(lr){5-6}\cmidrule(lr){7-8}\cmidrule(lr){9-10}\cmidrule(lr){11-12}
            & & Mean (std) & Median & Mean (std) & Median & Mean (std) & Median & Mean (std) & Median & Mean (std) & Median \\
            \midrule
            \multicolumn{12}{@{}l}{Per-image} \\
            2 & 24 & 0.380 (0.008) & 0.381 & 0.341 (0.028) & 0.343 & 0.422 (0.018) & 0.423 & 0.323 (0.042) & 0.328 & 0.342 (0.017) & 0.344 \\
            \textbf{4} & \textbf{24} & \textbf{0.397 (0.004)} & \textbf{0.398} & \textbf{0.366 (0.020)} & \textbf{0.370} & \textbf{0.432 (0.014)} & \textbf{0.432} & \textbf{0.335 (0.035)} & \textbf{0.341} & \textbf{0.364 (0.012)} & \textbf{0.366} \\
            8 & 24 & 0.390 (0.007) & 0.390 & 0.358 (0.025) & 0.360 & 0.418 (0.024) & 0.419 & 0.315 (0.043) & 0.318 & 0.358 (0.016) & 0.361 \\
            10 & 24 & 0.385 (0.010) & 0.385 & 0.349 (0.027) & 0.355 & 0.405 (0.030) & 0.408 & 0.310 (0.049) & 0.313 & 0.354 (0.022) & 0.356 \\
            4 & 16 & 0.388 (0.007) & 0.388 & 0.356 (0.026) & 0.357 & 0.411 (0.024) & 0.413 & 0.312 (0.047) & 0.318 & 0.355 (0.018) & 0.359 \\
            4 & 32 & 0.384 (0.009) & 0.384 & 0.348 (0.028) & 0.354 & 0.423 (0.020) & 0.425 & 0.327 (0.042) & 0.334 & 0.351 (0.019) & 0.352 \\
            \midrule
            \multicolumn{12}{@{}l}{Per-conformation} \\
            2 & 24 & 0.386 (0.004) & 0.386 & 0.363 (0.016) & 0.365 & 0.424 (0.022) & 0.424 & 0.324 (0.043) & 0.330 & 0.342 (0.018) & 0.343 \\
            \textbf{4} & \textbf{24} & \textbf{0.402 (0.001)} & \textbf{0.402} & \textbf{0.387 (0.010)} & \textbf{0.387} & \textbf{0.432 (0.016)} & \textbf{0.432} & \textbf{0.336 (0.034)} & \textbf{0.343} & \textbf{0.365 (0.012)} & \textbf{0.367} \\
            8 & 24 & 0.396 (0.003) & 0.396 & 0.378 (0.013) & 0.382 & 0.419 (0.025) & 0.420 & 0.316 (0.049) & 0.320 & 0.358 (0.016) & 0.361 \\
            10 & 24 & 0.392 (0.005) & 0.390 & 0.373 (0.015) & 0.374 & 0.406 (0.027) & 0.410 & 0.312 (0.050) & 0.314 & 0.355 (0.022) & 0.357 \\
            4 & 16 & 0.393 (0.003) & 0.392 & 0.380 (0.015) & 0.380 & 0.411 (0.025) & 0.414 & 0.313 (0.047) & 0.318 & 0.356 (0.018) & 0.359 \\
            4 & 32 & 0.389 (0.004) & 0.390 & 0.371 (0.017) & 0.374 & 0.425 (0.021) & 0.426 & 0.328 (0.040) & 0.336 & 0.351 (0.020) & 0.352 \\
            \bottomrule
        \end{tabular}%
    }
\end{table*}

\subsection{Particle selection}
\label{app:particle_selection}

\begin{figure*}[p]
    \centering
    \includegraphics[width=\linewidth, height=\dimexpr\textheight-18\baselineskip\relax, keepaspectratio]{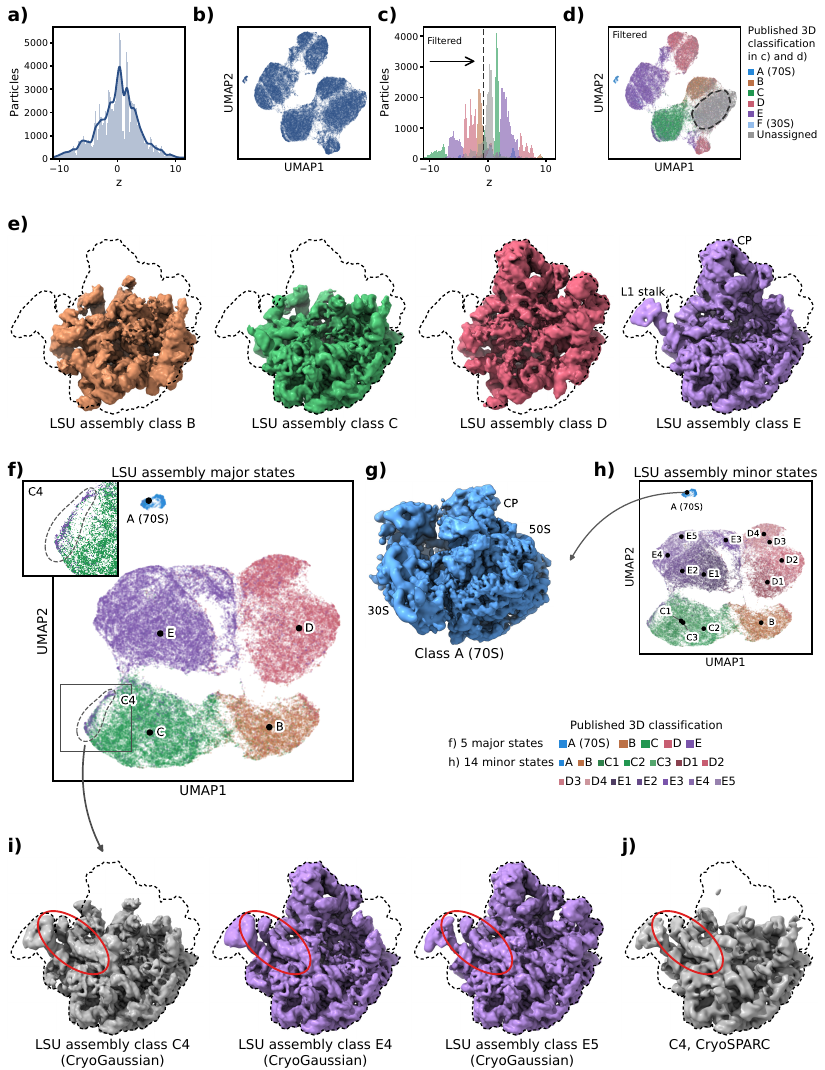}
    \caption{\textbf{Particle filtering and reconstruction of EMPIAR-10076 with CryoGaussian.}
\textbf{(a, b)} Inferred latents from one- and four-dimensional models trained on all
131,899 particle images: a latent histogram with a kernel-density estimate (a) and a
UMAP projection (b). \textbf{(c, d)} The same representations colored by published
class assignments~\citep{davis2016assembly}. The vertical dashed line in (c) marks the
lower boundary of the excluded latent interval; the dashed contour in (d) identifies
the excluded Gaussian-mixture component. The intersection of the two retained particle sets contains 97,651 images, of
which a random 90\% (87,885 images) are used for subsequent reconstruction.
\textbf{(e, g)} Reconstructed densities for assembly classes B--E and
the 70S class A. \textbf{(f, h)} Inferred latents colored by major and minor class
assignments; the inset identifies the released C4 particle
selection~\citep{zhong2021cryodrgn}. \textbf{(i)} C4, E4, and E5 densities.
\textbf{(j)} Homogeneous reconstruction of the selected C4 particle images with
CryoSPARC. Dashed lines trace the outline of the mature 50S subunit, and red ellipses mark
H68. Filtering and reconstruction settings are given in
Appendices~\ref{app:experimental_data} and~\ref{app:experimental_heterogeneity}.}
    \label{fig:empiar10076_filtering}
\end{figure*}

\begin{figure*}[p]
    \centering
    \includegraphics[width=\linewidth, height=\dimexpr\textheight-12\baselineskip\relax, keepaspectratio]{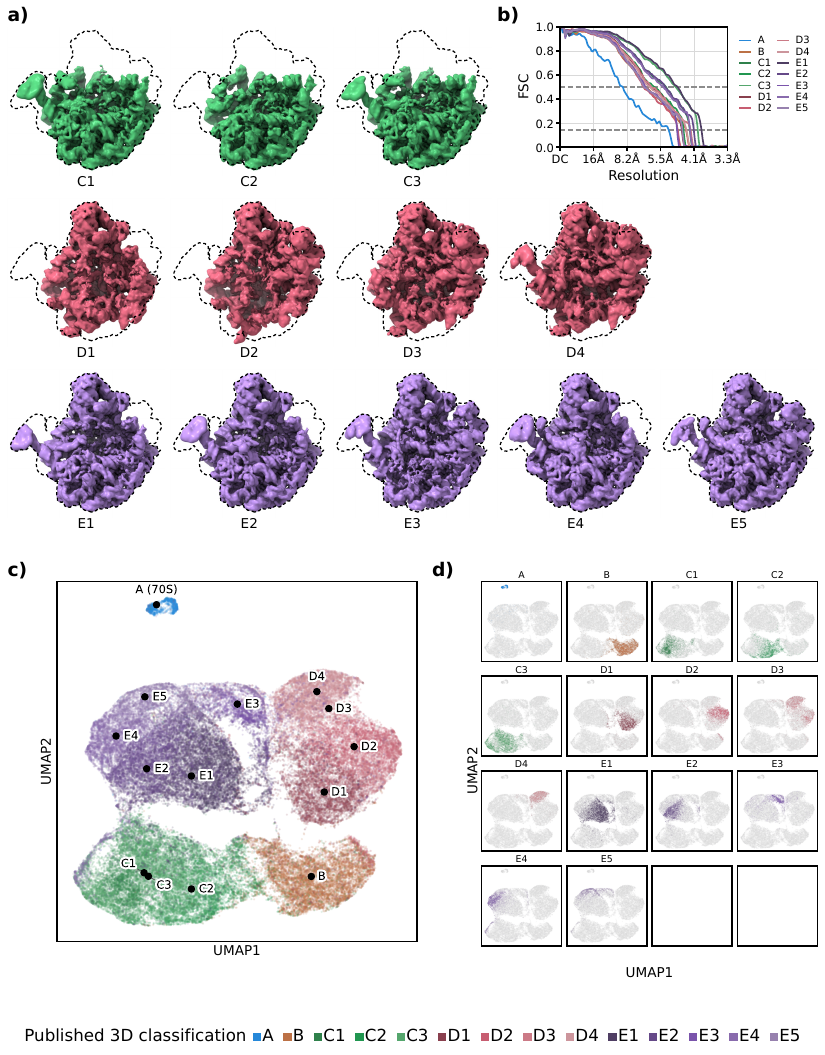}
    \caption{\textbf{Assembly-state reconstruction using the CryoGaussian particle selection for
EMPIAR-10076.} The model is trained on 87,885 particle images selected as described in
Figure~\ref{fig:empiar10076_filtering}. \textbf{(a)} Reconstructed densities for
C1--C3, D1--D4, and E1--E5, with dashed lines tracing the outline of the mature 50S subunit.
\textbf{(b)} Map-to-map FSC against the corresponding deposited
reconstructions~\citep{davis2016assembly}, including classes A and B. Horizontal
dashed lines indicate FSC values of 0.5 and 0.143. \textbf{(c, d)} UMAP projections
colored by published class assignments, shown jointly (c) and with one class
highlighted at a time (d). Colors are shared across panels (b)--(d). Evaluation
follows Figure~\ref{fig:empiar10076_supplied_states}.}
    \label{fig:empiar10076_filtered_states}
\end{figure*}

We examine EMPIAR-10076 using both the CryoDRGN-filtered and
CryoGaussian-filtered particle subsets.
Figure~\ref{fig:empiar10076_filtering} documents the CryoGaussian particle
selection (Appendix~\ref{app:experimental_data}). Of the
97,651 retained particles, a random 90\% (87,885 particles) are used for
subsequent heterogeneous reconstruction.
Figure~\ref{fig:empiar10076_filtered_states} shows the reconstructed minor
assembly states and their map-to-map FSC against deposited densities for
the CryoGaussian-filtered subset. Representative latents are selected as
described in Appendix~\ref{app:experimental_heterogeneity}.
Figure~\ref{fig:empiar10076_cross_comparison} compares reconstructed densities
for major assembly classes B--E with CryoDRGN and across the two
particle selections used for CryoGaussian reconstruction, and marks regions
in which the CryoDRGN density is fragmented and noise-like whereas both
CryoGaussian reconstructions are continuous.
Both particle selections recover the intermediate with H68 density but
no resolved central protuberance. Homogeneous CryoSPARC refinement
of the intersection between the released state selection and the
CryoGaussian-filtered particle set shows the same features
(Figure~\ref{fig:empiar10076_c4_comparison}(b)), as does refinement of the
954 particles selected in the CryoGaussian latent space
(Figure~\ref{fig:empiar10076_c4_comparison}(c, e)). Starting from the
unfiltered particle stack, and without the released state indices,
CryoGaussian accomplishes the particle selection, the identification of this
intermediate in its latent space, and its reconstruction.

\begin{figure*}[tp]
    \centering
    \includegraphics[width=1\linewidth]{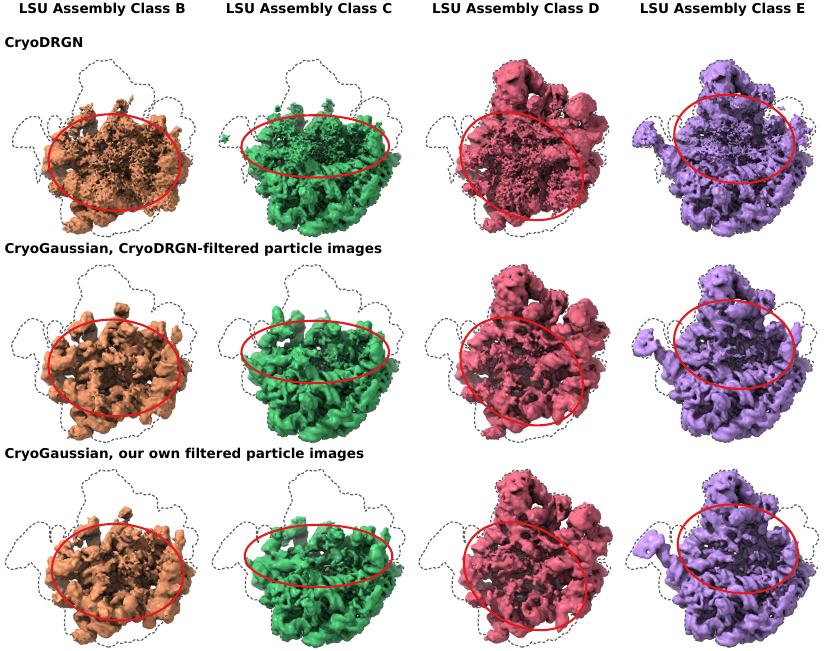}
    \caption{\textbf{Assembly-state comparison on EMPIAR-10076.}
    Columns show reconstructed densities for the major assembly classes B--E.
    Rows correspond to CryoDRGN (top), CryoGaussian using the CryoDRGN-filtered
    particle subset (middle), and CryoGaussian using the CryoGaussian-filtered
    subset (bottom). Dashed lines trace the outline of the mature 50S
    subunit. Red ellipses, drawn at corresponding positions in all three rows
    of each column, mark regions in which the CryoDRGN density is fragmented
    and noise-like, whereas both CryoGaussian reconstructions exhibit
    continuous density. Particle selection is described in
    Appendix~\ref{app:experimental_data}.}
    \label{fig:empiar10076_cross_comparison}
\end{figure*}

\subsection{Sensitivity to image noise}
\label{app:noise_sensitivity}

\begin{figure*}[tp]
    \centering
    \includegraphics[width=1\linewidth]{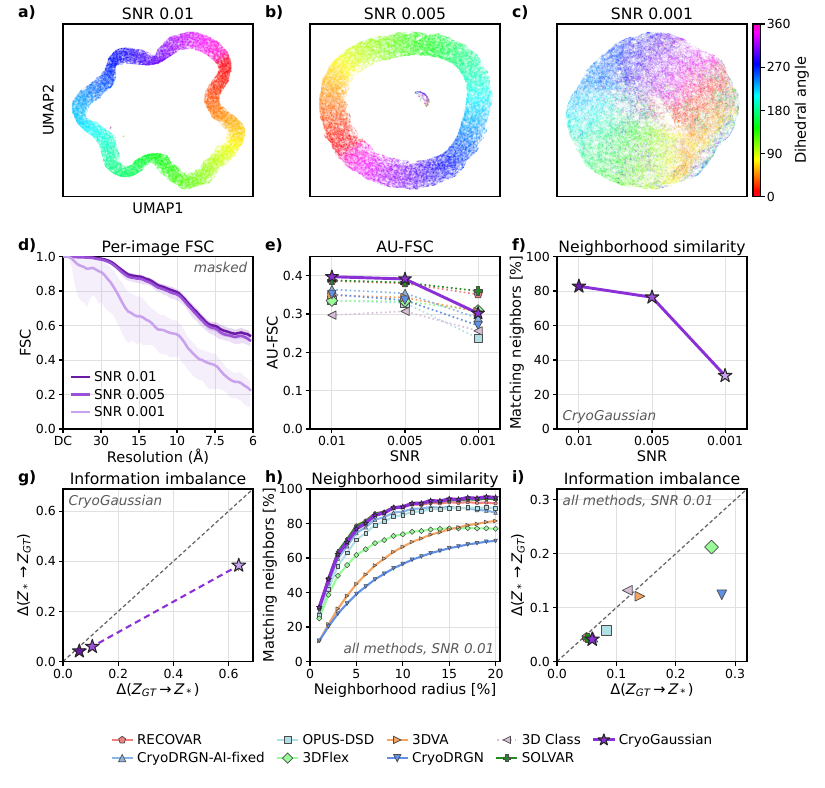}
    \caption{\textbf{Noise sensitivity of CryoGaussian on IgG-1D.}
    \textbf{(a--c)} UMAP projections of inferred latents at SNR 0.01,
    0.005, and 0.001, respectively, colored by the ground-truth dihedral
    angle. \textbf{(d)} Masked per-image FSC between reconstructed
    and reference densities. Solid curves show the mean, and shaded
    regions cover the 5th--95th percentiles across 100 evaluated
    particle images. \textbf{(e)} Mean per-image FSC AUC as a function
    of SNR for CryoGaussian and the compared methods.
    \textbf{(f, g)} Neighborhood similarity, averaged over neighborhood
    radii of 1--20\% of the evaluation subset, and information imbalance
    of the inferred latents at the three noise levels.
    \textbf{(h, i)} Corresponding comparisons with the other methods at
    SNR 0.01, as in Figure~\ref{fig:heterogeneity_latent_metrics}(a).
    The legend is shared by (e), (h), and (i), with stars denoting
    CryoGaussian.}
    \label{fig:igg1d_robustness}
\end{figure*}

We use the three released IgG-1D noise conditions at SNR 0.01, 0.005,
and 0.001~\citep{jeon2024cryobench}. Figure~\ref{fig:igg1d_robustness}
combines latent-space evaluation and per-image FSC for these conditions.
Mean masked per-image FSC AUC is 0.397, 0.391, and 0.302, respectively.
CryoGaussian achieves the highest mean per-image FSC AUC among the
compared methods at SNR 0.01 and 0.005. The decrease is more pronounced at
SNR 0.001. The corresponding mean per-conformation FSC AUC values are
0.402, 0.398, and 0.328 (Table~\ref{tab:cryobench_masking}). Neighborhood
similarity averaged over neighborhood radii of 1--20\% is 82.7\%, 76.4\%,
and 30.9\%, and information imbalance from inferred latents to the
ground-truth heterogeneity embedding is 0.041, 0.060, and 0.383 at the three noise
levels. At SNR 0.01, the information imbalance between the
ground-truth heterogeneity embedding and the inferred latents is 0.059 and
0.041 in the two directions, and the comparison with the other methods in
Figure~\ref{fig:igg1d_robustness}(h, i) corresponds to
Figure~\ref{fig:heterogeneity_latent_metrics}(a).

\begin{figure*}[p]
    \centering
    \includegraphics[width=\linewidth, height=\dimexpr\textheight-10\baselineskip\relax, keepaspectratio]{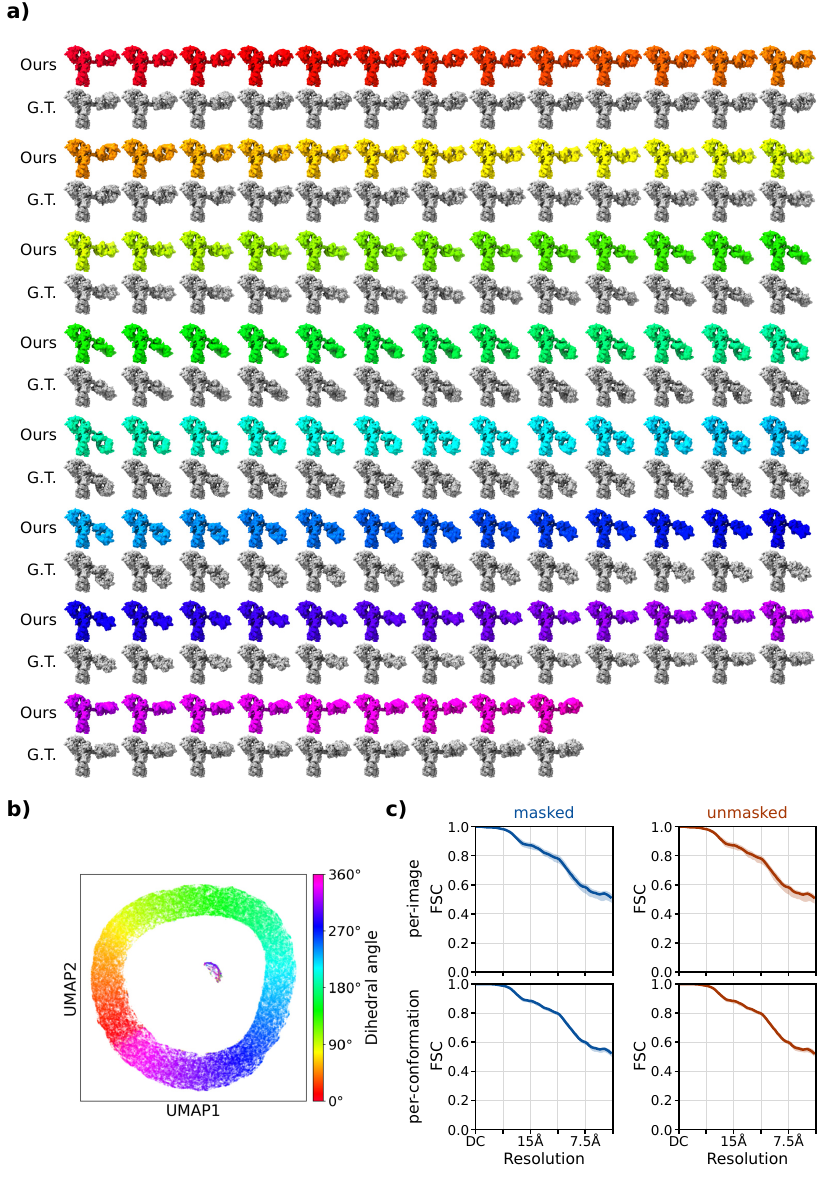}
    \caption{\textbf{IgG-1D reconstruction at SNR 0.005.} \textbf{(a)} Per-image reconstructed
densities and paired reference densities. \textbf{(b)} UMAP projection of inferred
latents, colored by the ground-truth dihedral angle as in
Figure~\ref{fig:igg1d_reconstruction}. \textbf{(c)} Per-image FSC (top) and
per-conformation FSC (bottom), with background masking (left) and without
masking (right). Curves and shading show the mean and 5th--95th percentiles across 100
evaluated particle images or 100 reference states, respectively.}
    \label{fig:igg1d_snr005}
\end{figure*}

\begin{figure*}[p]
    \centering
    \includegraphics[width=\linewidth, height=\dimexpr\textheight-10\baselineskip\relax, keepaspectratio]{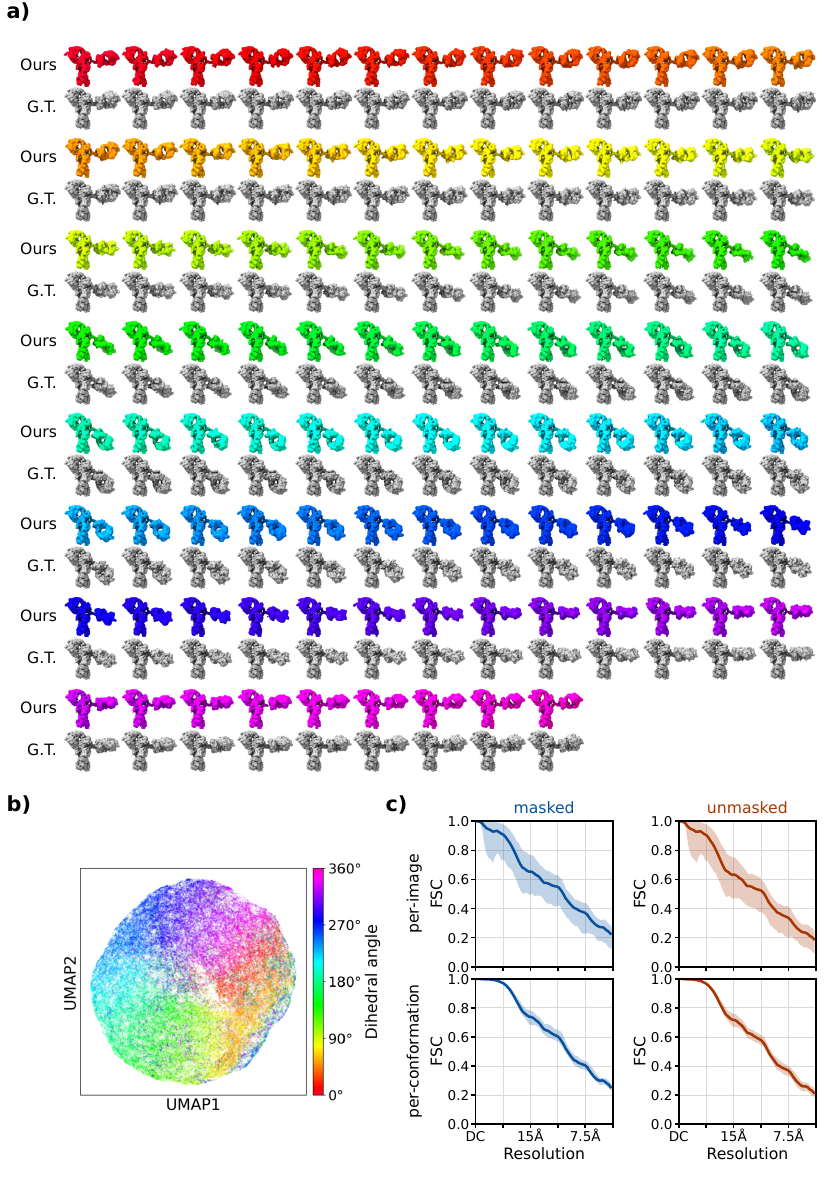}
    \caption{\textbf{IgG-1D reconstruction at SNR 0.001.} \textbf{(a)} Per-image reconstructed
densities and paired reference densities. \textbf{(b)} UMAP projection of inferred
latents, colored by the ground-truth dihedral angle as in
Figure~\ref{fig:igg1d_reconstruction}. \textbf{(c)} Per-image FSC (top) and
per-conformation FSC (bottom), with background masking (left) and without
masking (right). Curves and shading show the mean and 5th--95th percentiles across 100
evaluated particle images or 100 reference states, respectively.}
    \label{fig:igg1d_snr001}
\end{figure*}

Estimated numbers of Gaussians for these noise conditions are reported in
Table~\ref{tab:computational_cost}.
Figures~\ref{fig:igg1d_snr005} and~\ref{fig:igg1d_snr001}
show the lower-SNR reconstructed densities and FSC curves in detail.
Baseline sources and latent-space evaluation procedures for the composite
comparison are specified in Appendices~\ref{app:baseline_configurations}
and~\ref{app:latent_evaluation}.

\subsection{Sensitivity to masking}
\label{app:masking_sensitivity}

We assess the sensitivity of per-image and per-conformation FSC AUC to
background masking by evaluating the same reconstructed densities with and without the
evaluation mask (Table~\ref{tab:cryobench_masking}), following the
masked and unmasked comparison in CryoBench~\citep{jeon2024cryobench}.
Removing the mask changes the mean FSC AUC by at most 0.001 at the
displayed precision, except for IgG-1D at SNR 0.001, where the mean decreases
by approximately 0.010 for both per-image and per-conformation evaluation.
These results indicate limited sensitivity of the reported FSC AUC to
background masking, with a larger effect at the lowest SNR.

\begin{table*}[tp]
    \centering
    \caption{\textbf{Masked and unmasked FSC AUC for CryoGaussian on CryoBench.}
We report the mean (standard deviation), median, minimum, and maximum across evaluated particle
images for per-image FSC and reference states for per-conformation FSC
(Appendix~\ref{app:reference_fsc}).}
    \label{tab:cryobench_masking}
    \setlength{\tabcolsep}{3pt}
    \renewcommand{\arraystretch}{1.15}
    \small
    \resizebox{\linewidth}{!}{%
        \begin{tabular}{@{}ll*{2}{ccc}@{}}
            \toprule
            Dataset & Evaluation
            & \multicolumn{3}{c}{Masked}
            & \multicolumn{3}{c}{Unmasked} \\
            \cmidrule(lr){3-5}
            \cmidrule(lr){6-8}
            & & Mean (std) & Median & Min--max
            & Mean (std) & Median & Min--max \\
            \midrule
            IgG-1D (SNR 0.01) & Per-image
            & 0.397 (0.004) & 0.398 & 0.382--0.404
            & 0.397 (0.004) & 0.398 & 0.382--0.403 \\
             & Per-conformation
            & 0.402 (0.001) & 0.402 & 0.398--0.404
            & 0.401 (0.001) & 0.401 & 0.397--0.404 \\
            \addlinespace[2pt]
            IgG-1D (SNR 0.005) & Per-image
            & 0.391 (0.012) & 0.394 & 0.296--0.400
            & 0.391 (0.012) & 0.394 & 0.296--0.399 \\
             & Per-conformation
            & 0.398 (0.001) & 0.398 & 0.394--0.402
            & 0.397 (0.002) & 0.397 & 0.393--0.401 \\
            \addlinespace[2pt]
            IgG-1D (SNR 0.001) & Per-image
            & 0.302 (0.030) & 0.316 & 0.205--0.336
            & 0.292 (0.030) & 0.303 & 0.200--0.331 \\
             & Per-conformation
            & 0.328 (0.005) & 0.328 & 0.315--0.339
            & 0.318 (0.005) & 0.318 & 0.302--0.329 \\
            \addlinespace[2pt]
            IgG-RL & Per-image
            & 0.366 (0.020) & 0.370 & 0.318--0.400
            & 0.366 (0.020) & 0.369 & 0.318--0.400 \\
             & Per-conformation
            & 0.387 (0.010) & 0.387 & 0.352--0.410
            & 0.387 (0.010) & 0.387 & 0.352--0.410 \\
            \addlinespace[2pt]
            Ribosembly & Per-image
            & 0.432 (0.014) & 0.432 & 0.386--0.452
            & 0.432 (0.014) & 0.432 & 0.386--0.452 \\
             & Per-conformation
            & 0.432 (0.016) & 0.432 & 0.391--0.454
            & 0.432 (0.016) & 0.432 & 0.391--0.454 \\
            \addlinespace[2pt]
            Tomotwin-100 & Per-image
            & 0.335 (0.035) & 0.341 & 0.217--0.387
            & 0.335 (0.035) & 0.341 & 0.217--0.387 \\
             & Per-conformation
            & 0.336 (0.034) & 0.343 & 0.234--0.390
            & 0.336 (0.034) & 0.342 & 0.234--0.389 \\
            \addlinespace[2pt]
            Spike-MD & Per-image
            & 0.364 (0.012) & 0.366 & 0.326--0.386
            & 0.364 (0.012) & 0.366 & 0.326--0.386 \\
             & Per-conformation
            & 0.365 (0.012) & 0.367 & 0.327--0.386
            & 0.365 (0.012) & 0.367 & 0.327--0.386 \\
            \bottomrule
        \end{tabular}%
    }
\end{table*}

\clearpage
\section{Estimating the number of Gaussians}
\label{app:space_bandwidth}

This appendix gives the basis of the space--bandwidth rule for the number of
Gaussians, the backprojection used to estimate spatial support, and the interpretation of the
relative density threshold.

\subsection{Space--bandwidth basis}

We derive the space--bandwidth product in Equation~\eqref{eq:gaussian_count}
using the multidimensional space--frequency concentration
problem~\citep{slepian1964multidimensional,katsevich2015covariance}. Let $\Omega\subset\mathbb R^3$ be a
measurable spatial region of finite volume and $B\subset\mathbb R^3$ a frequency
region of finite volume. We use the Fourier convention
\begin{equation}
  \widehat f(\boldsymbol\xi)
  =\int_{\mathbb R^3}f(\mathbf x)
      e^{-2\pi\mathrm i\boldsymbol\xi^{\mathsf T}\mathbf x}\,d\mathbf x.
  \label{eq:capacity_fourier_convention}
\end{equation}
The concentration operator on $L^2(\Omega)$ is
\begin{equation}
\begin{aligned}
  {}[T_{\Omega,B}f](\mathbf x)
  &=\int_{\Omega}K_B(\mathbf x,\mathbf y)f(\mathbf y)\,d\mathbf y, \\
  K_B(\mathbf x,\mathbf y)
  &=\int_Be^{2\pi\mathrm i\boldsymbol\xi^{\mathsf T}(\mathbf x-\mathbf y)}
        \,d\boldsymbol\xi.
\end{aligned}
\label{eq:concentration_kernel}
\end{equation}
To justify the trace, define $\mathcal A:L^2(\Omega)\to L^2(B)$ by
\begin{equation}
  [\mathcal A f](\boldsymbol\xi)
  =\int_{\Omega}f(\mathbf x)
       e^{-2\pi\mathrm i\boldsymbol\xi^{\mathsf T}\mathbf x}\,d\mathbf x,
  \qquad \boldsymbol\xi\in B.
\end{equation}
Then $T_{\Omega,B}=\mathcal A^*\mathcal A$. Zero extension and Plancherel's theorem
give $\|\mathcal A\|\leq1$, while
$\|\mathcal A\|_{\mathrm{HS}}^2=\int_B\int_\Omega1\,d\mathbf x\,d\boldsymbol\xi
=|\Omega||B|<\infty$. Thus $T_{\Omega,B}$ is positive and trace class, has
eigenvalues $\lambda_j\in[0,1]$, and its trace is
\begin{align}
  \mathcal S(\Omega,B)
  &=\sum_j\lambda_j
   =\int_{\Omega}K_B(\mathbf x,\mathbf x)\,d\mathbf x \\
  &=\int_{\Omega}\int_B1\,d\boldsymbol\xi\,d\mathbf x
   =|\Omega|\,|B|.
  \label{eq:concentration_trace}
\end{align}
The trace is the space--bandwidth product. It provides a nominal effective dimension
for spatially and spectrally concentrated signals; its interpretation as a count of
well-concentrated modes is approximate, whereas the trace identity is
exact~\citep{simons2011spatiospectral}.

For the ball $B_d=\{\boldsymbol\xi:\|\boldsymbol\xi\|_2\leq1/d\}$,
\begin{equation}
  |B_d|=\int_0^{1/d}4\pi r^2\,dr
       =\frac{4\pi}{3d^3}.
\end{equation}
Treating the above-threshold voxels as a union of physical voxel cells gives
$|\Omega_\tau|=V_\tau=n_\tau a^3$. At $d=2a$,
\begin{align}
  \mathcal S_\tau=\mathcal S(\Omega_\tau,B_d)
  &=\frac{4\pi}{3}\frac{V_\tau}{d^3}
   =\frac{4\pi}{3}\frac{n_\tau a^3}{(2a)^3} \\
  &=\frac{4\pi}{3}\frac{n_\tau a^3}{8a^3}
   =\frac{\pi}{6}n_\tau.
  \label{eq:count_cancellation}
\end{align}
The factor $4\pi/3$ is the volume factor of the frequency-domain ball; the spatial
support need not be spherical or connected. We obtain the number of Gaussians
$G=\mathcal R(\mathcal S_\tau)$ by rounding the space--bandwidth product to an
integer.

The concentration identity concerns a linear space of signals, and we use
its value as the number of Gaussians in the nonlinear Gaussian
representation. The relative density threshold $\tau$ specifies the
estimated spatial support, and the sampling bandwidth specifies the spatial
scale at which the product is evaluated. The Gaussian FWHM constraint uses
the same spatial scale.

\subsection{Backprojection for spatial support estimation}
\label{app:capacity_backprojection}

We compute a CTF-weighted backprojection, the closed-form linear
reconstruction of the particle images at their supplied orientations, using
regularized gridding in CryoDRGN (version 3.4.4)~\citep{zhong2021cryodrgn}.
The backprojection uses all particle images in the corresponding dataset.
Particle images are corrected for the supplied
in-plane translations. Fourier
downsampling, when used, preserves the physical field of view. After any downsampling,
the pixel size $a$ is also the voxel spacing used for spatial support estimation. The supplied orientations are used to backproject the particle
images. Let $\mathcal X$ denote the voxel centers of the resulting density
$\rho_{\mathrm{BP}}$ on a grid with spacing $a$. For a relative density
threshold $\tau\in(0,1)$, the estimated spatial support is defined by
\begin{equation}
\begin{aligned}
  \rho_+(\mathbf x)&=\max(\rho_{\mathrm{BP}}(\mathbf x),0), \\
  \mathcal X_{\tau}
  &=\left\{\mathbf x\in\mathcal X:\rho_+(\mathbf x)\geq
                  \tau\max_{\mathbf x'\in\mathcal X}\rho_+(\mathbf x')\right\},
\end{aligned}
\label{eq:estimated_support}
\end{equation}
assuming $\max_{\mathbf x\in\mathcal X}\rho_+(\mathbf x)>0$, and
$\Omega_\tau$ is the union of the voxel cells of $\mathcal X_\tau$, with
$n_\tau=|\mathcal X_\tau|$. Only the number of Gaussians derived
from this support is used in subsequent reconstruction, and Gaussian attributes are
not initialized from the backprojected density.

\subsection{Interpretation of the relative density threshold}
\label{app:density_threshold}

The relative density threshold $\tau$ defines the estimated spatial support
$\Omega_\tau$ used in the space--bandwidth product, and its numerical value is reported
in Section~\ref{sec:gaussian_capacity}. Thresholding relative to the positive maximum
makes this estimate invariant to a positive multiplicative rescaling of the
backprojected density. A lower threshold includes more weak-density regions in the
estimated spatial support.

\paragraph{Support-volume interpretation.}
Let $\rho_{\max}=\max_{\mathbf x\in\mathcal X}\rho_+(\mathbf x)>0$.
The voxel set $\mathcal X_\tau$ admits an equivalent characterization as a
maximizer of a discrete excess-mass criterion:
\begin{equation}
  \mathcal X_\tau\in
  \operatorname*{arg\,max}_{\mathcal U\subseteq\mathcal X}
  \left[
    \sum_{\mathbf x\in\mathcal U}
      \frac{\rho_+(\mathbf x)}{\rho_{\max}}
    -\tau|\mathcal U|
  \right].
  \label{eq:support_excess_mass}
\end{equation}
Each voxel contributes $\rho_+(\mathbf x)/\rho_{\max}-\tau$; selecting all
nonnegative contributions gives $\mathcal X_\tau$. This criterion balances
retained peak-normalized density against support size. Since
$V_\tau=a^3n_\tau$ and $\mathcal S_\tau=(\pi/6)n_\tau$, the support-size
penalty also controls the space--bandwidth product used to set $G$.

\paragraph{Gaussian reference.}
A single three-dimensional Gaussian provides an analytical reference for
$\tau=0.02$. Its density superlevel sets are highest-density regions: each
encloses a given probability mass within the minimum
volume~\citep{hyndman1996hdr}. Let
$\varphi(\mathbf r)=\mathcal N_3(\mathbf r;\mathbf0,I_3)$ and
let $F_{\chi^2_3}$ denote the cumulative distribution function of a chi-square
random variable with three degrees of freedom. Then
\begin{align}
  \frac{\varphi(\mathbf r)}{\varphi(\mathbf0)}\geq\tau
  &\quad\Longleftrightarrow\quad
    \|\mathbf r\|_2^2\leq-2\ln\tau, \\
  \int_{\varphi(\mathbf r)\geq\tau\varphi(\mathbf0)}
       \varphi(\mathbf r)\,d\mathbf r
  &=F_{\chi^2_3}(-2\ln\tau).
  \label{eq:gaussian_threshold_reference}
\end{align}
The 95-percent mass ellipsoid has relative boundary density
$\exp[-F_{\chi^2_3}^{-1}(0.95)/2]\approx0.0201$; the relevant chi-square quantile
is approximately $7.815$~\citep{nistChiSquareTable}. Thus, $\tau=0.02$
approximates this relative boundary density and encloses approximately $0.9502$
of the Gaussian mass. An invertible affine coordinate transformation gives
the same relation for any nondegenerate three-dimensional Gaussian.
This calculation supplies an analytical reference for the fixed threshold;
the fraction of the integrated positive backprojected density enclosed by
$\Omega_\tau$ is data dependent.

\paragraph{Dependence on the threshold.}
For a fixed backprojection and voxel grid, two thresholds
$0<\tau_1<\tau_2<1$ satisfy
\begin{equation}
  n_{\tau_1}-n_{\tau_2}
  =\left|\left\{\mathbf x\in\mathcal X:
      \tau_1\leq\frac{\rho_+(\mathbf x)}{\rho_{\max}}<\tau_2
    \right\}\right|.
  \label{eq:threshold_count_difference}
\end{equation}
The change in estimated spatial support is therefore determined exactly by
the voxels whose normalized densities lie between the two thresholds.
Its effect on the number of Gaussians follows through
$G=\mathcal R((\pi/6)n_\tau)$. The space--bandwidth identity gives $\mathcal S_\tau$ conditional on
$\Omega_\tau$; it does not determine $\tau$.
Figure~\ref{fig:gaussian_count_sensitivity} and Table~\ref{tab:tau_sensitivity}
evaluate the empirical effect of varying $\tau$ on heterogeneity estimation
and per-state reconstruction; the corresponding results are discussed in
Appendix~\ref{app:count_sensitivity}.

\section{Efficient cryo-EM Gaussian splatting}
\label{app:renderer_optimization}

This appendix describes the gradient reductions and the partitioning of Gaussian
lists for image tiles underlying
Section~\ref{sec:efficient_optimization}. We use the same Gaussian density,
CTF modulation, and reconstruction objective as in
Section~\ref{sec:gaussian_projection}, where the projection operator in
Equation~\eqref{eq:analytic_projection} is the line integral
$[\mathcal P_{R_i}\rho_{\Theta}](\mathbf v;\mathbf z_i)
=\int_{\mathbb R}\rho_{\Theta}(R_i^{\mathsf T}(v_1,v_2,w)^{\mathsf T};\mathbf z_i)\,dw$
along the beam direction $w$. Let $\psi_{ig}(\mathbf v)$ denote the
contribution of Gaussian $g$ to the predicted image $\widehat Y_i^{\mathrm{pre}}$
before CTF modulation: it is the $g$th summand in Equation~\eqref{eq:analytic_projection},
multiplied by $\kappa$ in
Equation~\eqref{eq:ctf_projection}. Rasterization groups the projected Gaussian
footprints into image tiles. A Gaussian is assigned to a tile only when its
footprint overlaps that tile, as in Speedy-Splat~\citep{hanson2025speedy},
rather than to every tile overlapping the axis-aligned bounding box of the
footprint. The following reductions refer to the Gaussian--pixel
interactions selected by this assignment.

\subsection{Gradients through CTF modulation}
\label{app:ctf_adjoint}

The CTF-modulated prediction is related to the projection by a linear operator.
Applying its adjoint to the image residual gives the gradient supplied to the
rasterizer:
\begin{equation}
  \delta_i
  \equiv\frac{\partial\mathcal L}
               {\partial\widehat Y_i^{\mathrm{pre}}}
  =\frac{2}{NHW}\mathcal F_2^{-1}\!\left[
       \overline{C_i}\,
       \mathcal F_2\!\left[\widehat Y_i-Y_i^{\mathrm{aligned}}\right]
     \right],
  \label{eq:ctf_adjoint}
\end{equation}
where the overline denotes complex conjugation; the CTFs used here are real-valued.
The factor $2/(NHW)$ follows from Equation~\eqref{eq:image_objective}.
CTF modulation couples spatial locations in the image domain, while the rasterizer receives
$\delta_i$ after this operation has been differentiated. It can therefore accumulate
gradient contributions over each projected Gaussian footprint using $\delta_i$.

\subsection{Per-Gaussian gradient accumulation}
\label{app:gaussian_gradient_accumulation}

For differentiation, regard the effective Gaussian attributes for each particle
image as separate intermediate variables,
$\boldsymbol\vartheta_{ig}=(p_g(\mathbf z_i),\boldsymbol\mu_g(\mathbf z_i),
\boldsymbol\Sigma_g(\mathbf z_i))$. These attributes are functions of $\Theta$ and, in heterogeneous reconstruction,
$\mathbf z_i$, as defined in Section~\ref{sec:gaussian_projection}.
Let $\mathcal V_{ig}$ be the image-grid locations assigned to Gaussian $g$ by the
rasterizer. With these interactions held fixed, the gradient reduction is
\begin{equation}
  \nabla_{\boldsymbol\vartheta_{ig}}\mathcal L
  =\sum_{\mathbf v\in\mathcal V_{ig}}
       \delta_i(\mathbf v)\,
       \nabla_{\boldsymbol\vartheta_{ig}}\psi_{ig}(\mathbf v).
  \label{eq:gaussian_gradient_reduction}
\end{equation}
A pixel-wise scatter computes contributions at each pixel and uses atomic updates
to accumulate them into Gaussian gradients. Our gather implementation assigns one
thread block to each Gaussian for each particle image. Threads in that block
accumulate contributions to the projected Gaussian attributes and reduce them before
writing the result, avoiding concurrent writes from different blocks to that
Gaussian's gradient in this reduction. The projection derivatives then propagate
these gradients to the three-dimensional Gaussian attributes in
Equation~\eqref{eq:gaussian_gradient_reduction}. The per-splat backward pass
of Taming 3DGS~\citep{mallick2024taming} performs this reduction under
depth-ordered compositing, where it also requires the compositing state of
the pixels in front of each splat. In additive projection,
Equation~\eqref{eq:gaussian_gradient_reduction} involves only $\delta_i$ and
the footprint of the Gaussian.

For attributes shared across particle images, the resulting gradients are summed
across those images. Gradients for attributes decoded separately for each particle
image are kept separate until they are propagated through the Gaussian attribute
decoders to the trainable representation parameters.

\subsection{Partitioned additive projection}
\label{app:partitioned_projection}

Let $\mathcal G_i(\mathcal T)$ be the Gaussian list assigned to image tile
$\mathcal T$ for particle image $i$. A long list can be divided into disjoint
consecutive sublists $\mathcal G_i^{(q)}(\mathcal T)$ whose union is
$\mathcal G_i(\mathcal T)$. For each pixel location $\mathbf v$ in that tile,
linearity gives
\begin{equation}
  \sum_{g\in\mathcal G_i(\mathcal T)}\psi_{ig}(\mathbf v)
  =\sum_q\sum_{g\in\mathcal G_i^{(q)}(\mathcal T)}
     \psi_{ig}(\mathbf v).
  \label{eq:partitioned_projection}
\end{equation}
Each inner sum forms a partial projection that can be evaluated independently,
and the partial projections are then accumulated. This exposes additional
parallelism when a small number of tiles contain long Gaussian lists. Under
depth-ordered compositing, 3DGS implementations balance such lists by
redistributing work across thread blocks~\citep{gui2024balanced,wang2024adr}.
Linearity makes the partition direct here.
Batching particle images also supplies parallel work and amortizes preprocessing
and scheduling overhead. We retain batched projection when the batch supplies sufficient parallelism and
partition long tile lists when parallelism across particle images is limited.

The partitioning of Gaussian lists for image tiles and the gradient reductions
above retain the same Gaussian--pixel interactions. The forward pass
allocates workspaces for the projected geometry of the Gaussians, the
Gaussian--tile keys and their sorted order, and the per-tile ranges, which
3DGS rasterizers retain for the backward pass because back-to-front
compositing requires the sorted lists and the per-pixel
transmittance~\citep{kerbl2023gaussian,ye2025gsplat}. In additive
projection, Equation~\eqref{eq:gaussian_gradient_reduction} requires
neither, so the rasterizer releases these workspaces once the projections
are formed, retaining the Gaussian parameters, the viewing matrices, the
projections, and the computation graph to the Gaussian attribute decoders,
and the backward pass recomputes the projected geometry of each Gaussian
from the retained quantities and gathers gradients over the pixel footprint
determined by its projected center and radius. This trades recomputation
for memory, as in activation recomputation~\citep{chen2016training}, and
the memory held between the two passes scales with the number of Gaussians
rather than with the total length of the tile lists. Releasing the
workspaces lowers peak allocated memory per GPU by 8.0--10.2\% on the
datasets of Table~\ref{tab:rasterizer_speedup}
(Appendix~\ref{app:computational_cost}).

\section{Future directions}
\label{app:extended_discussion}
\label{app:future_implications}

CryoGaussian could serve as a reconstruction component in agentic workflows
that repeatedly select particles, reconstruct structural states, and
evaluate their agreement with experimental observations. Scientific agents
such as Coscientist demonstrate how language models can coordinate
specialized tools and use feedback to plan
experiments~\citep{boiko2023autonomous}. In cryo-EM, such a workflow could
invoke CryoGaussian to compare alternative particle selections and
reconstruct candidate structural subpopulations. Accelerating projection
and gradient computation with our CUDA rasterizer and distributed training
could reduce the cost of these repeated analyses, allowing more structural
hypotheses to be examined at a given computational cost. Where local map
quality supports atomic interpretation, reconstructed states could be
passed to automated model-building tools such as
ModelAngelo~\citep{jamali2024automated}, connecting heterogeneity analysis
to molecular models for downstream study.

A further direction is to connect experimentally resolved structural
states with molecular design. ProteinZero improves inverse-folding models
through online reinforcement learning with computational
feedback~\citep{wang2026proteinzero}, while PocketXMol uses atomic prompts
to unify structure prediction and the design of pocket-interacting small
molecules and peptides~\citep{peng2026pocketxmol}. These capabilities
motivate workflows in which reconstructed conformations inform structural
targets or binding environments, and subsequent experiments test the
resulting designs. CryoGaussian could provide the reconstruction component
as new particle images become available, enabling comparisons across
experimental conditions and structural states. Evaluating this integration
would establish whether accelerated reconstruction can make iterative
structure-guided design more practical.

\end{document}